\documentclass{article}

\usepackage[accepted]{util/icml2020}

\usepackage[utf8]{inputenc} 
\usepackage[T1]{fontenc}    
\usepackage{hyperref}       
\usepackage{url}            
\usepackage{booktabs}       
\usepackage{amsfonts}       
\usepackage{amsmath}
\usepackage{nicefrac}       
\usepackage{microtype}      
\usepackage{graphicx}
\usepackage{subcaption}
\usepackage{wrapfig}
\usepackage[ruled,vlined,algo2e]{algorithm2e}
\usepackage{enumitem}
\usepackage{amsthm}
\usepackage{todonotes}

\SetKwInput{KwInput}{In}
\SetKwInput{KwOutput}{Out}

\icmltitlerunning{Real-time Classification using Input-filtering Neural ODEs}

\begin{document}

\twocolumn[
\icmltitle{Real-time Classification from Short Event-Camera Streams \\
           using Input-filtering Neural ODEs}

\icmlkeywords{DVS, Event Camera, ODE, Neural, NODE, continuous}


\begin{icmlauthorlist}
\icmlauthor{Giorgio Giannone}{nn}
\icmlauthor{Asha Anoosheh}{nn}
\icmlauthor{Alessio Quaglino}{nn}
\icmlauthor{Pierluca D'Oro}{nn}
\icmlauthor{Marco Gallieri}{nn}
\icmlauthor{Jonathan Masci}{nn}
\end{icmlauthorlist}

\icmlaffiliation{nn}{NNAISENSE, Lugano, Switzerland}
\icmlcorrespondingauthor{}{\texttt{\{firstname\}@nnaisense.com}}
\vskip 0.3in

{\renewcommand\twocolumn[1][]{#1}
\begin{center}
    \centering
    \includegraphics[width=.9\textwidth,height=\textheight,keepaspectratio]{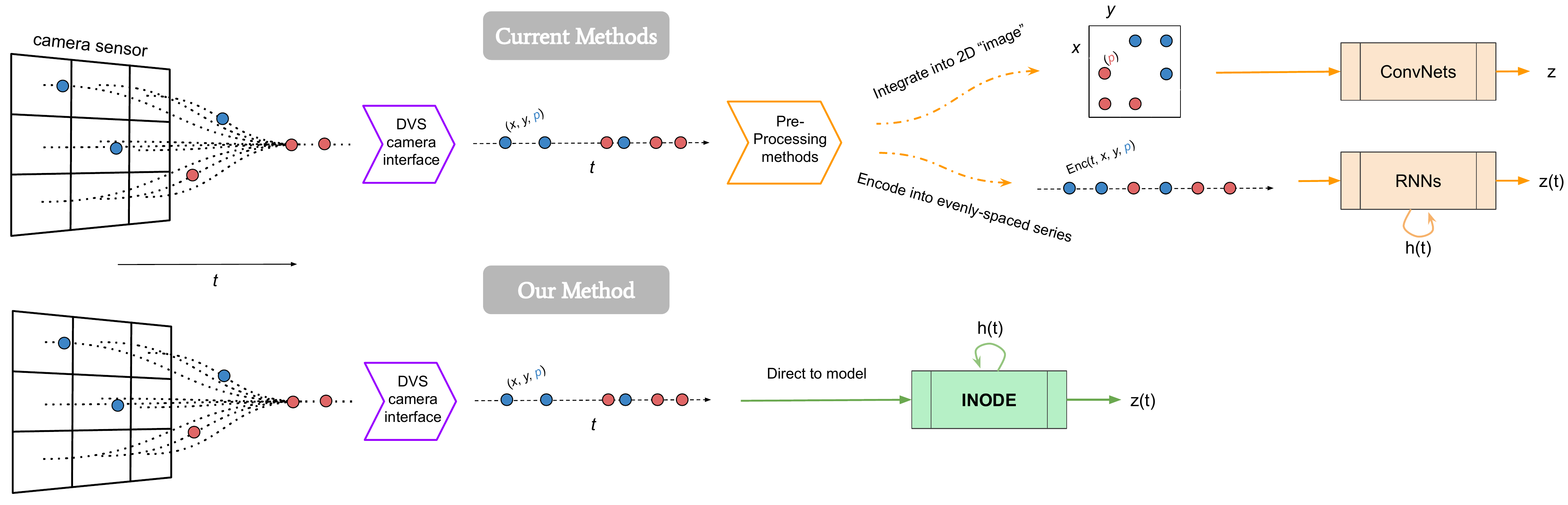}
    \captionof{figure}{{\bf Approach rationale.} The standard way to perform machine learning on asynchronous event stream data from DVS cameras consists in either integrating/binning events into a 2D grid to be fed into convolutional networks -- wherein there is a trade-off between sparsity and timestamp information loss -- or converting them into traditional time series with some time discretization scheme -- wherein the model must somehow consider the effect of time as the one of any other input feature. On the contrary, our method requires no preprocessing or loss of information, and inherently handles the data stream's asynchronous timing. In the figure, blue and red dots represent events of different polarity respectively.}
\label{fig:scheme}
\end{center}
\hrulefill
\vskip 0.2in}

\begin{abstract}
\vskip 0.2in
Event-based cameras are novel, efficient sensors inspired by the human vision system, 
generating an asynchronous, pixel-wise stream of data. 
Learning from such data is generally performed through heavy preprocessing and event integration into images. This requires buffering of possibly long sequences and can limit the response time of the inference system. 
In this work, we instead propose to directly use events from a DVS camera, a stream of intensity changes and their spatial coordinates. This sequence is used as the input for a novel \emph{asynchronous} RNN-like architecture, the Input-filtering Neural ODEs (INODE). This is inspired by the dynamical systems and filtering literature. INODE is an extension of Neural ODEs (NODE) that allows for input signals to be continuously fed to the network, like in filtering. The approach naturally handles batches of time series with irregular time-stamps by implementing a batch forward Euler solver. INODE is trained like a standard RNN, it learns to discriminate short event sequences and to perform event-by-event online inference. We demonstrate our approach on a series of classification tasks, comparing against a set of LSTM baselines. We show that, independently of the camera resolution, INODE can outperform the baselines by a large margin on the ASL task and it's on par with a much larger LSTM for the NCALTECH task. Finally, we show that INODE is accurate even when provided with very few events. 

\vskip 0.2in
\end{abstract}
]

\printAffiliationsAndNotice{}

\newcommand{\todopier}[1]{\todo[color=orange, inline]{#1}}
\newcommand{\todopierout}[1]{\todo[color=orange]{#1}}



\section{Introduction}
Event-based cameras are asynchronous sensors that capture changes in pixel intensity as binary events, with very high frequency compared to RGB sensors. This makes them suitable for high speed applications, such as robotics \cite{kim2016real, dimitrova2019towards} and other safety-critical scenarios. The Dynamic Vision Sensor (DVS) \cite{dvs_tobi_2008} is an event camera that, compared to traditional sensors, has low power consumption, high dynamic range, no motion blur, and microsecond latency times.

Unfortunately, due to their asynchronous and binary format, there is no obvious choice of a model class for handling DVS data, unlike the predominant use of convolution-based models for RGB images. In this paper, we propose the use of a deep-learning and differential-equation hybrid method for such tasks, inspired by Neural Ordinary Differential Equations, (NODE) \cite{Chen2018NeuralOD}. NODE pioneered a novel machine learning approach where the data is modeled as an ODE in latent space, which can in principle be adjusted to process multiple asynchronous inputs.

Most recent works using machine-learning to model DVS data integrate individual events to convert them into formats that can be fed as input into existing models, but lose precise timing information. The work of \cite{akolkar_2015_question} studies the benefit of using precise temporal event data over aggregated event techniques. In particular, the study states: 
\begin{quote}\small
The use of information theory to characterize separability between classes for each temporal resolution shows that high temporal acquisition provides up to 70\% more information than conventional spikes generated from frame-based acquisition as used in standard artificial vision, thus drastically increasing the separability between classes of objects.
\end{quote}
This provides motivation to research methods that can directly handle asynchronous data.

\paragraph{Summary of contributions.} This work develops a novel real-time online classification model for event-based camera data streams. Moreover, it proposes INODE, an extension of the NODE architecture, which can directly take as input the stream of a possibly-high-frequency signal. This can be seen a continuous-time extension of Recurrent Neural Networks (RNNs). INODE is trained to perform continuous-time event filtering in order to infer classification labels online, based on its hidden state at a given moment. At test time, the classification prediction and the hidden state are updated as each (asynchronous) camera event is received. The event polarity and spatial coordinates are fed directly as inputs to the network without using convolutional layers or event integration. Importantly, we remark that we do not process input data in any form beyond normalization. 


\paragraph{Summary of experiments.}
We demonstrate that the proposed approach excels in sample efficiency and real-time performance, significantly outperforming several LSTM architectures using short sequencing during online inference at test time. Furthermore, our method works with raw, noisy camera readings and is also invariant to the camera resolution used to capture the data.

\section{Related Works}
We review previous works related to our method, first describing alternative approaches to process events and discussing their relative advantages, then briefly introducing NODE methods.

\subsection{Learning from event data}
Event data from DVS cameras, being asynchronously streamed per sensor array pixel, requires careful processing to be compatible with traditional machine learning models. Methods for handling event data can be, in general, divided into grouped-event-based and {per\nobreakdash-event\nobreakdash-based}. The former employ a scheme to integrate multiple events into a single data structure that can be handled by spatially-based (e.g., convolutional) models, while the latter process the data stream on an event-by-event basis. Figure \ref{fig:scheme} illustrates the main differences between the reviewed works and the proposed approach.  


\begin{figure}[t]
    \centering
    \includegraphics[trim=175 225 460 80,clip, width=0.99\columnwidth]{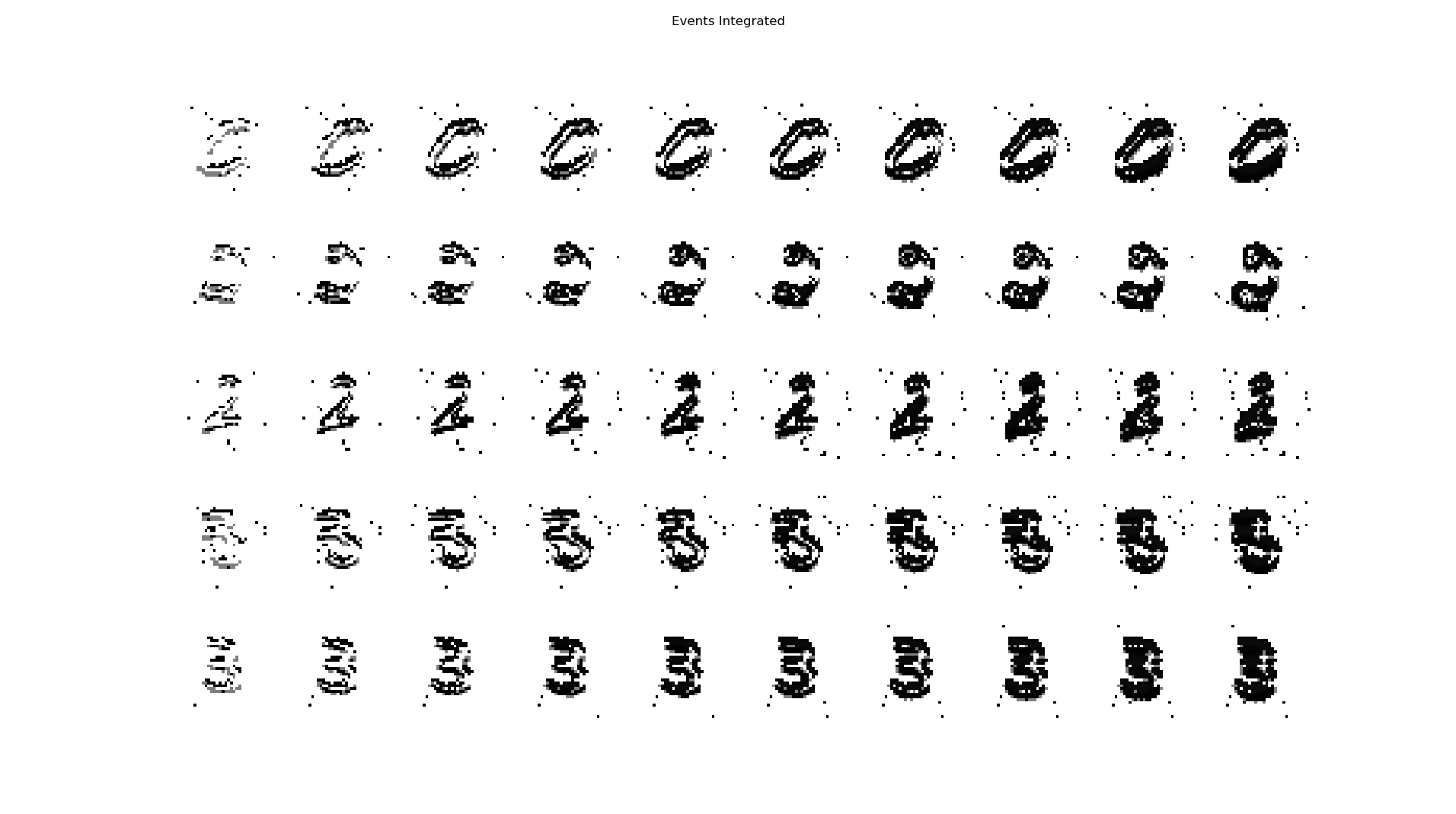}
    \caption{{\bf Event integration.} Most standard methods rely on accumulating potentially-hundreds of events in order to obtain images. This figure shows the result of integrating 7 $\times$ 100 consecutive events into a pixel grid. Our method trains and performs inference \emph{without directly integrating or buffering} events, but instead processing \emph{one event at a time}.} 
\label{fig:integration}
\end{figure}

\paragraph{Grouped-event methods.}
One of the more evident strategies in this category is to integrate time windows of data into grayscale intensity images, then apply existing computer vision techniques on these reconstructions. This is used, for example, in optical flow estimation~\cite{bardow2016simultaneous}, SLAM~\cite{kim2016real} and face recognition~\cite{barua_2016}. Such a process requires various filtering, tracking, and/or inertial measurement integration to properly compute frame offsets. This integration method itself is also the subject of~\cite{rebecq2019events}, that uses RNNs to obtain usable intensity video from events. The main advantage of these methods is the possibility of directly plugging-in existing algorithms on top of grayscale images. This comes at the cost of including pipeline buffering (latency) due to event collection over some time window, loosing the timestamp information, and potentially needing external IMU integration for long-term odometry.

Many techniques avoid the reconstruction of a full intensity image over a long buffer, but still rely on machine learning methods made for image data, such as Convolutional Neural Networks~\cite{fukushima1980neocognitron,lecun1998gradient}, and thus require formatting events into a sparse 2D grid structure. This has been applied to optical flow estimation~\cite{zhu_ev-flownet_2018, cannici_matrix-lstm_2020}, object detection~\cite{cannici_asynchronous_2018, cannici_matrix-lstm_2020}, and depth estimation~\cite{tulyakov2019learning}. Various aggregation schemes can be used, such as time-window binning or voxel volumes. Different grid sampling schemes are proposed in~\cite{gehrig_end--end_2019} and \cite{cannici_matrix-lstm_2020}. Advantages of these methods include compatibility with image-based learning algorithms, but disadvantages include, once again, inefficiency over sparse grids, loss of precise event timings, and a delay required to collect frames over time windows. 

A distinct approach, evaluated on image classification, samples events until they form a connected graph, with a combination of spatial and temporal distances as a measure of edge length~\cite{Bi2019Graph}. A neural network able to work on graph data~\cite{bronstein2017geometric} is then used to process the inputs. The use of spatial graph convolutions addresses the issue of sparsity found in grid-based approaches but still requires to collect data over a time window.

\paragraph{Per-event methods.}
Since event-cameras are considered a neuromorphic system, researchers theorized they would go hand-in-hand with a more biologically-grounded model for processing. Spiking Neural Networks (SNNs)~\cite{maass1997networks} are a class of neural networks based on human-vision perception principles, asynchronously activating specific neurons. This makes them a theoretical candidate for processing DVS events, one at a time~\cite{akolkar_2015, paulun_2018}. In their original form, SNNs are non-differentiable and thus incompatible with backpropagation-based training; therefore, most SNN methods require either proxy-based procedures~\cite{stromatias2017event} or an approximation of the original SNN formulation~\cite{lee2016training}. Nevertheless, these models tend to have lower performance than more modern methods.

Another clear choice for event-by-event classification are RNNs~\cite{elman1990finding}, neural networks specifically designed to handle sequential data. Such models, however, usually assume evenly-spaced series inputs, therefore neglecting one of the main features of DVS sensors. To address this, an extension of the LSTM~\cite{hochreiter1997long} architecture, named PhasedLSTM~\cite{neil_phased_2016}, was devised. This model added time gates to the previous and current intermediate hidden states. These gates open cyclically, modulated by the current input timestamp. PhasedLSTM was tested on event classification, using an embedding for the event coordinates, showing an improvement over LSTM for performance on the same task. Note that this is the closest existing method to our own.

\subsection{Neural ODEs}
NODEs are a recent methodology for modeling data as a dynamical system, governed by a neural network and solved using traditional ODE solvers \cite{Chen2018NeuralOD}. Inference is performed using gradient-based optimization through several time steps of the discretized ODE, typically using \emph{explicit} time-stepping schemes \cite{butcher_runge-kutta_1996}. To reduce memory requirements, researchers have proposed using the adjoint method \cite{Chen2018NeuralOD, Gholami2019}. NODEs have been applied to the time-series domain~\cite{rubanova_latent_2019}, by employing an LSTM to preprocess irregularly-spaced samples before feeding it into a NODE solver. This adds flexibility to the original formulation, at the cost of additional parameters and increased processing time. Moreover, there is high risk that the conditioning network could perform most of the inference and therefore the NODE results only in an integration task. In this work, we instead consider ODEs with an input connection, similarly to the SNODE architecture in \cite{Quaglino2020SNODE}.


\section{Input-filtering Neural ODE (INODE)} \label{sec_inode}
The proposed approach builds upon the architecture proposed in \cite{Quaglino2020SNODE}, with the difference that here we do not focus on the improvement of training efficiency and use standard back-propagation through time. We implement a batch Euler ODE solver so that our network can be dealt with as  an RNN. This allows for the state to be unmeasured (hidden), for instance like in LSTMs. The result is an recurrent architecture with skip connections that can handle unevenly-spaced points in time. We also add a decoder network as a classifier. 

\paragraph{Input-filtering Neural ODE.}
Consider the constrained differential optimization problem,
\begin{align}
	\min_{\theta_f, \theta_g \in \mathbb{R}^m} \int_{t_0}^{t_1} & L(z(t),\bar{z}(t)) ~ dt, \label{problem} \\
	\mathrm{s.t.} \quad h'(t) &= f(h(t), u(t);\theta_f), \nonumber \\
	z(t) &= g(h(t); \theta_g), \nonumber \\ 
	h(t_0) &= h_0, \nonumber 
\end{align}
where $h(t)$ is the hidden state, $u(t)$ is the input, $z(t)$ is the predicted output, $\bar{z}(t)$ is the desired output, the loss $L$ is given, $f$ and $g$ are neural networks with a fixed architecture defined by, respectively,  $\theta_f$, and $\theta_g$ which are parameters that have to be learned. The first two equality constraints in \eqref{problem} define an ODE. Problems of this form have been used to represent several inverse problems, for instance in machine learning, estimation, filtering and optimal control \cite{stengel_optimal_1994, law2015data, Ross2015}. Since this architecture can act as a general filter for the input signal, $u(t)$, we refer to it as the \emph{Input-filtering Neural ODE} ({INODE}). We consider this as a general framework for handling event data in a machine-learning scenario.

\paragraph{Application to DVS cameras.} We propose to use INODE to build a system that predicts (labels) online by filtering a live-stream of DVS-camera events. The aim is to learn the ODE in problem \eqref{problem}, given short excitation event sequences $u(t)$. Ideally, this model should produce the fastest trajectory from the initial state $h_0$ to an appropriate (unknown) state $\bar{h}$ such that $\bar{z}=g(\bar{h})$, where $g$ serves as a classification layer and $\bar{z}$ are the labels to be predicted. Hence, we fix the target to $\bar{z}(t) = \bar{z}$, $\:\forall t$. 

\paragraph{Event inputs}
Events are high-frequency signals, and solving a high-frequency ODE is difficult. Event streams are also extremely dense: the time between events is, in general, very small (often $< 100 \mu$s). We propose the use of a sample-and-hold approach, where events are held constant for up to a maximum delta-time $\text{d}_{\max}$. In the rare case that no events occur after $\text{d}_{\max}$, then we simply wait for the next event and hold the previous result without running the forward pass. 

\paragraph{Problem discretization.} 
A neuromorphic dataset $D$ is a collection $\mathbf{e}=\{ e_i \}^{M}_{i=0}$ of events $e_i=(x_i, y_i, p_i, t_i)$,
where $M$ is the number of events considered for a given sample (typically on the order of thousands), and labels $\bar{z} \in \{0, ...., C-1 \}$ for $C$ classes. A digit is represented by a tuple $(\mathbf{e}, \bar{z})$ and the dataset by $D=\{(\mathbf{e}, \bar{z})_n\}^{N}_{n=0}$, where $N$ is the number of samples. 
Thus, the integral in \eqref{problem} is discretized for each sample using a subset of size $S$ evaluation points $\left[ t_1, ..., t_S \right]$ as:
\begin{equation*}   
    \mathcal{L}(\mathbf{e}, \bar{z}) = \frac{1}{S} \sum^{S}_{i=1} L(z(t_i),  \bar{z}),
\end{equation*}
where $L$ is the cross-entropy loss. For each evaluation point, a new input event is used, i.e., $u(t_i) = (x(t_i), y(t_i), p(t_i))$. Finally, the sample loss is averaged over the dataset
$\mathcal{L}_D = \mathbb{E}_{(\mathbf{e}, \bar{z})} \left[ \mathcal{L}(\mathbf{e}, \bar{z}) \right]$ and used for optimization. 

\paragraph{Time step normalization.} To accurately use the time-steps $dt$, they can be normalized to values smaller than one (timestamps are recorded in microseconds and thus quickly reach very large values). At the same time, $dt$ should not be very small to avoid optimization issues, such as vanishing gradients.
We compute $dt$ from the raw time-steps and divide by the 98th quantile $d_q$ from the empirical distribution of $dt$ for each training dataset, pre-computed and fixed, with an upper threshold at 1. The normalized step is $d \tau = \tfrac{dt}{\text{d}_{q}}$. The complete training procedure is summarised in Algorithm \ref{alg:example}. 



\begin{algorithm}[tb]
   \caption{INODE}
   \label{alg:example}
\begin{algorithmic}
   \STATE {\bfseries Inputs:} $\mathbf{e}$, $\text{d}_{\max}$, $\text{d}_{q}$, $S \ll M$
   \REPEAT
   \STATE Sample $\{u_i, t_i \}_{i=s}^{s+S}$ from $\mathbf{e}$
   \FOR{$i=0$ {\bfseries to} $S-1$}
   \STATE $\text{d}\tau = \min ( (t_{i+1} - t_{i}) / \text{d}_{q}, \text{d}_{\max} )$
   \STATE $h(t_{i+i}) = h(t_{i}) + \text{d}\tau\ f(h(t_i), u(t_{i}))$ \\
   \STATE $z(t_{i+1}) = g(h(t_{i+1}))$ \\ 
   \STATE $L_{i+1} = L(z(t_{i+1}), \bar{z})$ \\
   \ENDFOR
   \STATE $\mathcal{L} = \frac{1}{S } \sum^{S}_{i=1} L_i$
   \STATE $\theta \leftarrow \nabla_{\theta} \mathcal{L}$
   \UNTIL{Convergence}
\end{algorithmic}
\end{algorithm}
\vspace{-12pt}
\section{Experiments}
We consider multiple classification tasks to validate our method, benchmarking against LSTM variants.
During these, we always learn from short event subsequences (up to 100 events).
Performance is evaluated with the same number of events used during training.
This allows for potential real-time classification (when properly optimized), as inference time increases with number of events processed.

\paragraph{Setup.}
We use the same configurations, architectures, and hyper-parameters for all of the datasets and model variants.
We train all models with different $\rho = \{1, 0.4, 0.2\}$ levels, where $\rho$ is 
the fraction of train dataset used for training.
 For each sequence, we sample a random offset and relative sub-sequence of length $S \ll M$. In all of the experiments we set $S=100$.
We then use such sub-sequence as input $u(t)$ for the model with batch size  $B_{\rho} = \rho B_{\rho=1} $.


At test time, we consider different scenarios: a standard case, where the models are evaluated with $S=100$ on the test set, and more challenging ones, in which they are evaluated with short sub-sequences in the range $S=\{10, 20, 30, ..., 100 \}$.

\paragraph{Baselines.}
We first compare INODE against LSTM and bidirectional LSTM (bi-LSTM). The
 LSTMs and bi-LSTMs receive the event time-step as additional input. 
We consider three bi-LSTM models with hidden states of dimension $\{ 36, 72, 128 \}$.
The $\text{bi-LSTM}_{72}$ has approximately the same capacity of INODE, while $\text{bi-LSTM}_{128}$ is 3x larger.

We also consider a variant of LSTM, the PhasedLSTM \cite{neil_phased_2016} without coordinate-grid embedding. 
This model explicitly handles asynchronous data learning an additional phase gate. 
Such approach is -- according to the authors -- fruitful for \emph{long} sequences (>1000 steps), in which the phase gate can exploit periodic mechanism in the data.
Given our use case, \emph{short} sequences of events (<100), we do not expect improvements over a standard LSTM.
To the best of our knowledge, this is the only known method which -- like ours -- inherently handles asynchronous timing within the model and does not need to learn an external transition model. Unfortunately, our initial results with standard PhasedLSTM were rather poor. However, combining phased and bidirectional LSTM seemed promising. We denote this as P-bi-LSTM.

The number of states, parameters, and input features for each model are summarized in Table \ref{table:parameters}. 
\begin{table}[ht!]
\caption{Models setup and complexity}
\label{sample-table}
\begin{center}
\begin{scriptsize}
\begin{sc}
\begin{tabular}{lcccc}
\toprule
    model & n states & n params & input \\
\midrule
   $\textbf{inode}_{30}$ (ours)     & 30    & 42,161   & $(x, y, p)$ \\ \midrule
  $\text{lstm}_{164}$     & 164   & 111,520 & $(x, y, p, t)$\\
$\text{p-lstm}_{164}$   & 164   & 111,192 & $(x, y, p)$   \\
$\text{lstm}_{104}$     & 104   & 45,760  & $(x, y, p, t)$\\
$\text{p-lstm}_{104}$   & 104   & 45,552  & $(x, y, p)$   \\
$\text{lstm}_{72}$      & 72    & 22,464  & $(x, y, p, t)$\\
$\text{p-lstm}_{72}$    & 72    & 22,320  & $(x, y, p)$   \vspace{2pt}  \\
$\text{bi-lstm}_{128}$     & 128   & 137,216    & $(x, y, p, t)$\\
$\text{p-bi-lstm}_{128}$   & 128   & 136,704    & $(x, y, p)$\\
$\text{bi-lstm}_{72}$      & 72    & 44,928     & $(x, y, p, t)$ \\
$\text{p-bi-lstm}_{72}$    & 72    & 44,640     & $(x, y, p)$ \\
$\text{bi-lstm}_{36}$      & 36    & 12,096     & $(x, y, p, t)$ \\
$\text{p-bi-lstm}_{36}$    & 36    & 11,952     & $(x, y, p)$ \\
\bottomrule
\end{tabular}
\end{sc}
\end{scriptsize}
\end{center}
\vskip -0.13in
\label{table:parameters}
\end{table}

\paragraph{Datasets.}
We consider three neuromorphic datasets:
\subparagraph{1) NMNIST}
The NMNIST dataset \cite{orchard2015converting} is a neuromorphic version of MNIST. 
It is an artificial dataset, generated by moving a DVS sensor in front of an LCD monitor displaying static images.
It consists of 60k training samples and 10k test samples, for 10 different digits on a grid of 34 $\times$ 34 pixels.
We consider only the first 2,000 (of potentially up to 6,000) events for each sequence.
We do not stabilize the events spatially nor attempt to remove noisy events, which are options available in the dataset.
\subparagraph{2) ASL (12-16k)}
The ASL-DVS dataset, is a neuromorphic dataset, obtained for a stream of real-world events \cite{Bi2019Graph}.
It consists of around 100k samples for 24 different letters from the American Sign Language, with spatial resolution 180 $\times$ 240.
Its sequences range from 1-500k events, with length distribution peaking in the 12-16k range. To avoid inconsistencies, we consider a subset containing only samples with a number of events between 12k and 16k. 
The resulting dataset contains 12,275 training samples plus 1,364 test samples.
\subparagraph{3) NCALTECH}
Similarly, the NCALTECH dataset \cite{orchard2015converting} is the neuromorphic version of CALTECH101, produced in the same fashion as NMNIST.
It consists of 100 heavily unbalanced classes of objects plus a background, with spatial resolution 172 $\times$ 232.
The dataset contains 6,634 training samples and 1,608 test samples, after removing the background images.
As with NMNIST, we again avoid stabilizing/denoising the images.

\paragraph{Solver.}
We train each model using ADAM for 300 epochs, with $S=100$ and learning rate of 1e-3.
The batch size $B_{\rho=1}$ is $1000$ for NMNIST, and $100$ for the other datasets. We consider a simple multi-layer perceptron for $f$:
$$f(x,u) = \text{FC}_{3}(\sigma(\text{FC}_{2}(\sigma(\{ \text{FC}_{1}(x), \text{FC}_{u}(u)\})),$$
where $\{ \cdot,\cdot \}$ denotes the concatenation operation, FC is a fully-connected layer, and $\sigma = \tanh$ is the activation. 

\begin{table}[h!]
\caption{$f$ parameterization for INODE and classifier.}
\label{sample-table}
\vskip 0.15in
\begin{center}
\begin{scriptsize}
\begin{sc}
\begin{tabular}{lcccc|c}
\toprule
    &  $\text{FC}_{1}$ & $\text{FC}_{u}$ & $\text{FC}_{2}$ & $\text{FC}_{3}$ & $\text{FC}_{c}$\\
\midrule
input dim           & 30   & 3(+1)  & 256 & 128 & 30 \\
output dim          & 128  & 128    & 128 & 30  & \text{n classes} \\
\bottomrule
\end{tabular}
\end{sc}
\end{scriptsize}
\end{center}
\vskip -0.1in
\end{table}

\begin{table}[h]
\caption{Classification accuracy on NMNIST test set increasing the number of events (10 classes).}
\begin{center}
\begin{scriptsize}
\begin{sc}
\begin{tabular}{lccccc|c}
\toprule
model                        & dataset \% & \multicolumn{5}{c}{n events test} \\
\midrule
                            &         & 10 & 20 & 30 & 40 & 100\\
\midrule
$\textbf{inode}_{30}$  & 100  & {0.48} & {0.66} & 0.75 & 0.80          & 0.89         \vspace{2pt} \\
$\text{lstm}_{164}$        & 100     & \textbf{0.63} & \textbf{0.80} & \textbf{0.86} & \textbf{0.89} & \textbf{0.94}        \\ 
$\text{lstm}_{104}$        & 100     & 0.55          & 0.71          & 0.78          & 0.81          & 0.88          \\
$\text{p-lstm}_{104}$      & 100     & 0.27          & 0.29          & 0.27          & 0.24          & 0.18          \\
$\text{lstm}_{72}$         & 100     & 0.46          & 0.61          & 0.68          & 0.73          & 0.81        \vspace{2pt}  \\ 
$\text{bi-lstm}_{128}$         & 100 & 0.39          & {0.66} & {0.77} & {0.84} & {0.93} \\ 
$\text{p-bi-lstm}_{128}$       & 100 & 0.28          & 0.34          & 0.39          & 0.44          & 0.55          \\ 
$\text{bi-lstm}_{72}$          & 100 & 0.31          & 0.50          & 0.62          & 0.70          & 0.84          \\ 
$\text{p-bi-lstm}_{72}$        & 100 & 0.26          & 0.32          & 0.36          & 0.40          & 0.51          \\  
$\text{bi-lstm}_{36}$          & 100 & 0.22          & 0.34          & 0.43          & 0.48          & 0.61          \\ 
$\text{p-bi-lstm}_{36}$        & 100 & 0.24          & 0.30          & 0.32          & 0.35          & 0.44          \\
\midrule
$\textbf{inode}_{30}$  & 40  & {0.46} & {0.65} & {0.73} & {0.79}  & 0.88      \vspace{2pt} \\
$\text{lstm}_{164}$        & 40      & \textbf{0.61} & \textbf{0.79} & \textbf{0.85} & \textbf{0.88}  & \textbf{0.93}        \\ 
$\text{lstm}_{104}$        & 40      & 0.46          & 0.62          & 0.69          & 0.73           & 0.80        \\ 
$\text{p-lstm}_{104}$      & 40      & 0.24          & 0.26          & 0.24          & 0.21           & 0.15        \\
$\text{lstm}_{72}$         & 40      & 0.44          & 0.59          & 0.65          & 0.70           & 0.78       \vspace{2pt} \\ 
$\text{bi-lstm}_{128}$         & 40  & 0.30          & 0.53          & 0.68          & 0.77           & {0.89} \\ 
$\text{p-bi-lstm}_{128}$       & 40  & 0.27          & 0.33          & 0.37          & 0.41           & 0.51          \\  
$\text{bi-lstm}_{72}$          & 40  & 0.27          & 0.39          & 0.49          & 0.56           & 0.72          \\ 
$\text{p-bi-lstm}_{72}$        & 40  & 0.25          & 0.30          & 0.34          & 0.37           & 0.45          \\  
$\text{bi-lstm}_{36}$          & 40  & 0.25          & 0.36          & 0.42          & 0.47           & 0.58          \\ 
$\text{p-bi-lstm}_{36}$        & 40  & 0.23          & 0.27          & 0.30          & 0.32           & 0.40          \\  
\midrule
$\textbf{inode}_{30}$    & 20  & \textbf{0.46} & 0.63          & 0.73          & 0.78          & 0.87               \vspace{2pt} \\
$\text{lstm}_{164}$        & 20      & \textbf{0.46} & 0.62          & 0.68          & 0.72          & 0.79\\
$\text{lstm}_{104}$        & 20      & 0.29          & 0.36          & 0.41          & 0.43          & 0.49      \\
$\text{p-lstm}_{104}$      & 20      & 0.22          & 0.25          & 0.23          & 0.20          & 0.17      \\
$\text{lstm}_{72}$         & 20      & 0.26          & 0.33          & 0.36          & 0.39          & 0.42     \vspace{2pt} \\
$\text{bi-lstm}_{128}$    & 20  & 0.42          & \textbf{0.64} & \textbf{0.75} & \textbf{0.80} & \textbf{0.90}       \\ 
$\text{p-bi-lstm}_{128}$  & 20  & 0.25          & 0.30          & 0.34          & 0.37          & 0.47                \\  
$\text{bi-lstm}_{72}$     & 20  & 0.30          & 0.47          & 0.58          & 0.64          & 0.77                \\ 
$\text{p-bi-lstm}_{72}$   & 20  & 0.23          & 0.28          & 0.30          & 0.33          & 0.41                \\ 
$\text{bi-lstm}_{36}$     & 20  & 0.21          & 0.30          & 0.36          & 0.40          & 0.49                \\
$\text{p-bi-lstm}_{36}$   & 20  & 0.21          & 0.24          & 0.26          & 0.28          & 0.34                \\ 
\midrule
random &   & 0.10 & 0.10 & 0.10 & 0.10 & 0.10\\
\bottomrule
\end{tabular}
\end{sc}
\end{scriptsize}
\end{center}
\vskip -0.1in
\label{table:short_nmnist}
\end{table}

\paragraph{Results.} 
When testing the models, we vary both the size of the \emph{training} dataset and the number of \emph{test} events used for the classification ($10 \leq S \leq 100$). The former is used to show INODE's learning efficiency when using a small amount of training data, while the latter demonstrates INODE's real-time scenario usability. Tables \ref{table:short_nmnist}, \ref{table:short_asl}, and \ref{table:short_ncaltech} report accuracies for each of our datasets.

The LSTM with 164 states outperforms the proposed architecture on NMNIST, see Table \ref{table:short_ncaltech}. On the ASL dataset (Table \ref{table:short_asl}) our approach consistently outperforms all of the unidirectional baselines with a margin of 20\%. We believe this is important since, among the considered datasets, ASL contains by far the most realistic data, being the only one not generated from static images. For NCALTECH, our approach is either on par or better than the LSTM  when a small percentage of event is used (Table \ref{table:short_nmnist}).

For the bidirectional baselines, with approximately the same capacity ($\text{INODE}_{30}$ and $\text{bi-LSTM}_{72}$), INODE performs better then the bi-LSTMs on all of the datasets. 
Increasing the baseline capacity ($\text{bi-LSTM}_{128}$), INODE performs better on NCALTECH and ASL, while slightly losing its edge to the $\text{bi-LSTM}_{128}$ on NMNIST.
Decreasing the training-set size has essentially no impact on NMNIST for all models -- confirmation of a relatively simple dataset.

One can also notice that, with a couple of exceptions on NMNIST, INODE outperforms the bidirectional methods regardless of number of input events. These are as low as $S=10$ and in principle even $S=1$ is possible without modifying our approach. Interestingly, with only 10 events, the model can correctly classify NMNIST digits about half of the time.
As such, we demonstrate INODE's ability to extract information in the case of exceptional sparsity and data unavailability. 
This could be extremely important in scenarios such as collision avoidance and human-machine interaction, where safety is a paramount requisite.

Finally, Figure \ref{fig:results_summary_bi}, \ref{fig:results_summary} and more comprehensive figures found in the Appendix further illustrate how INODE trains faster using fewer samples and events, especially on the ASL dataset. 

\begin{table}[ht!]
\caption{Classification accuracy on ASL test set increasing the number of events (24 classes).}
\begin{center}
\begin{scriptsize}
\begin{sc}
\begin{tabular}{lccccc|c}
\toprule
model                       & dataset \% & \multicolumn{5}{c}{n events test} \\
\midrule
                           &         & 10            & 20           & 30           & 40            & 100\\
\midrule
$\textbf{inode}_{30}$ & 100            & \textbf{0.37} & \textbf{0.51}& \textbf{0.61}& \textbf{0.67} & \textbf{0.79}      \vspace{2pt} \\
$\text{lstm}_{164}$        & 100     & 0.35          & 0.44         & 0.51         & 0.55          & 0.59         \\ 
$\text{p-lstm}_{164}$      & 100     & 0.22          & 0.25         & 0.25         & 0.24          & 0.21         \\ 
$\text{lstm}_{104}$        & 100     & 0.27          & 0.31         & 0.32         & 0.34          & 0.37         \\ 
$\text{p-lstm}_{104}$      & 100     & 0.21          & 0.21         & 0.23         & 0.22          & 0.20         \\ 
$\text{lstm}_{72}$         & 100     & 0.27          & 0.30         & 0.33         & 0.32          & 0.35         \\ 
$\text{p-lstm}_{72}$       & 100     & 0.24          & 0.26         & 0.27         & 0.28          & 0.24       \vspace{2pt}   \\
$\text{bi-lstm}_{128}$        & 100     & 0.18          & 0.26         & 0.35         & 0.40          & 0.54         \\ 
$\text{p-bi-lstm}_{128}$      & 100     & 0.28          & 0.33         & 0.36         & 0.41          & 0.47         \\ 
$\text{bi-lstm}_{72}$         & 100     & 0.25          & 0.36         & 0.43         & 0.49          & 0.61         \\ 
$\text{p-bi-lstm}_{72}$       & 100     & 0.29          & 0.32         & 0.36         & 0.38          & 0.45         \\ 
$\text{bi-lstm}_{36}$         & 100     & 0.17          & 0.25         & 0.29         & 0.32          & 0.38         \\ 
$\text{p-bi-lstm}_{36}$       & 100     & 0.23          & 0.27         & 0.30         & 0.31          & 0.36         \\
\midrule
$\textbf{inode}_{30}$ & 40     & \textbf{0.36} & \textbf{0.50}& \textbf{0.58}     & \textbf{0.64} & \textbf{0.69}       \vspace{2pt} \\
$\text{lstm}_{164}$        & 40      & 0.32          & 0.39         & 0.44         & 0.47               & 0.46          \\ 
$\text{p-lstm}_{164}$      & 40      & 0.19          & 0.19         & 0.19         & 0.19               & 0.18          \\
$\text{lstm}_{104}$        & 40      & 0.28          & 0.32         & 0.34         & 0.36               & 0.39          \\ 
$\text{p-lstm}_{104}$      & 40      & 0.18          & 0.19         & 0.20         & 0.19               & 0.21          \\
$\text{lstm}_{72}$         & 40      & 0.26          & 0.29         & 0.29         & 0.30               & 0.31          \\ 
$\text{p-lstm}_{72}$       & 40      & 0.24          & 0.26         & 0.25         & 0.27               & 0.25        \vspace{2pt}   \\ 
$\text{bi-lstm}_{128}$        & 40      & 0.29          & 0.40         & 0.48         & 0.55               & 0.65          \\ 
$\text{p-bi-lstm}_{128}$      & 40      & 0.27          & 0.32         & 0.35         & 0.37               & 0.41          \\
$\text{bi-lstm}_{72}$         & 40      & 0.23          & 0.26         & 0.31         & 0.35               & 0.40          \\ 
$\text{p-bi-lstm}_{72}$       & 40      & 0.23          & 0.28         & 0.30         & 0.33               & 0.36          \\
$\text{bi-lstm}_{36}$         & 40      & 0.19          & 0.22         & 0.24         & 0.28               & 0.34          \\ 
$\text{p-bi-lstm}_{36}$       & 40      & 0.23          & 0.27         & 0.30         & 0.31               & 0.35          \\ 
\midrule
$\textbf{inode}_{30}$ & 20      & \textbf{0.32}          & \textbf{0.47}& \textbf{0.55} & \textbf{0.60} & \textbf{0.71}     \vspace{2pt} \\
$\text{lstm}_{164}$        & 20      & 0.25          & 0.29         & 0.31         & 0.31                    & 0.31         \\
$\text{p-lstm}_{164}$      & 20      & 0.17          & 0.17         & 0.16         & 0.16                    & 0.15         \\
$\text{lstm}_{104}$        & 20      & 0.26          & 0.29         & 0.31         & 0.32                    & 0.33         \\
$\text{p-lstm}_{104}$      & 20      & 0.19          & 0.18         & 0.19         & 0.19                    & 0.17         \\
$\text{lstm}_{72}$         & 20      & 0.26          & 0.32         & 0.34         & 0.33                    & 0.35         \\ 
$\text{p-lstm}_{72}$       & 20      & 0.19          & 0.19         & 0.16         & 0.17                    & 0.18      \vspace{2pt}    \\ 
$\text{bi-lstm}_{128}$        & 20      & 0.28          & 0.37         & 0.43         & 0.48                    & 0.55          \\
$\text{p-bi-lstm}_{128}$      & 20      & 0.25          & 0.28         & 0.30         & 0.34                    & 0.37          \\
$\text{bi-lstm}_{72}$         & 20      & 0.20          & 0.26         & 0.32         & 0.34                    & 0.39          \\
$\text{p-bi-lstm}_{72}$       & 20      & 0.24          & 0.28         & 0.30         & 0.30                    & 0.35          \\
$\text{bi-lstm}_{36}$         & 20      & 0.21          & 0.26         & 0.28         & 0.30                    & 0.33          \\ 
$\text{p-bi-lstm}_{36}$       & 20      & 0.23          & 0.26         & 0.27         & 0.28                    & 0.31          \\ 
\midrule
random &   & 0.04 & 0.04 & 0.04 & 0.04 & 0.04\\
\bottomrule
\end{tabular}
\end{sc}
\end{scriptsize}
\end{center}
\vskip -0.1in
\label{table:short_asl}
\end{table}

\begin{table}[ht]
\caption{Classification accuracy on NCALTECH test set increasing the number of events (100 classes).}
\begin{center}
\begin{scriptsize}
\begin{sc}
\begin{tabular}{lccccc|c}
\toprule
model                        & dataset \% & \multicolumn{5}{c}{n events test} \\
\midrule
      &                     & 10                  & 20            &  30           & 40            & 100            \\
\midrule
$\textbf{inode}_{30}$  & 100 & {0.22} & {0.26} & {0.29} & 0.30          & 0.34          \vspace{2pt} \\ 
$\text{lstm}_{164}$        & 100     & \textbf{0.25} & \textbf{0.29} & \textbf{0.32} & \textbf{0.32} & \textbf{0.36}          \\ 
$\text{p-lstm}_{164}$      & 100     & 0.22          & 0.24          & 0.24          & 0.24          & 0.21                   \\ 
$\text{lstm}_{104}$        & 100     & 0.24          & 0.27          & 0.28          & 0.29          & 0.31                   \\ 
$\text{p-lstm}_{104}$      & 100     & 0.23          & 0.25          & 0.25          & 0.24          & 0.21                   \\ 
$\text{lstm}_{72}$         & 100     & 0.24          & 0.27          & 0.29          & 0.30          & 0.31                   \\ 
$\text{p-lstm}_{72}$       & 100     & 0.23          & 0.24          & 0.23          & 0.24          & 0.24                 \vspace{2pt}   \\
$\text{bi-lstm}_{128}$         & 100 & 0.16          & 0.24          & 0.28 & {0.31} & {0.35}  \\ 
$\text{p-bi-lstm}_{128}$       & 100 & 0.21          & 0.24          & 0.26          & 0.26          & 0.28           \\ 
$\text{bi-lstm}_{72}$          & 100 & 0.16          & 0.24          & 0.28          & 0.29          & 0.30           \\ 
$\text{p-bi-lstm}_{72}$        & 100 & 0.21          & 0.24          & 0.25          & 0.26          & 0.28           \\
$\text{bi-lstm}_{36}$          & 100 & 0.12          & 0.21          & 0.24          & 0.27          & 0.28           \\
$\text{p-bi-lstm}_{36}$        & 100 & 0.21          & 0.23          & 0.24          & 0.25          & 0.26           \\ 
\midrule
$\textbf{inode}_{30}$  & 40  & {0.23} & {0.27} & {0.29} & \textbf{0.31}  & \textbf{0.34}      \vspace{2pt} \\
$\text{lstm}_{164}$        & 40      & 0.25          & 0.27          & \textbf{0.30} & \textbf{0.31}  & 0.33          \\ 
$\text{p-lstm}_{164}$      & 40      & 0.22          & 0.23          & 0.22          & 0.22           & 0.20          \\
$\text{lstm}_{104}$        & 40      & \textbf{0.26} & \textbf{0.28} & 0.29          & 0.30           & 0.30          \\ 
$\text{p-lstm}_{104}$      & 40      & 0.21          & 0.22          & 0.22          & 0.22           & 0.22          \\
$\text{lstm}_{72}$         & 40      & 0.25          & 0.27          & 0.28          & 0.29           & 0.31          \\ 
$\text{p-lstm}_{72}$       & 40      & 0.20          & 0.21          & 0.20          & 0.21           & 0.20        \vspace{2pt}   \\ 
$\text{bi-lstm}_{128}$         & 40  & 0.17          & 0.22          & 0.24          & 0.25           & 0.29          \\ 
$\text{p-bi-lstm}_{128}$       & 40  & 0.22          & 0.24          & 0.26          & 0.26           & 0.28          \\  
$\text{bi-lstm}_{72}$          & 40  & 0.20          & 0.24          & 0.25          & 0.27           & 0.28          \\ 
$\text{p-bi-lstm}_{72}$        & 40  & 0.21          & 0.23          & 0.24          & 0.25           & 0.27          \\ 
$\text{bi-lstm}_{36}$          & 40  & 0.18          & 0.21          & 0.23          & 0.24           & 0.25          \\ 
$\text{p-bi-lstm}_{36}$        & 40  & 0.20          & 0.21          & 0.22          & 0.23           & 0.25          \\ 
\midrule
$\textbf{inode}_{30}$  & 20  & {0.22} & \textbf{0.25} & {0.26} & \textbf{0.28} & \textbf{0.30}      \vspace{2pt} \\
$\text{lstm}_{164}$        & 20      & \textbf{0.24} & \textbf{0.25} & 0.26          & 0.27          & \textbf{0.30}\\
$\text{p-lstm}_{164}$      & 20      & 0.21          & 0.22          & 0.22          & 0.22          & 0.20         \\
$\text{lstm}_{104}$        & 20      & 0.22          & 0.24          & 0.25          & 0.25          & 0.27         \\
$\text{p-lstm}_{104}$      & 20      & 0.20          & 0.23          & 0.22          & 0.23          & 0.21         \\
$\text{lstm}_{72}$         & 20      & 0.23          & 0.25          & \textbf{0.27} & 0.27          & 0.28         \\ 
$\text{p-lstm}_{72}$       & 20      & 0.19          & 0.20          & 0.20          & 0.20          & 0.20        \vspace{2pt}  \\ 
$\text{bi-lstm}_{128}$         & 20  & 0.19          & 0.23          & 0.25          & 0.26          & 0.28           \\ 
$\text{p-bi-lstm}_{128}$       & 20  & 0.21          & 0.23          & 0.25          & 0.26          & 0.26           \\ 
$\text{bi-lstm}_{72}$          & 20  & 0.11          & 0.15          & 0.20          & 0.22          & 0.25           \\ 
$\text{p-bi-lstm}_{72}$        & 20  & 0.20          & 0.21          & 0.23          & 0.24          & 0.25           \\ 
$\text{bi-lstm}_{36}$          & 20  & 0.17          & 0.18          & 0.21          & 0.20          & 0.22           \\
$\text{p-bi-lstm}_{36}$        & 20  & 0.20          & 0.20          & 0.23          & 0.23          & 0.24           \\ 
\midrule
random &   & 0.01 & 0.01 & 0.01 & 0.01 & 0.01\\
\bottomrule
\end{tabular}
\end{sc}
\end{scriptsize}
\end{center}
\vskip -0.1in
\label{table:short_ncaltech}
\end{table}

\begin{figure*}[h!t]
    \captionsetup[subfigure]{justification=centering}
    \centering
    \includegraphics[width=0.4\linewidth]{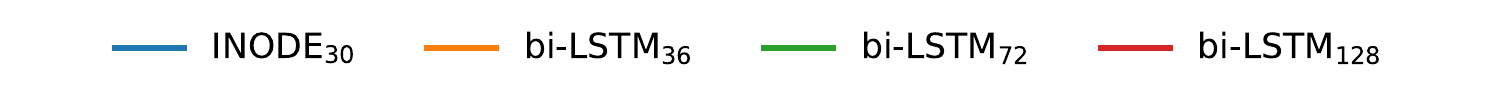}\\\vspace{5pt}
    \begin{subfigure}[b]{0.31\textwidth}
        \includegraphics[height=3.6cm, width=1.78cm]{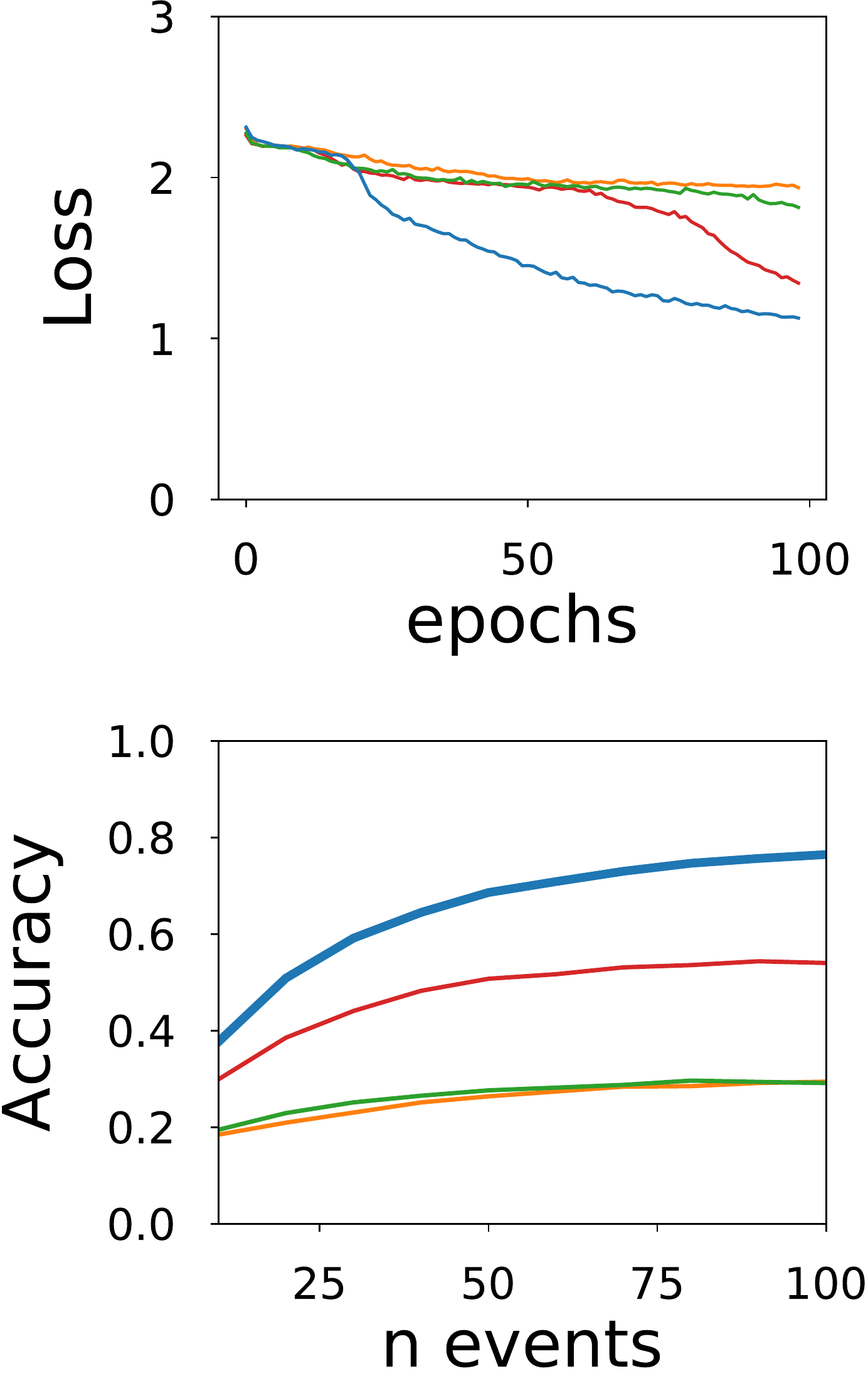}
        \includegraphics[height=3.6cm, width=1.36cm]{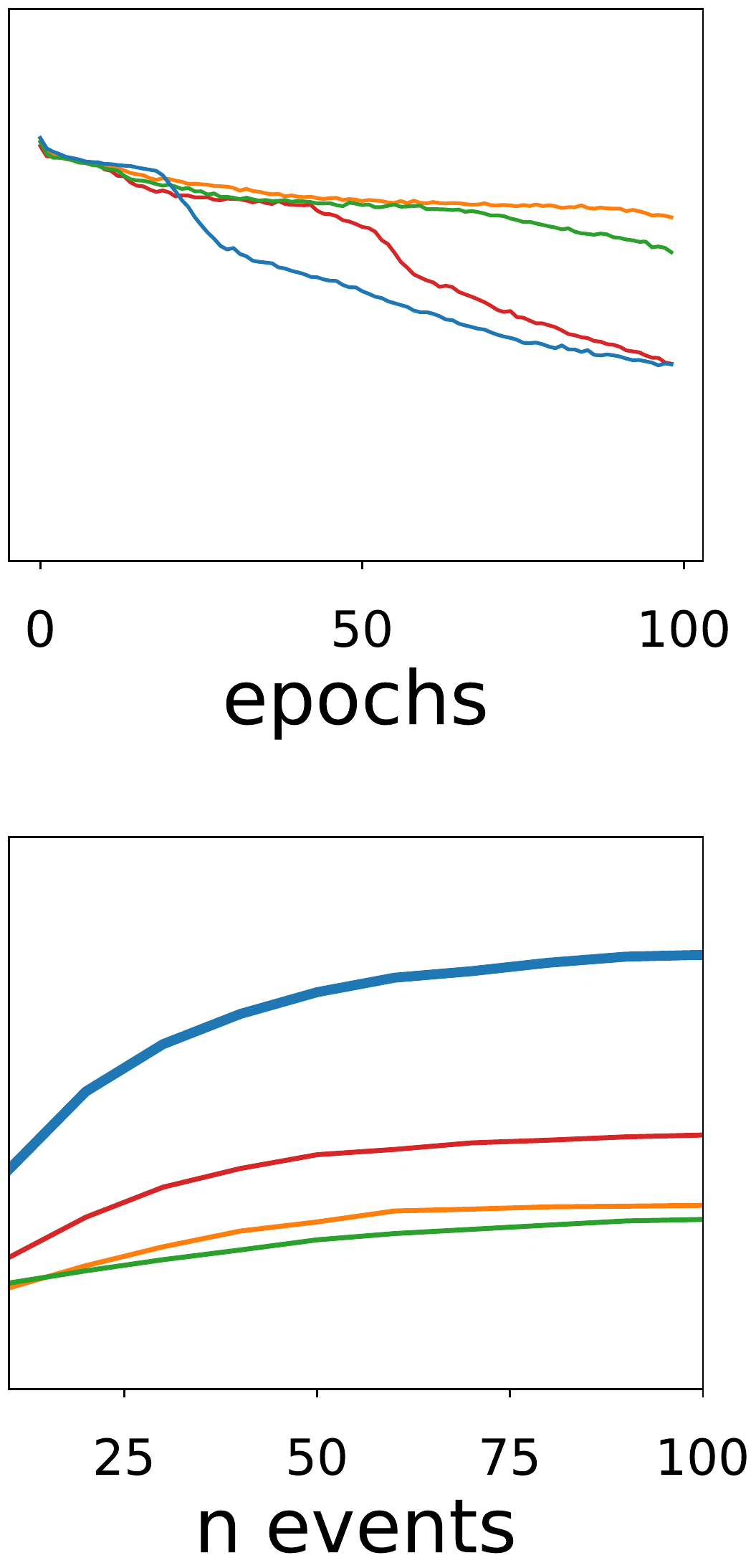}
        \includegraphics[height=3.6cm, width=1.36cm]{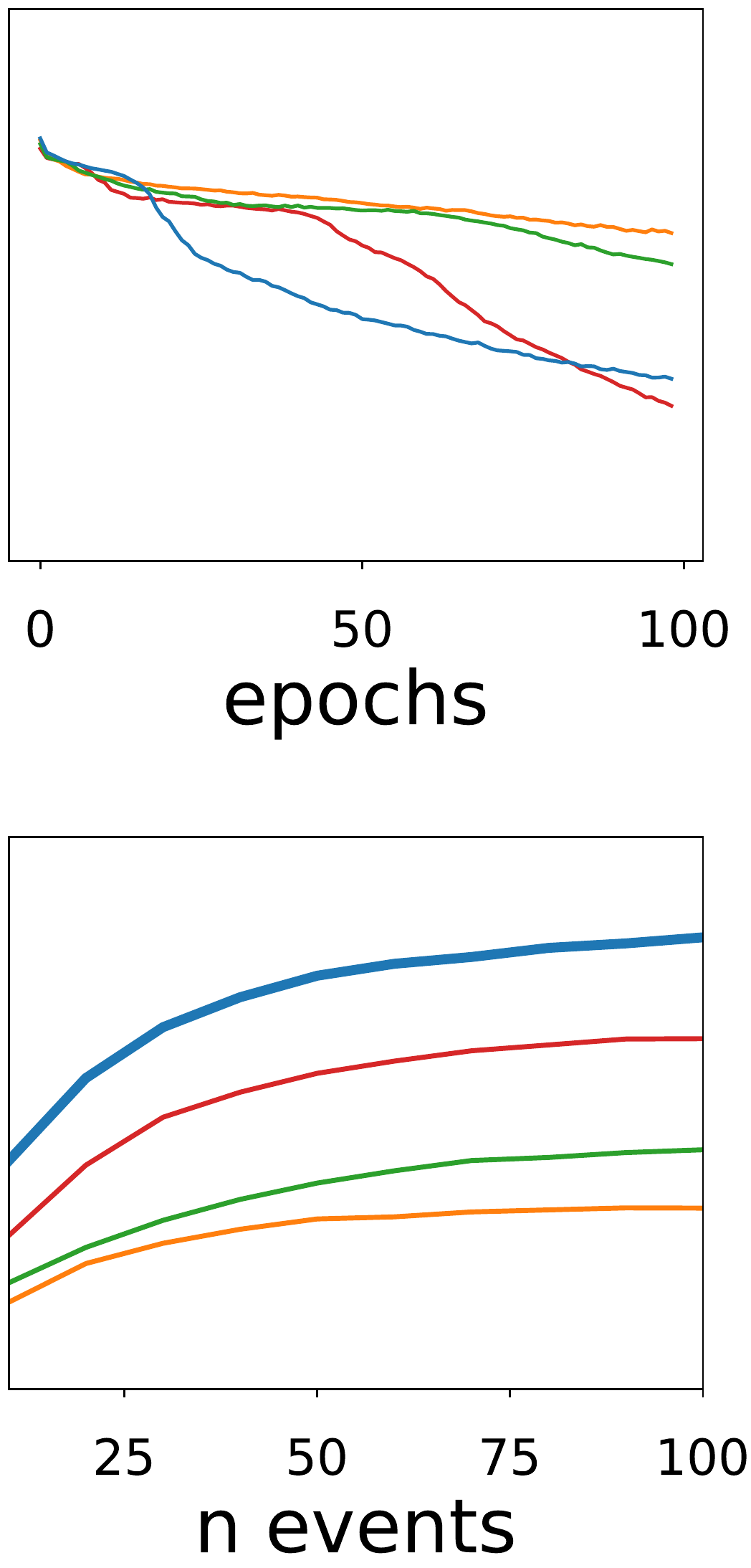}
        \caption{MNIST (100 epochs)}
        \vspace{5pt}
    \end{subfigure}
    \begin{subfigure}[b]{0.31\textwidth}
        \includegraphics[height=3.6cm, width=1.78cm]{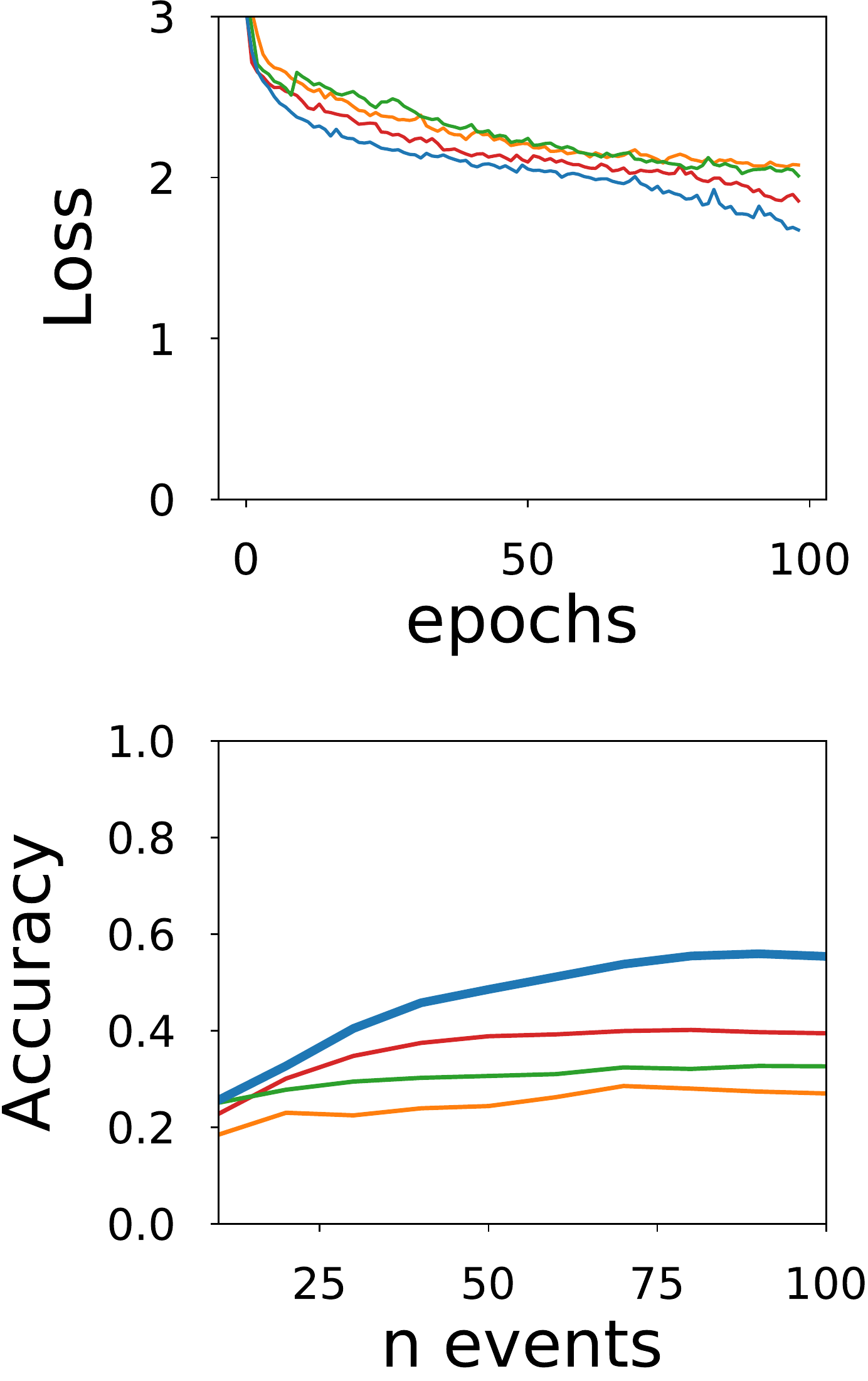}
        \includegraphics[height=3.6cm, width=1.36cm]{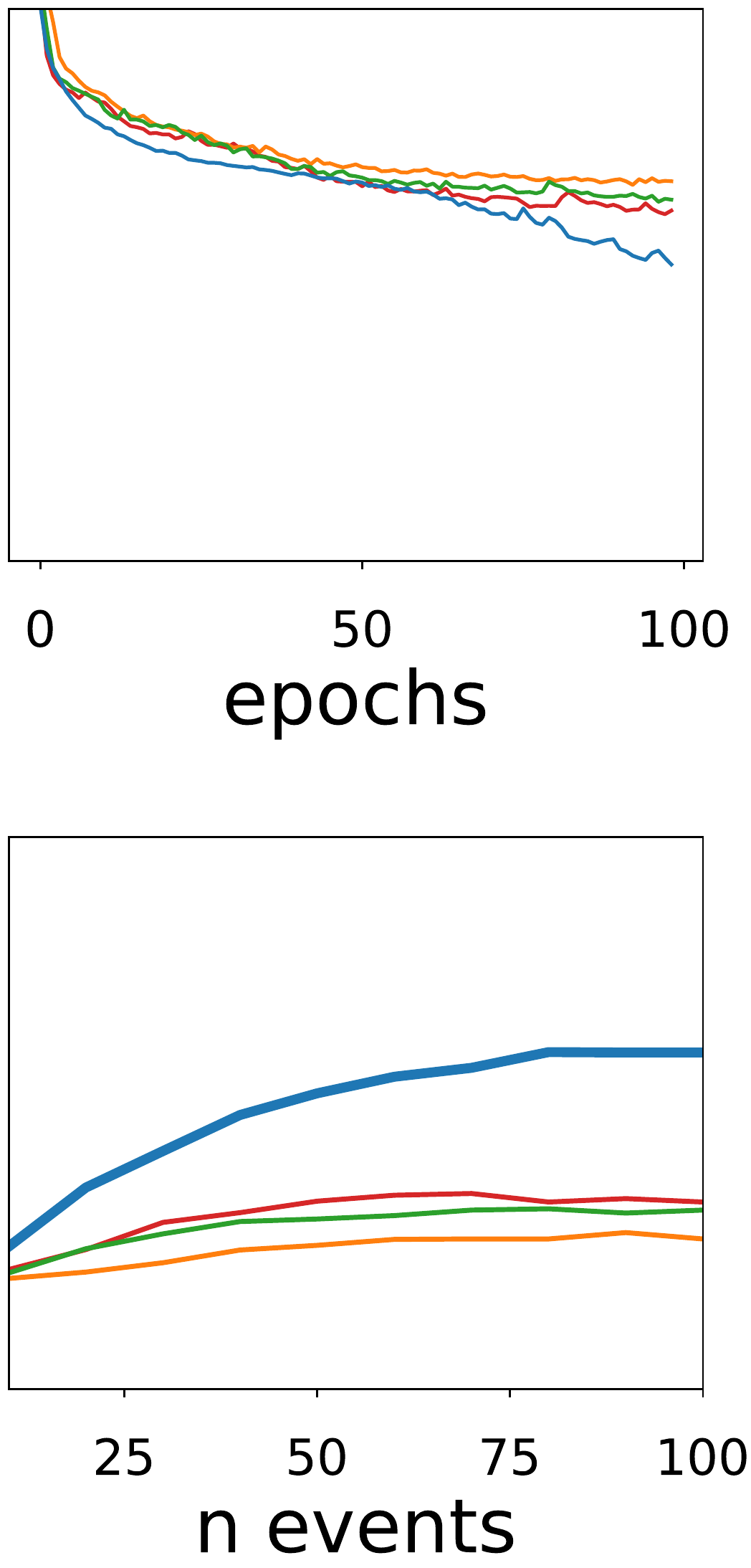}
        \includegraphics[height=3.6cm, width=1.36cm]{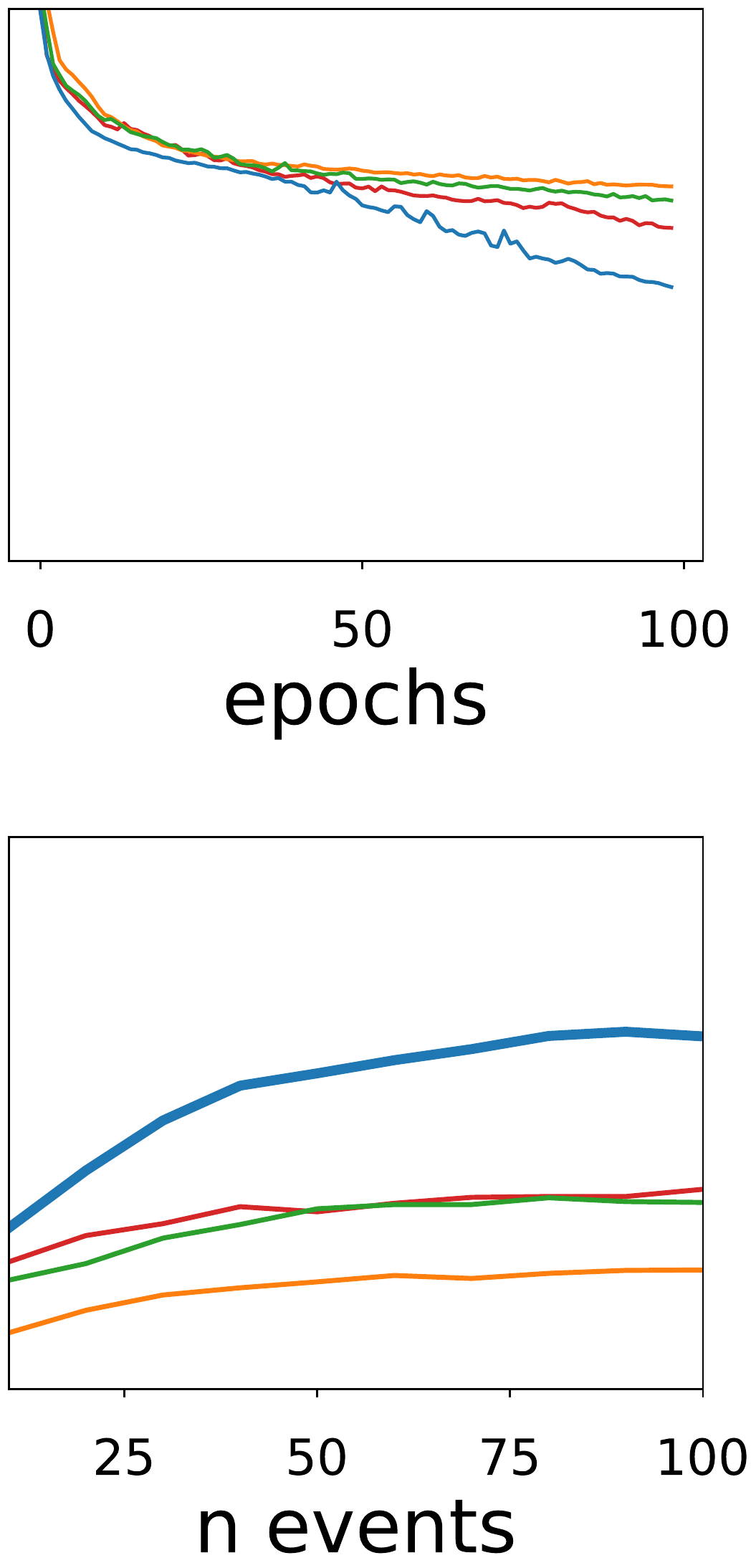}
        \caption{ASL (100 epochs)}
        \vspace{5pt}
    \end{subfigure}
    \begin{subfigure}[b]{0.31\textwidth}
        \includegraphics[height=3.6cm, width=1.78cm]{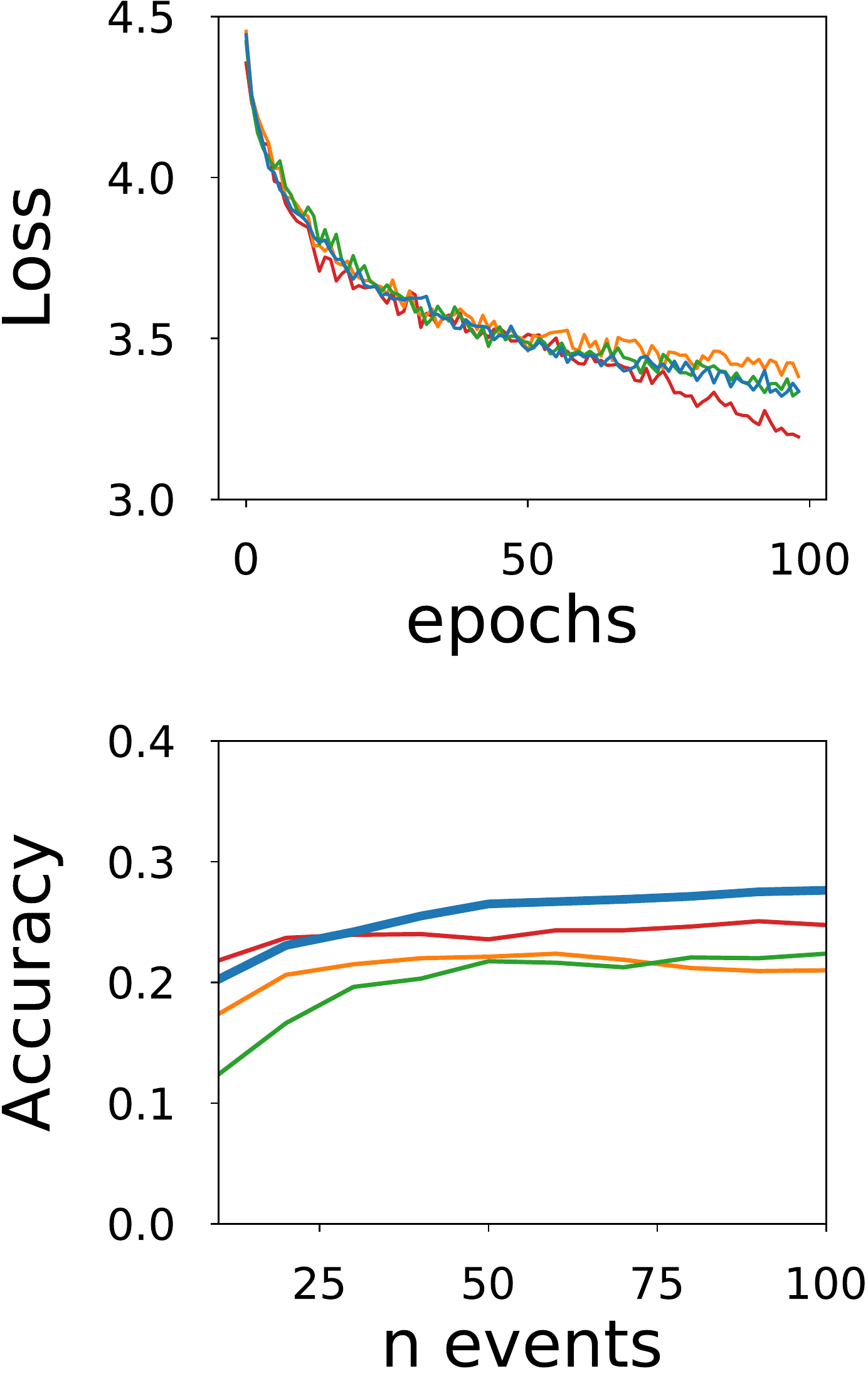}
        \includegraphics[height=3.6cm, width=1.36cm]{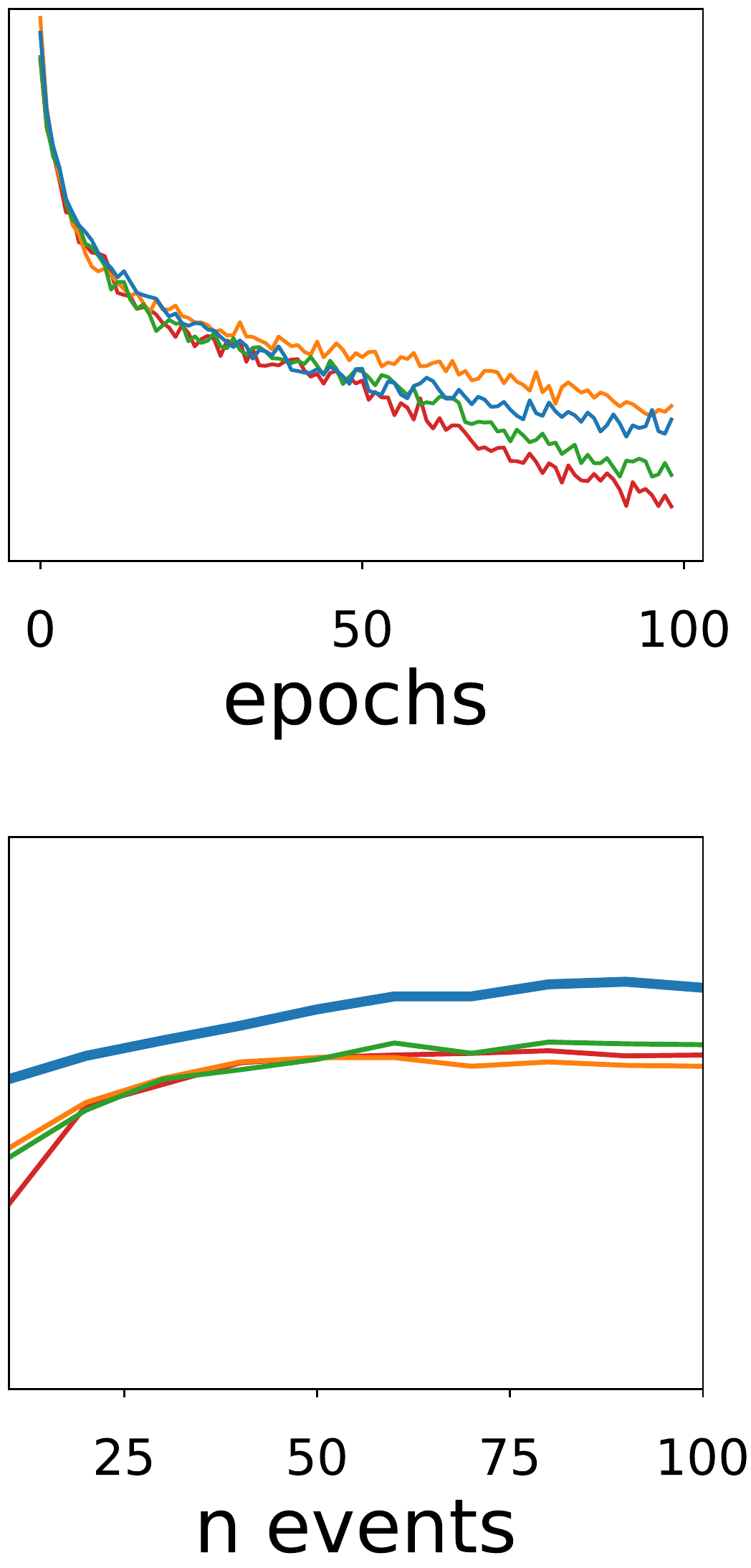}
        \includegraphics[height=3.6cm, width=1.36cm]{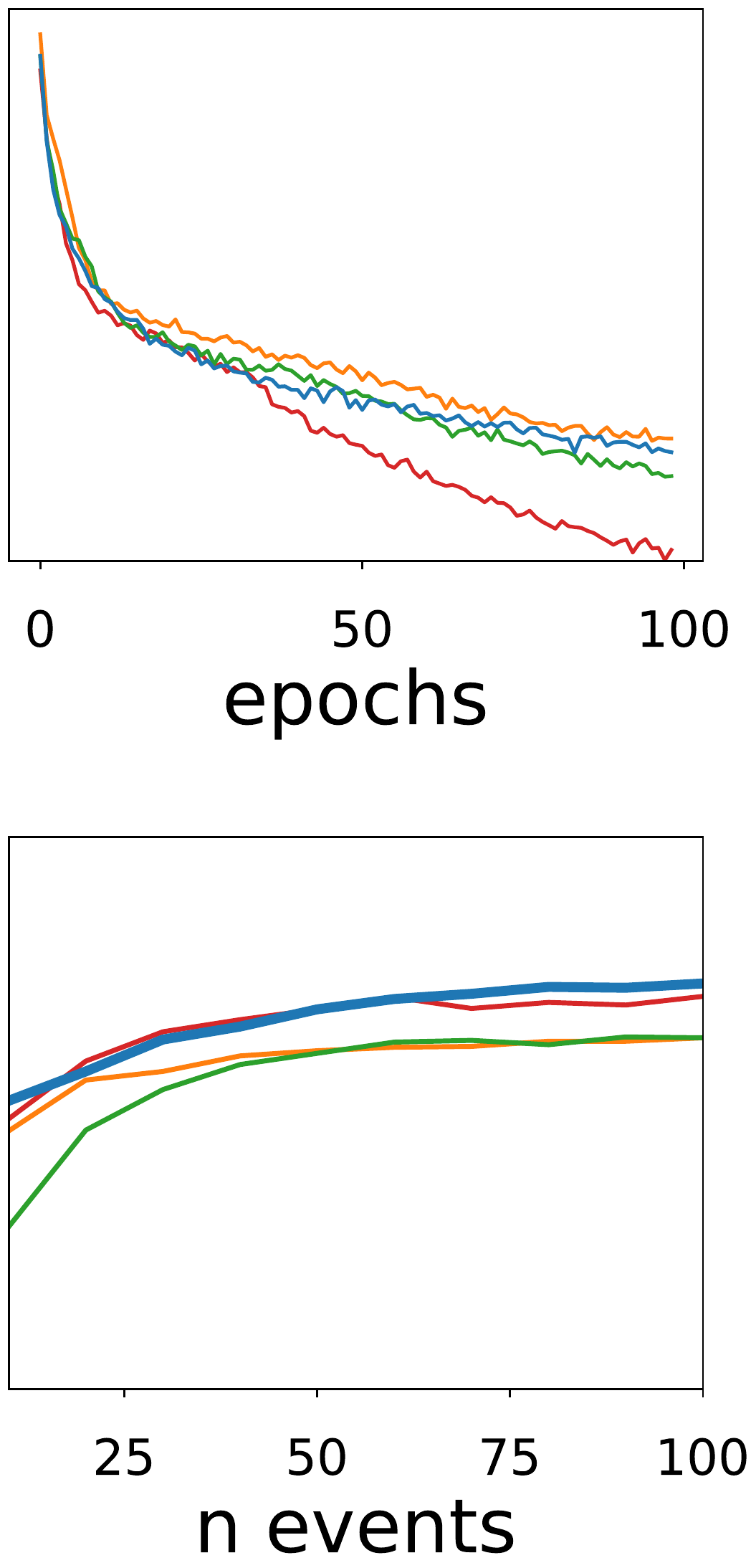}
        \caption{NCALTECH (100 epochs)}
        \vspace{5pt}
    \end{subfigure}
    \begin{subfigure}[b]{0.31\textwidth}
        \includegraphics[height=3.6cm, width=1.78cm]{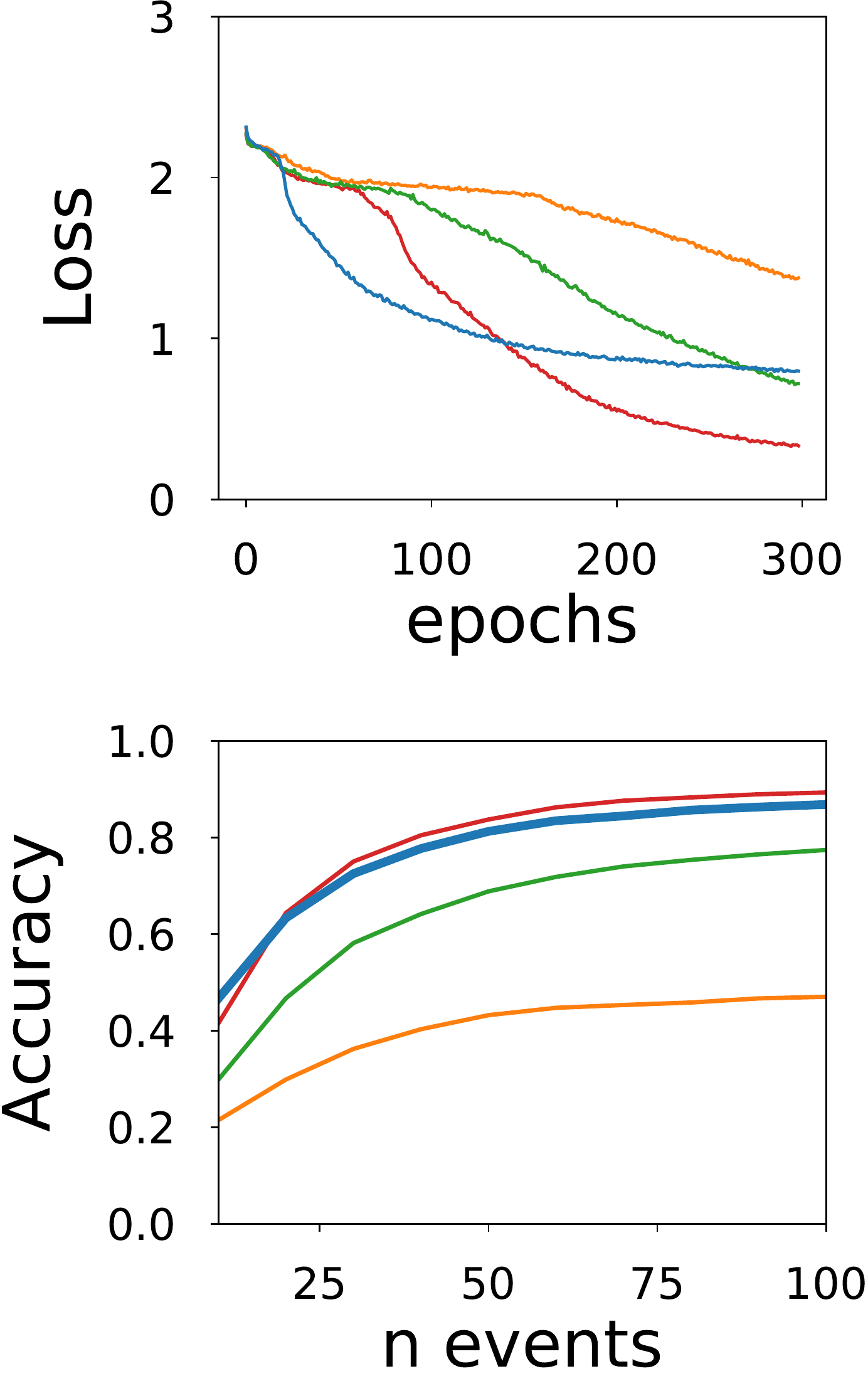}
        \includegraphics[height=3.6cm, width=1.36cm]{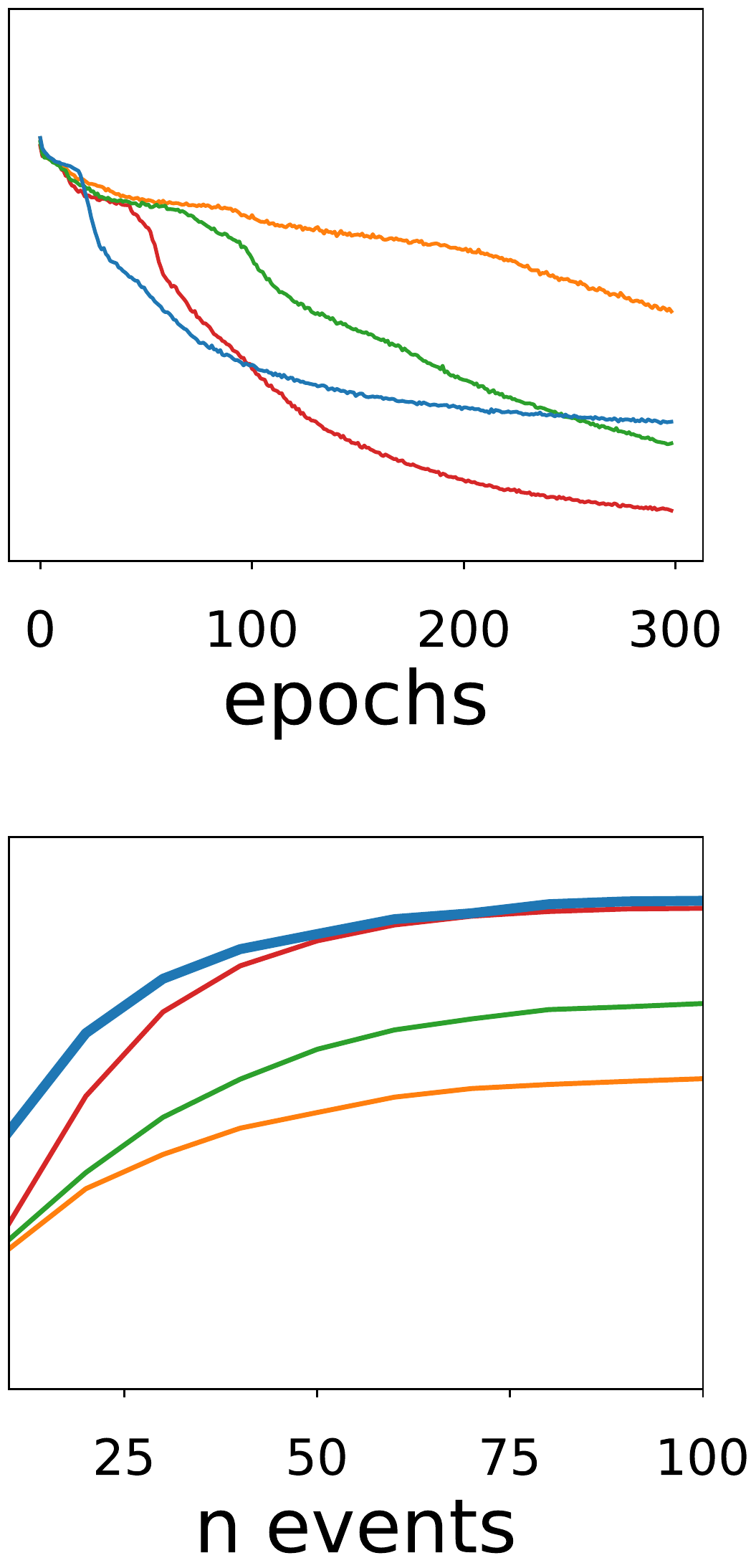}
        \includegraphics[height=3.6cm, width=1.36cm]{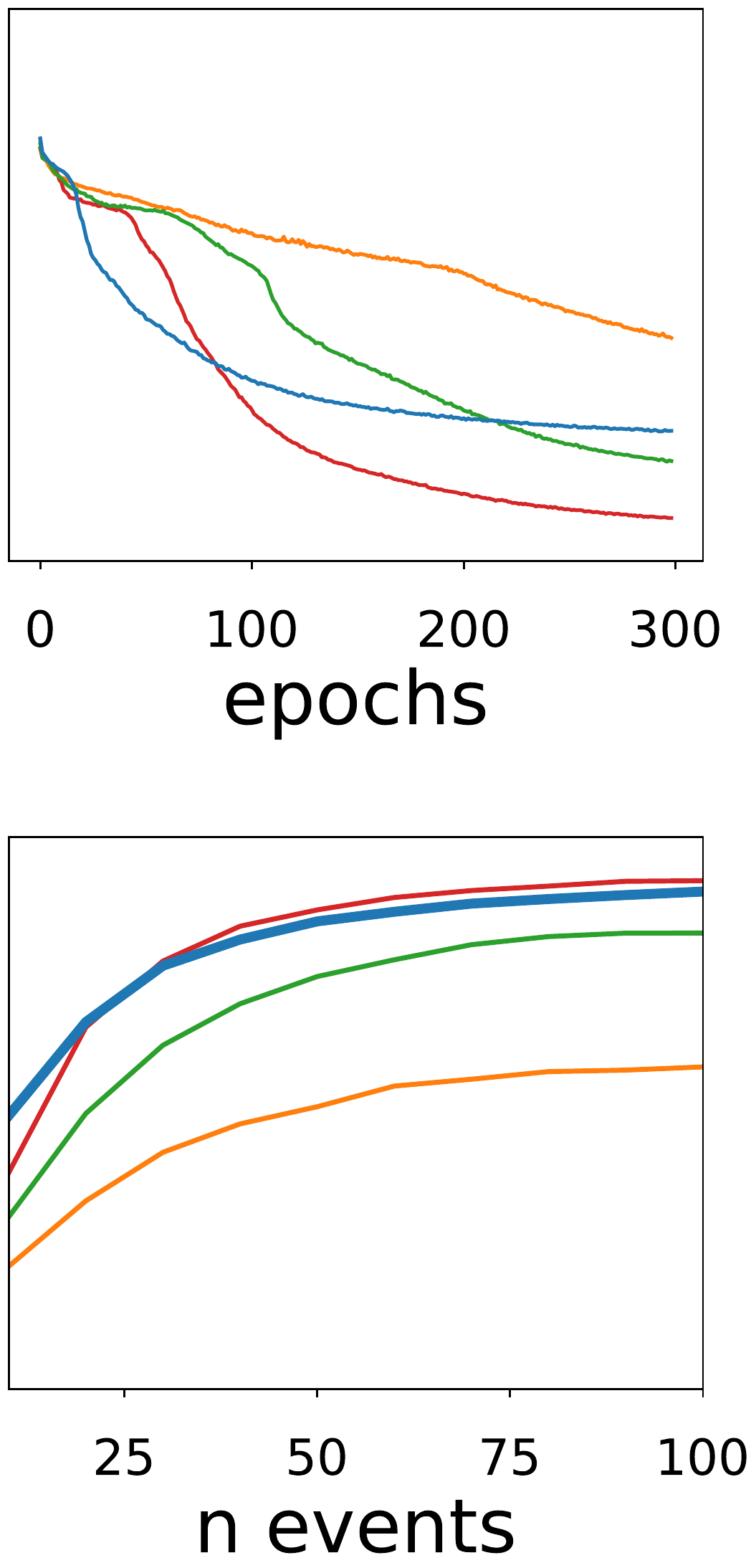}
        \caption{MNIST (300 epochs)}
    \end{subfigure}
    \begin{subfigure}[b]{0.31\textwidth}
        \includegraphics[height=3.6cm, width=1.78cm]{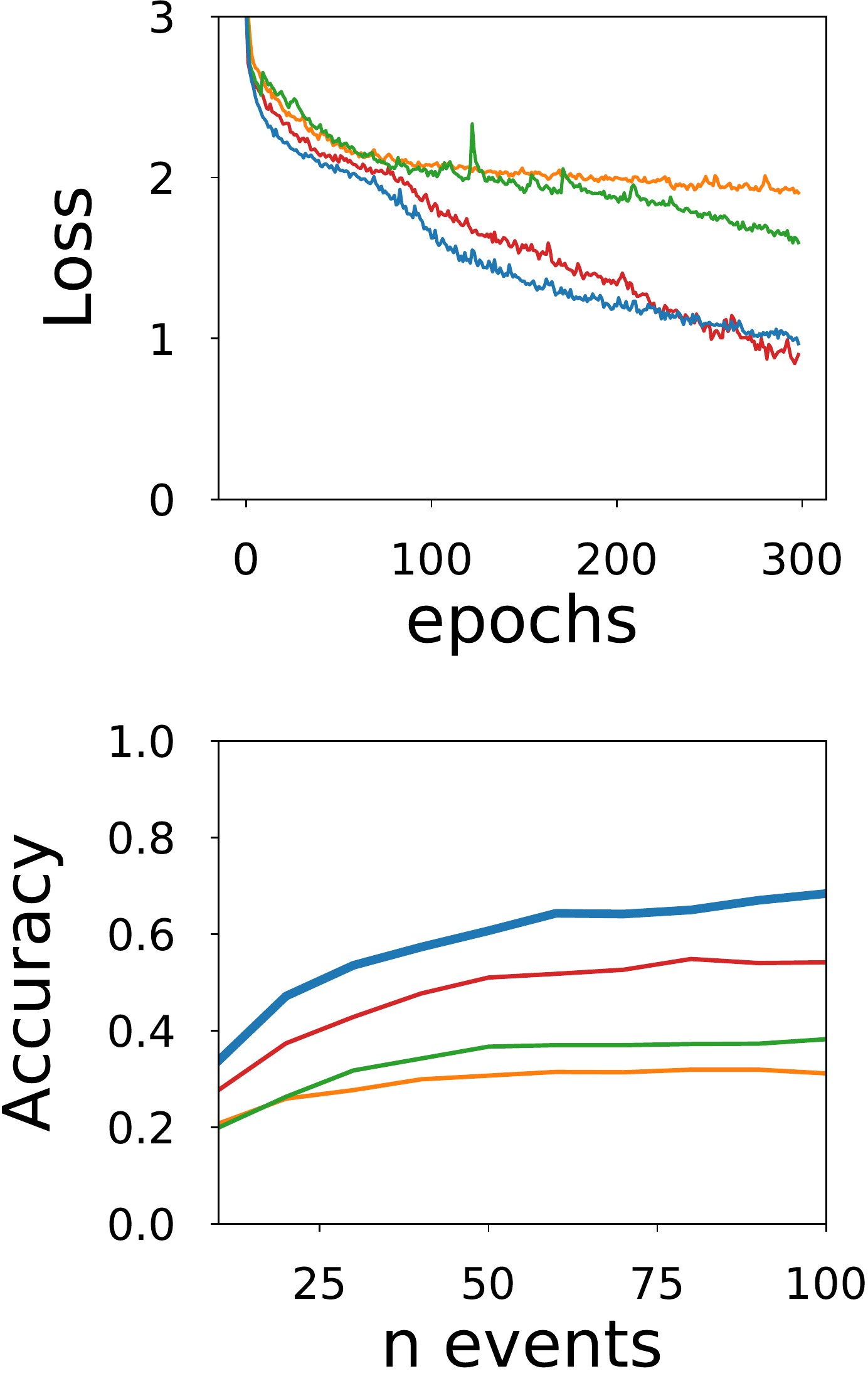}
        \includegraphics[height=3.6cm, width=1.36cm]{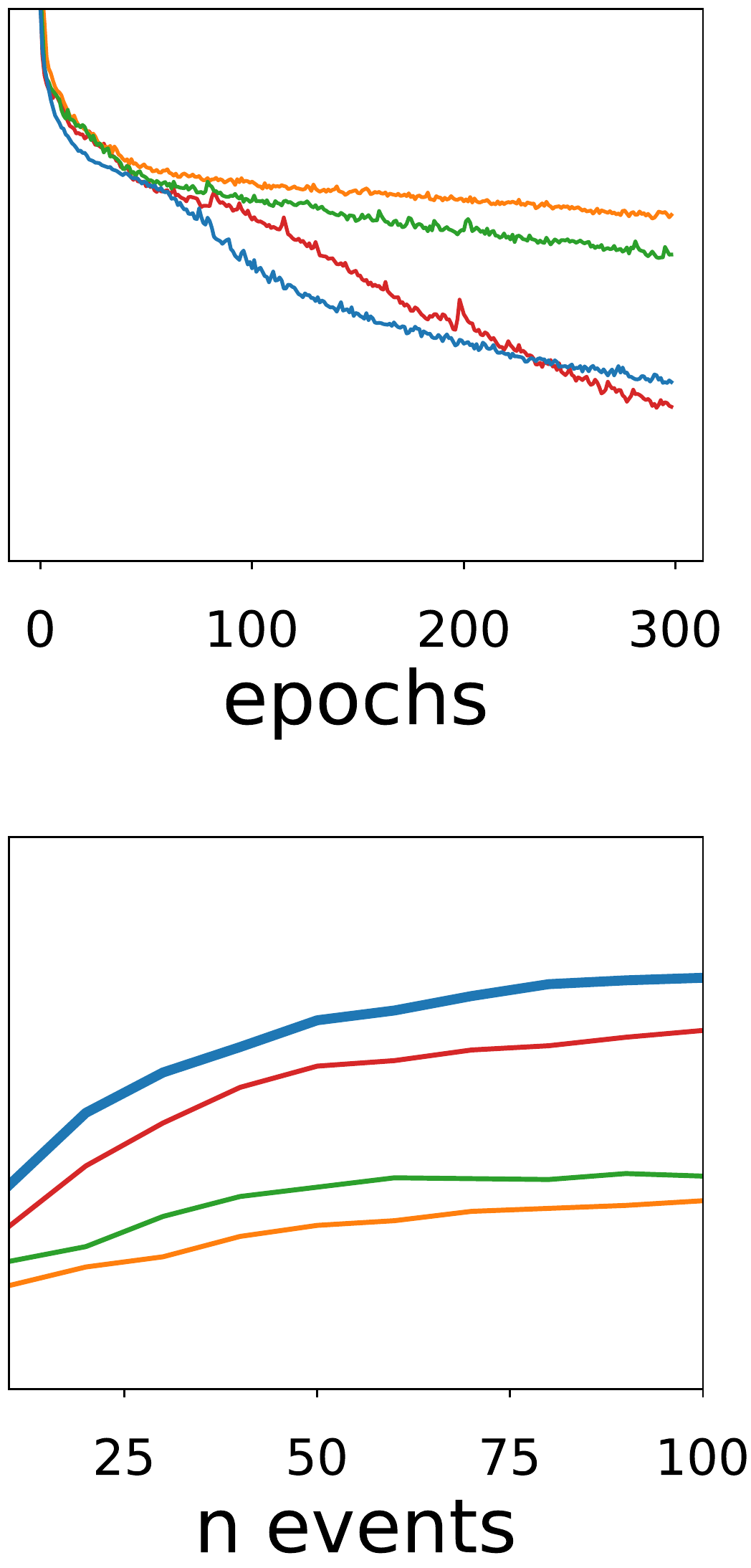}
        \includegraphics[height=3.6cm, width=1.36cm]{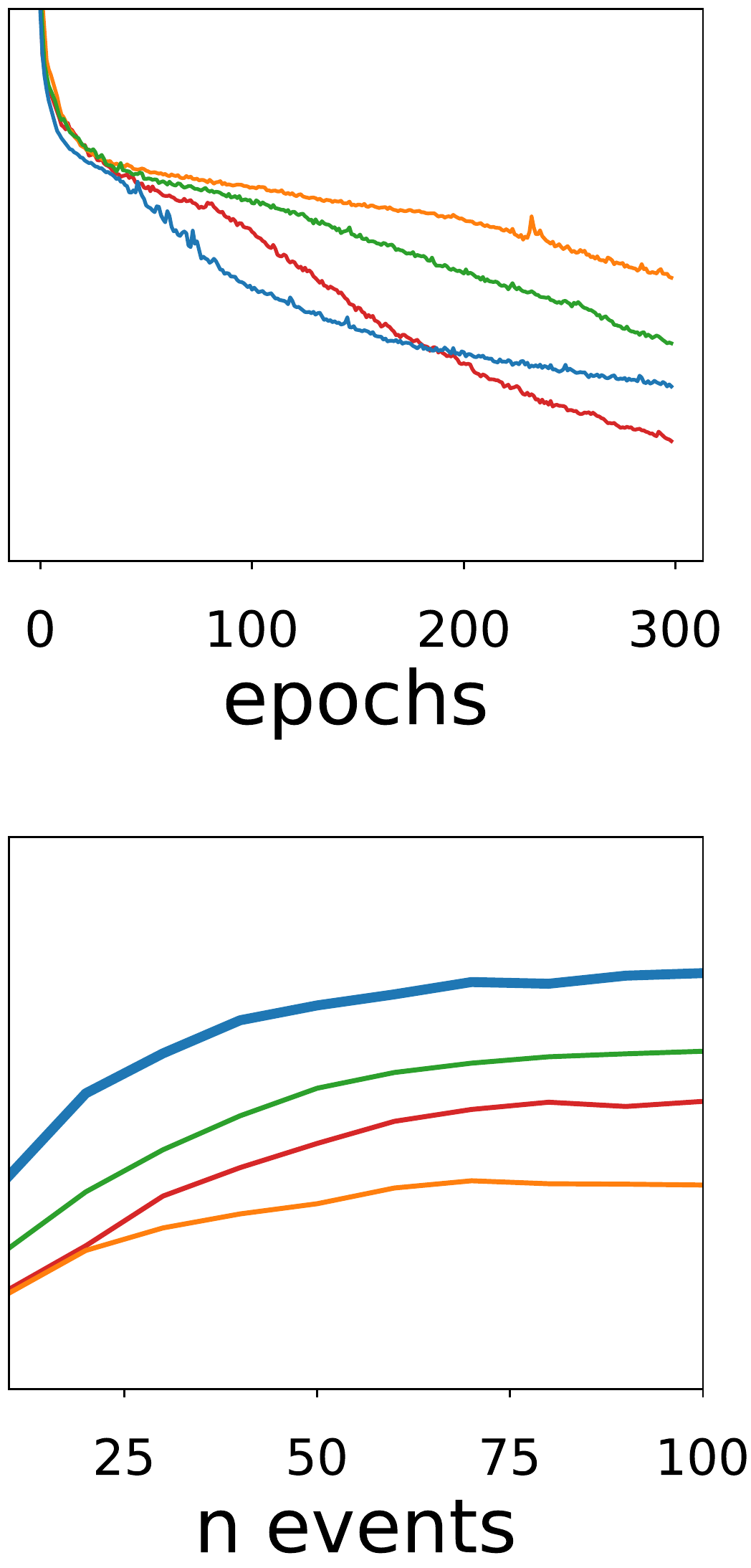}
        \caption{ASL (300 epochs)}
    \end{subfigure}
    \begin{subfigure}[b]{0.31\textwidth}
        \includegraphics[height=3.6cm, width=1.78cm]{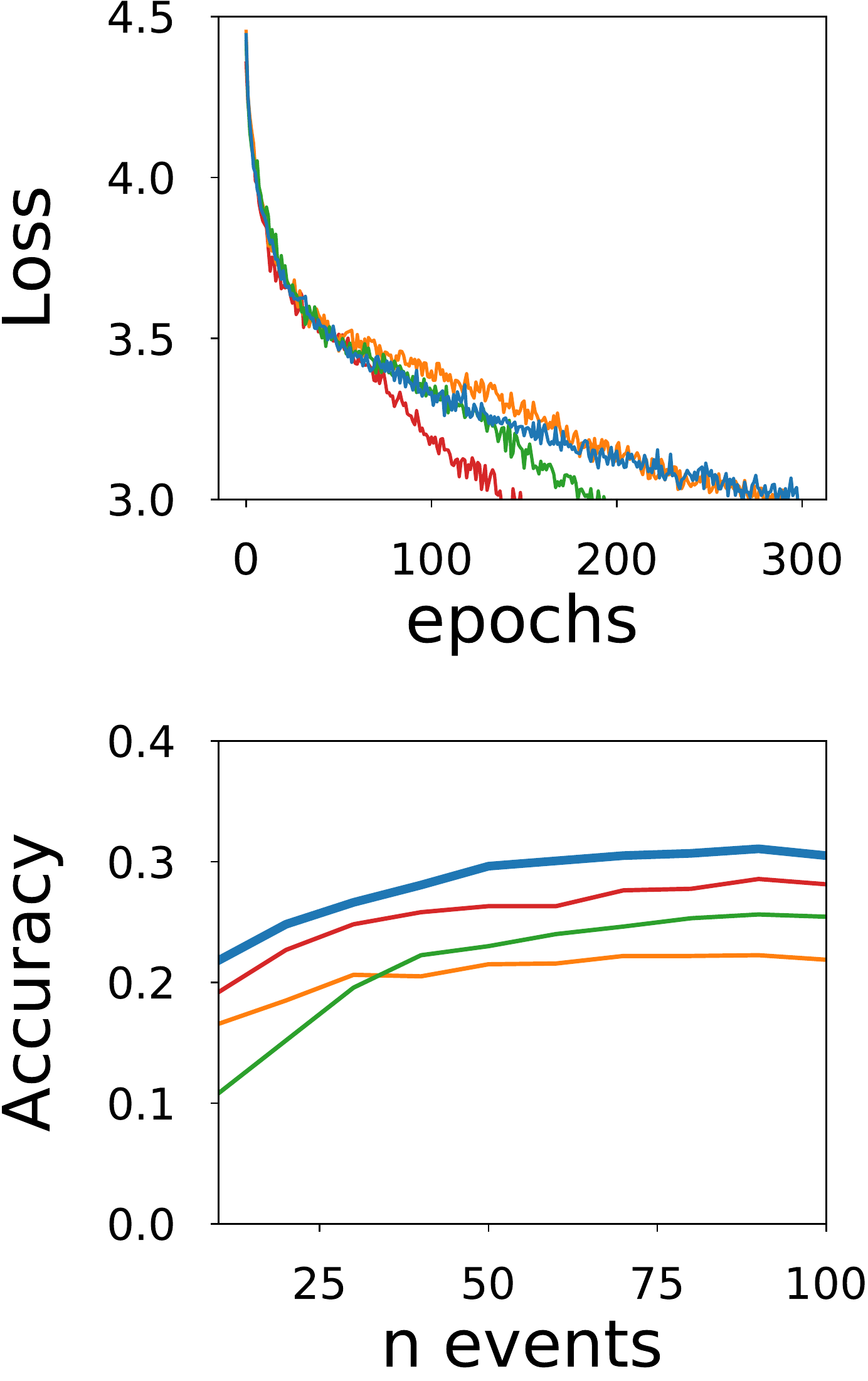}
        \includegraphics[height=3.6cm, width=1.36cm]{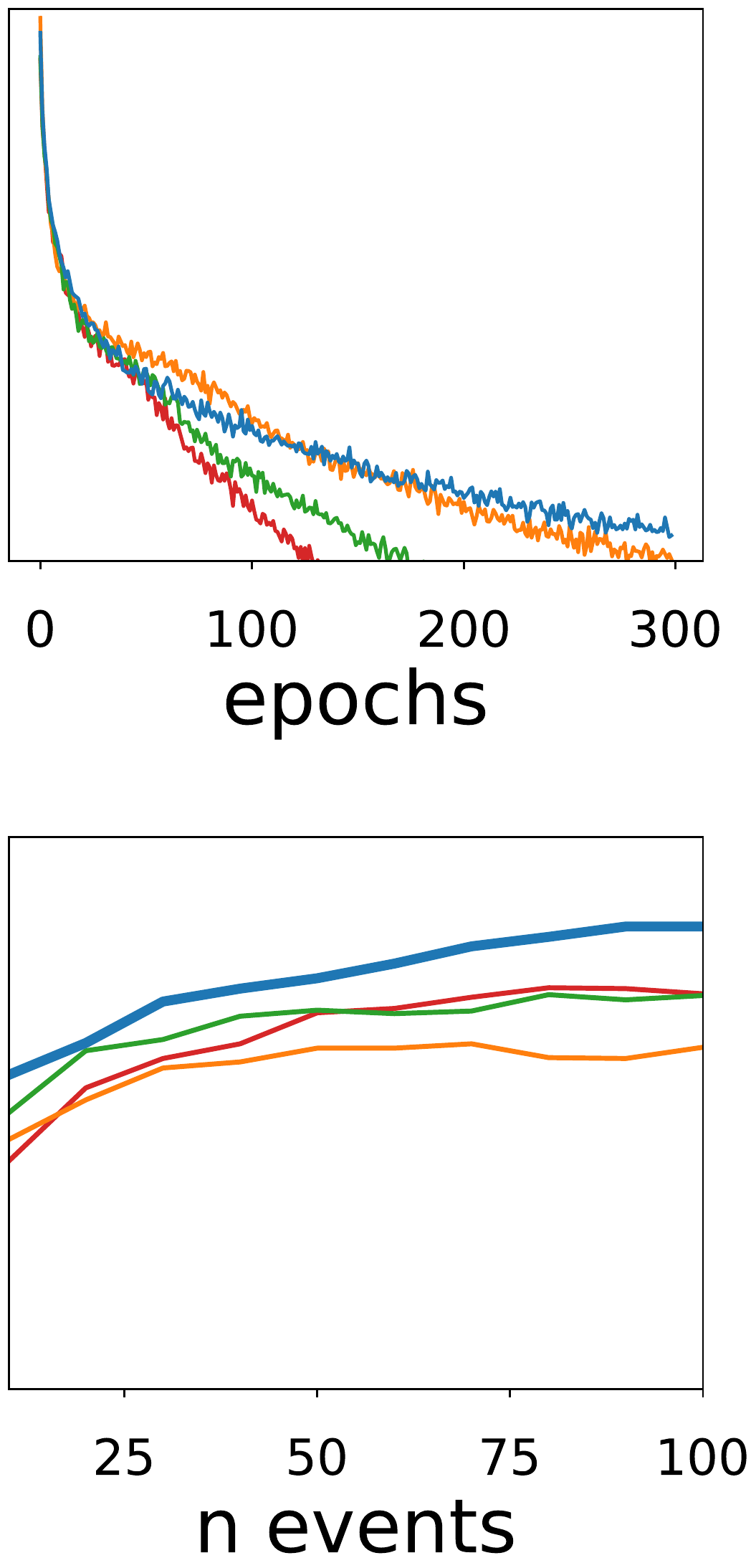}
        \includegraphics[height=3.6cm, width=1.36cm]{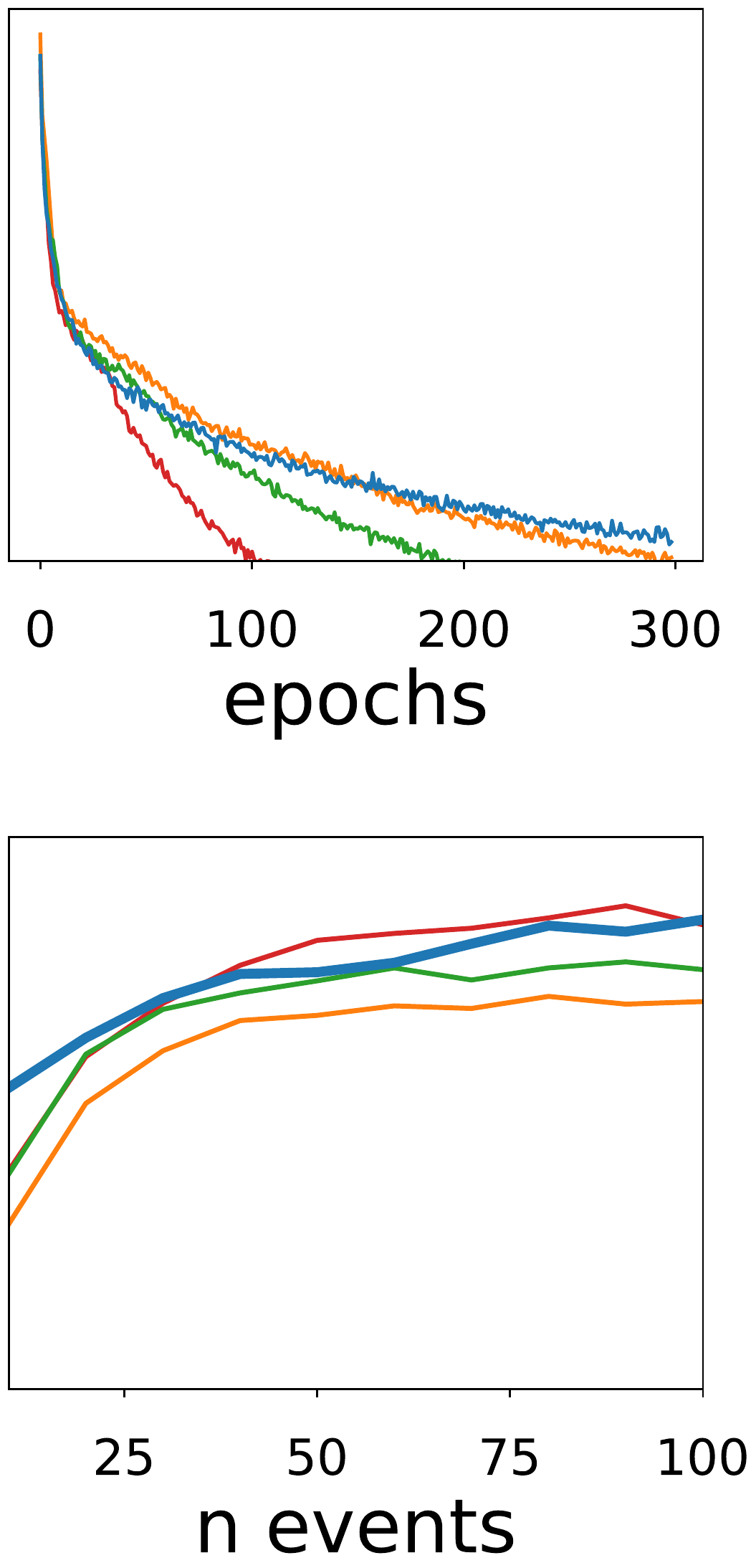}
        \caption{NCALTECH (300 epochs)}
    \end{subfigure}
    \caption{{\bf Summary of results.} Train/test losses and classification performance for INODE and multiple bi-LSTM baselines, with increasing number of inference events per digit from 10 to 100. The three images for each dataset sub-figure correspond to training-set fraction of $20\%$ (left), $40\%$ (center), and $100\%$ (right).}
    \label{fig:results_summary_bi}
    \vspace{-0.5cm}
\end{figure*}

\begin{figure*}[h!t]
    \captionsetup[subfigure]{justification=centering}
    \centering
    \includegraphics[width=0.4\linewidth]{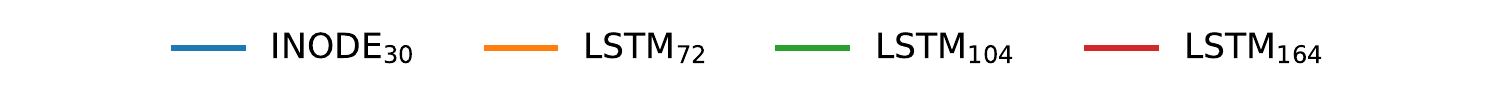}\\\vspace{5pt}
    \begin{subfigure}[b]{0.31\textwidth}
        \includegraphics[height=3.6cm, width=1.78cm]{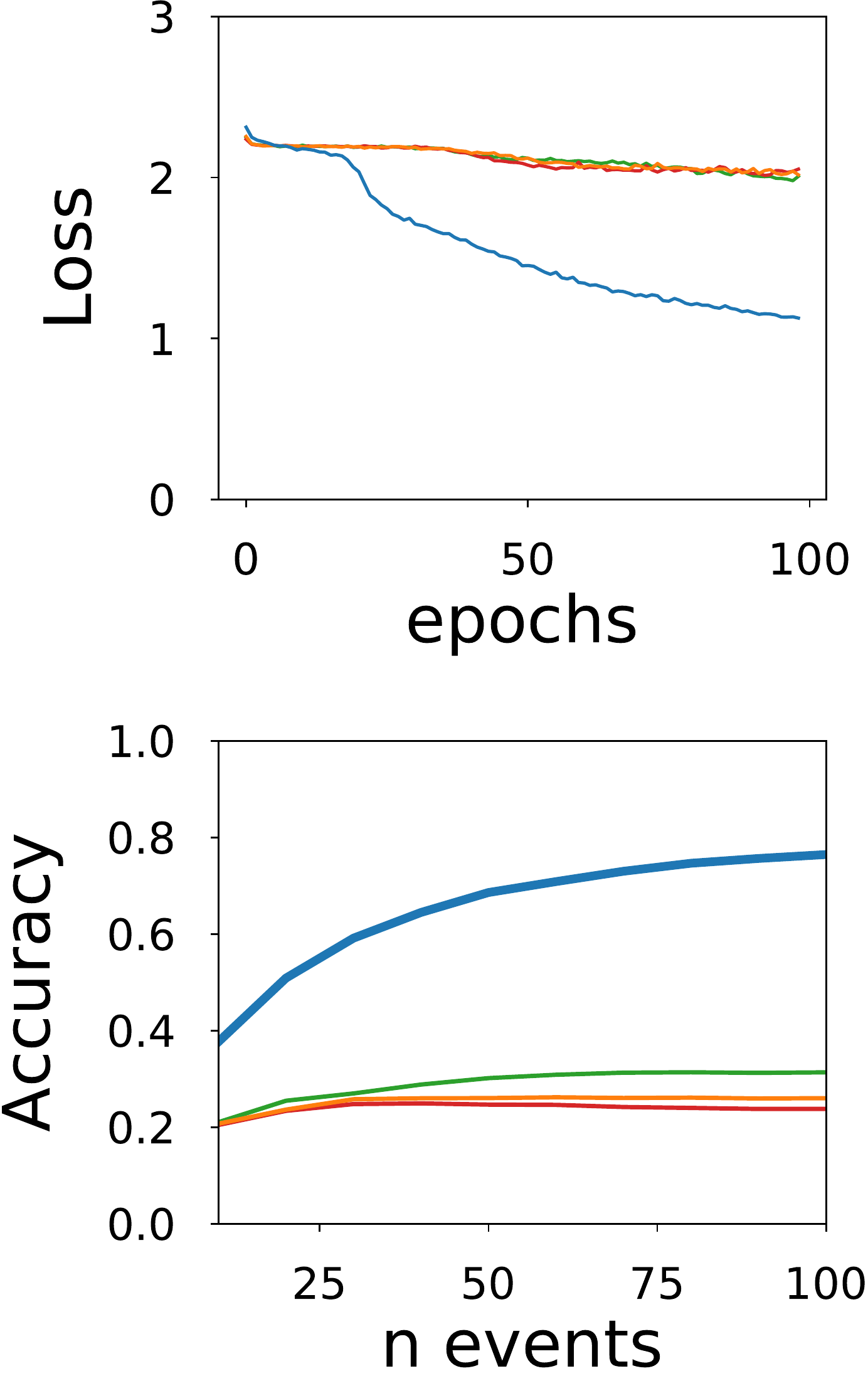}
        \includegraphics[height=3.6cm, width=1.36cm]{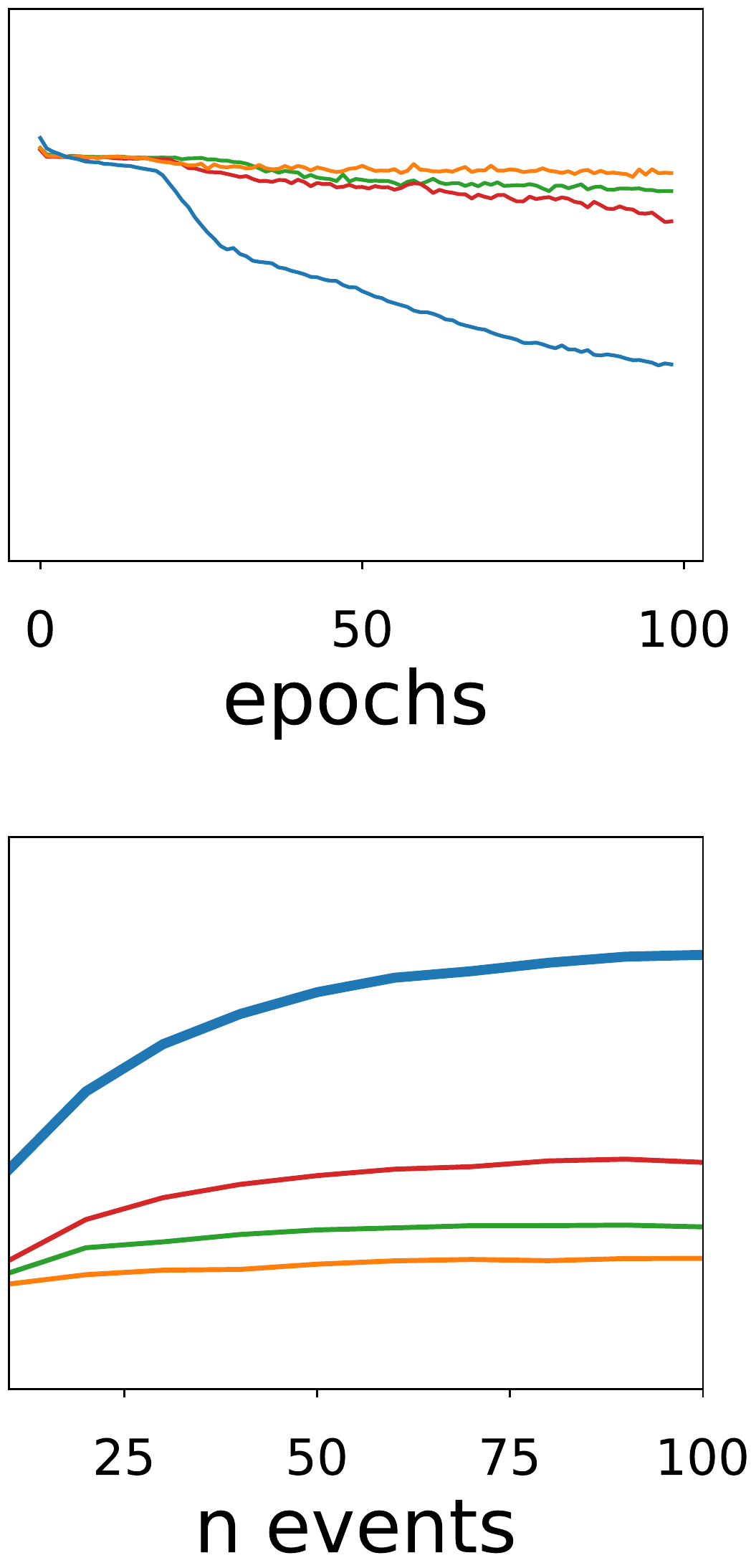}
        \includegraphics[height=3.6cm, width=1.36cm]{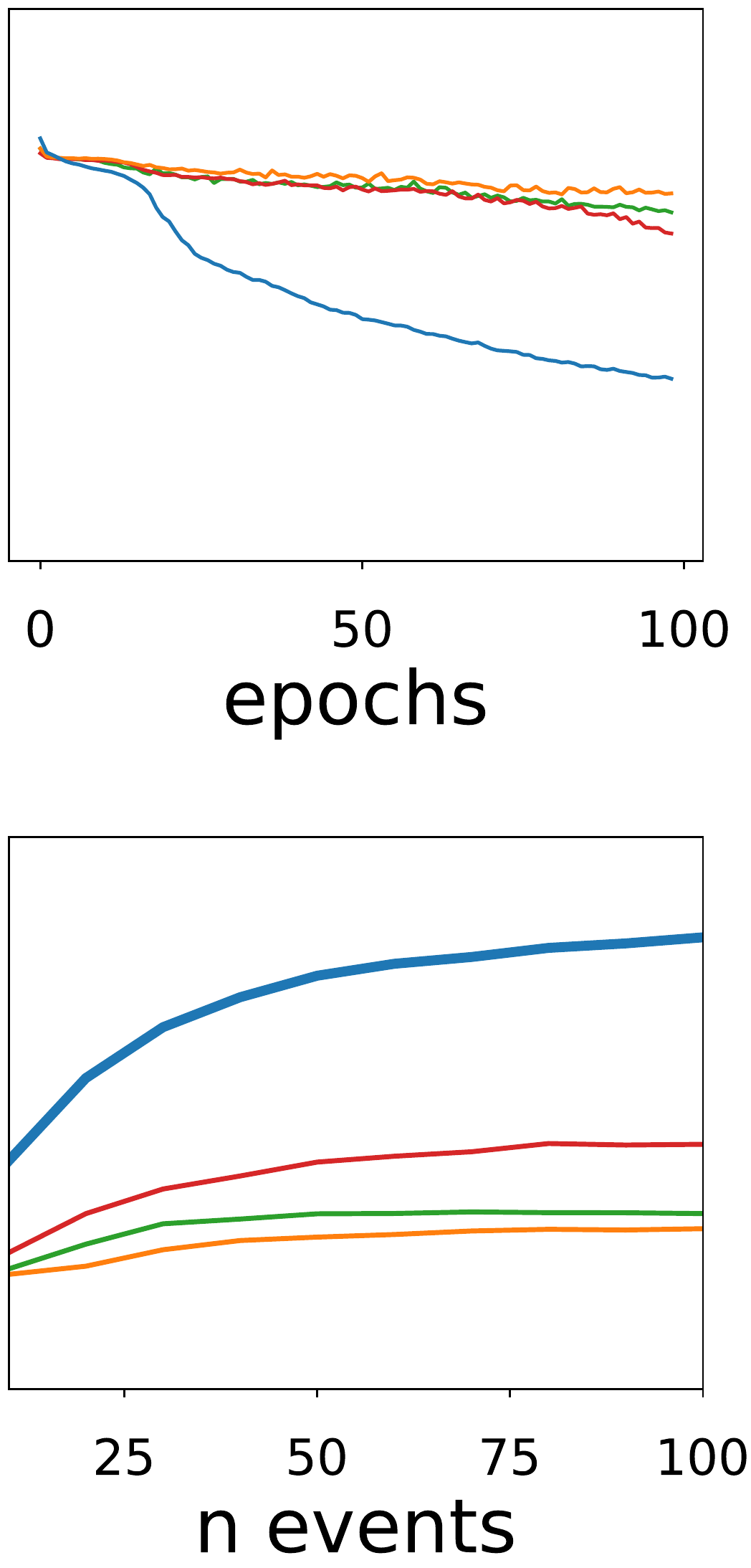}
        \caption{MNIST (100 epochs)}
        \vspace{5pt}
    \end{subfigure}
    \begin{subfigure}[b]{0.31\textwidth}
        \includegraphics[height=3.6cm, width=1.78cm]{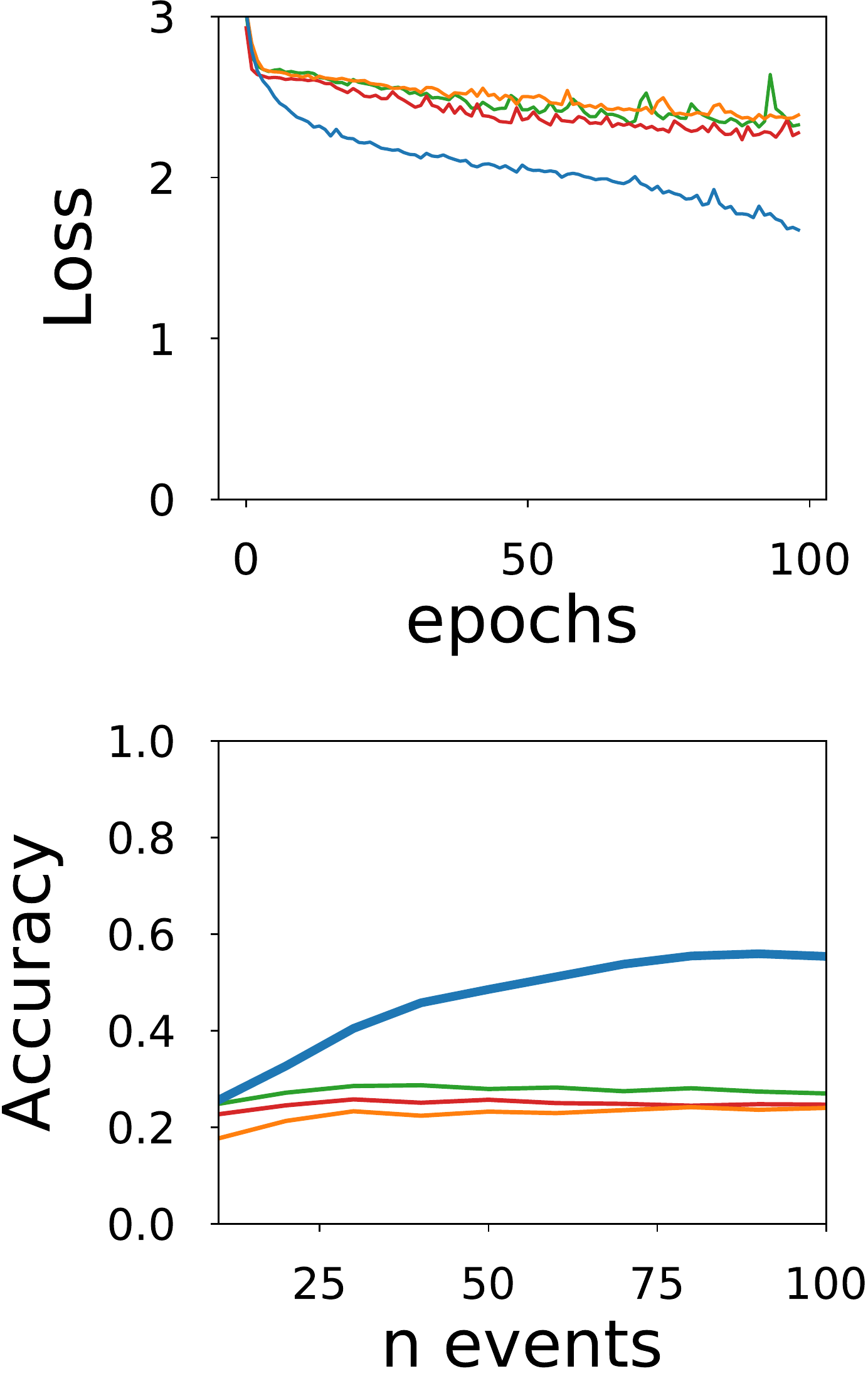}
        \includegraphics[height=3.6cm, width=1.36cm]{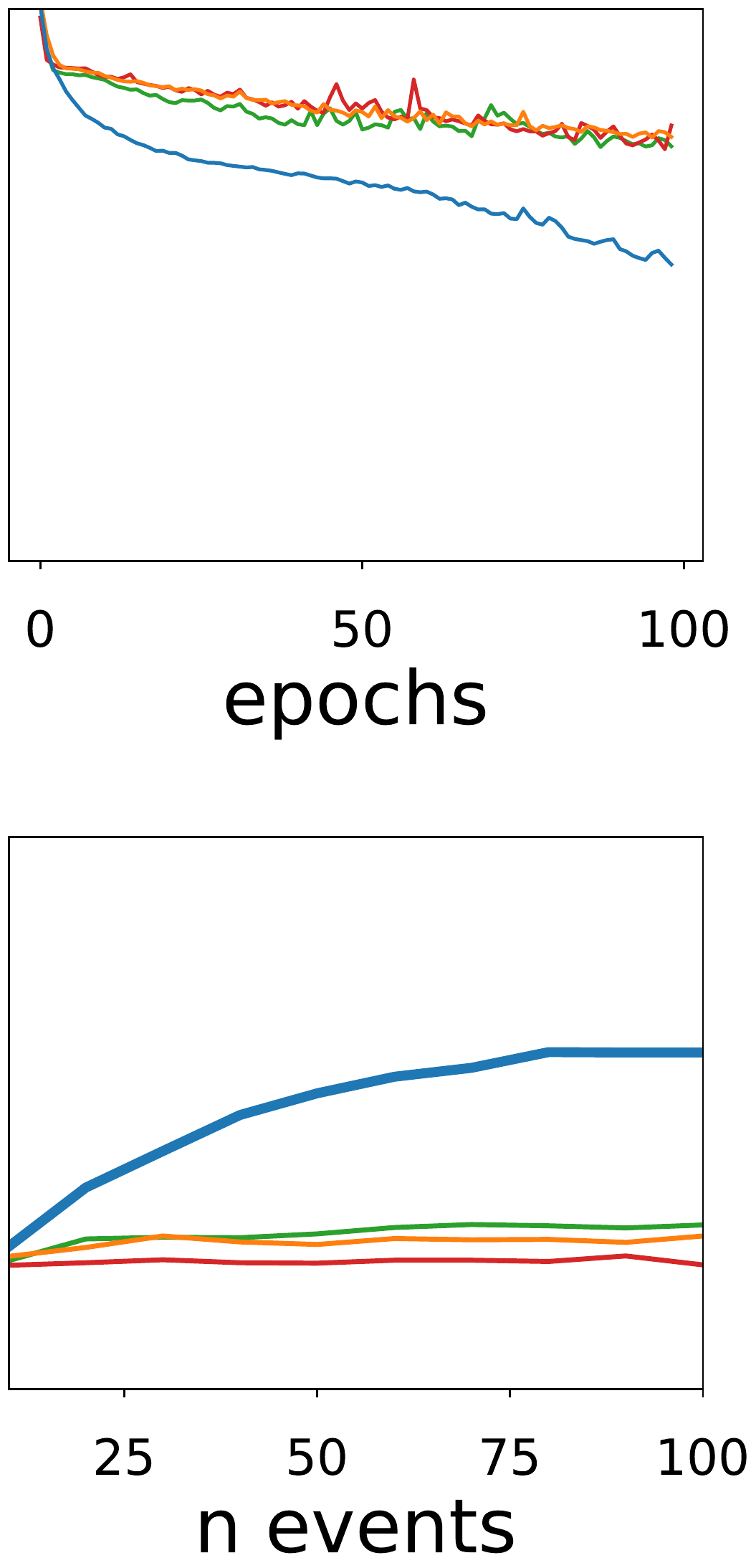}
        \includegraphics[height=3.6cm, width=1.36cm]{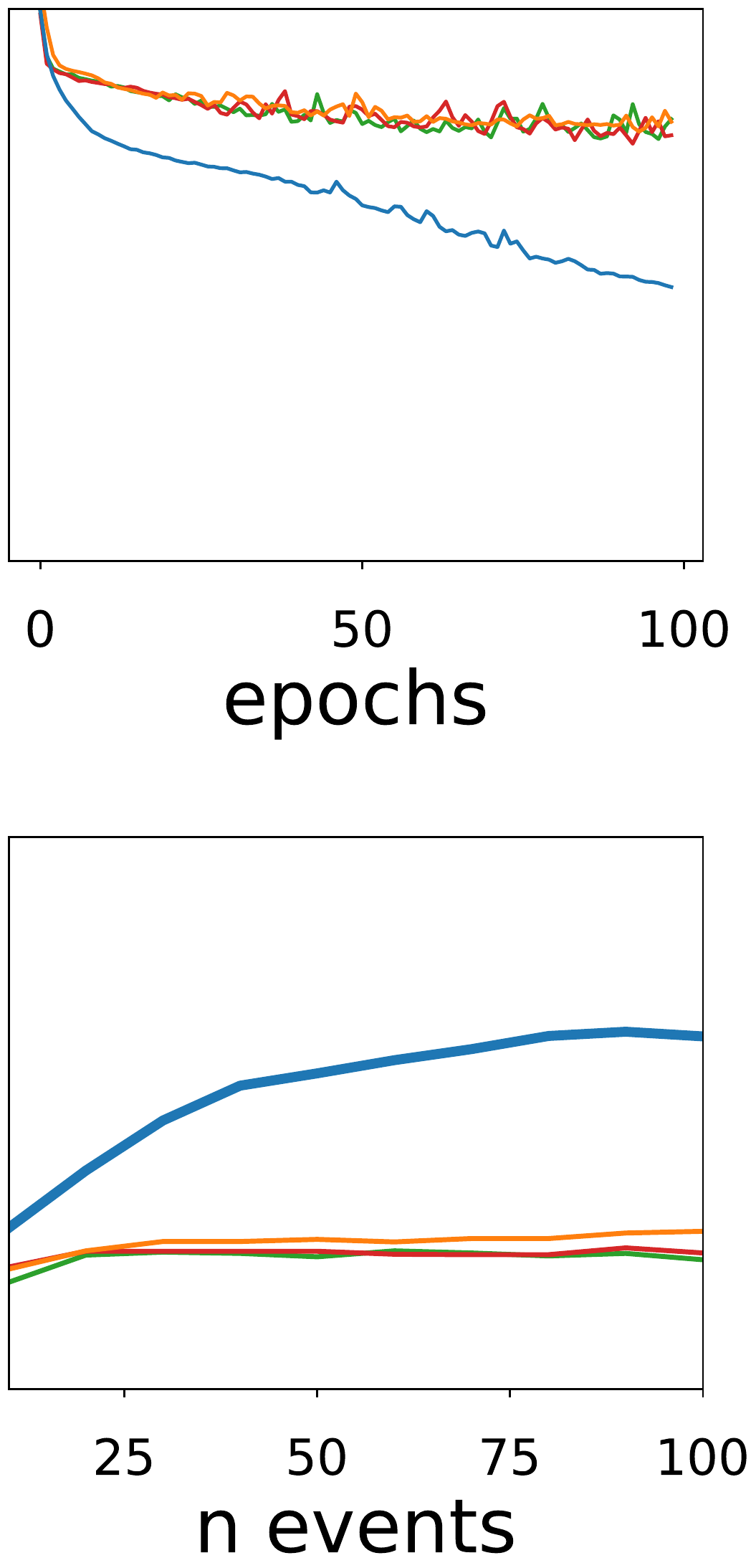}
        \caption{ASL (100 epochs)}
        \vspace{5pt}
    \end{subfigure}
    \begin{subfigure}[b]{0.31\textwidth}
        \includegraphics[height=3.6cm, width=1.78cm]{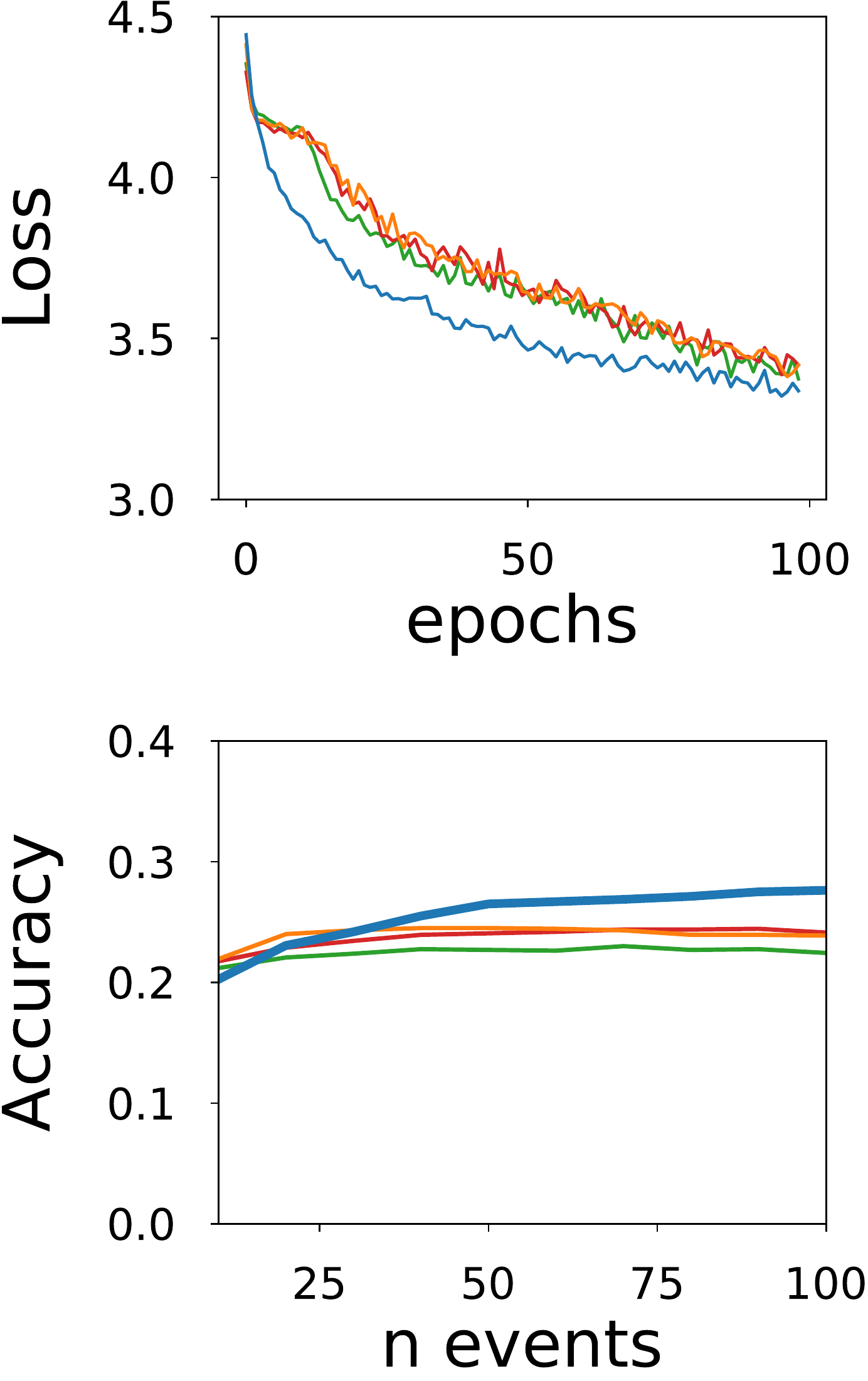}
        \includegraphics[height=3.6cm, width=1.36cm]{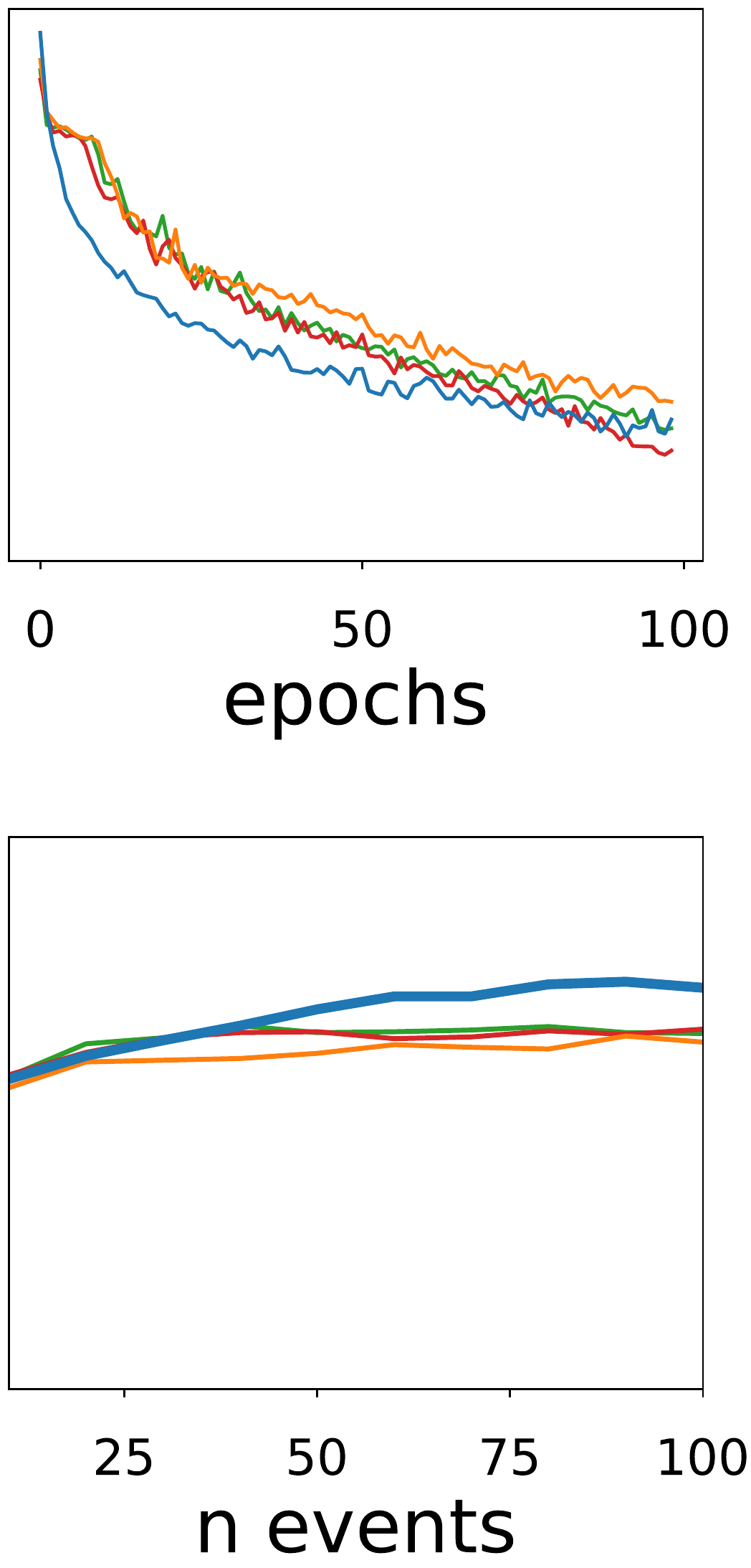}
        \includegraphics[height=3.6cm, width=1.36cm]{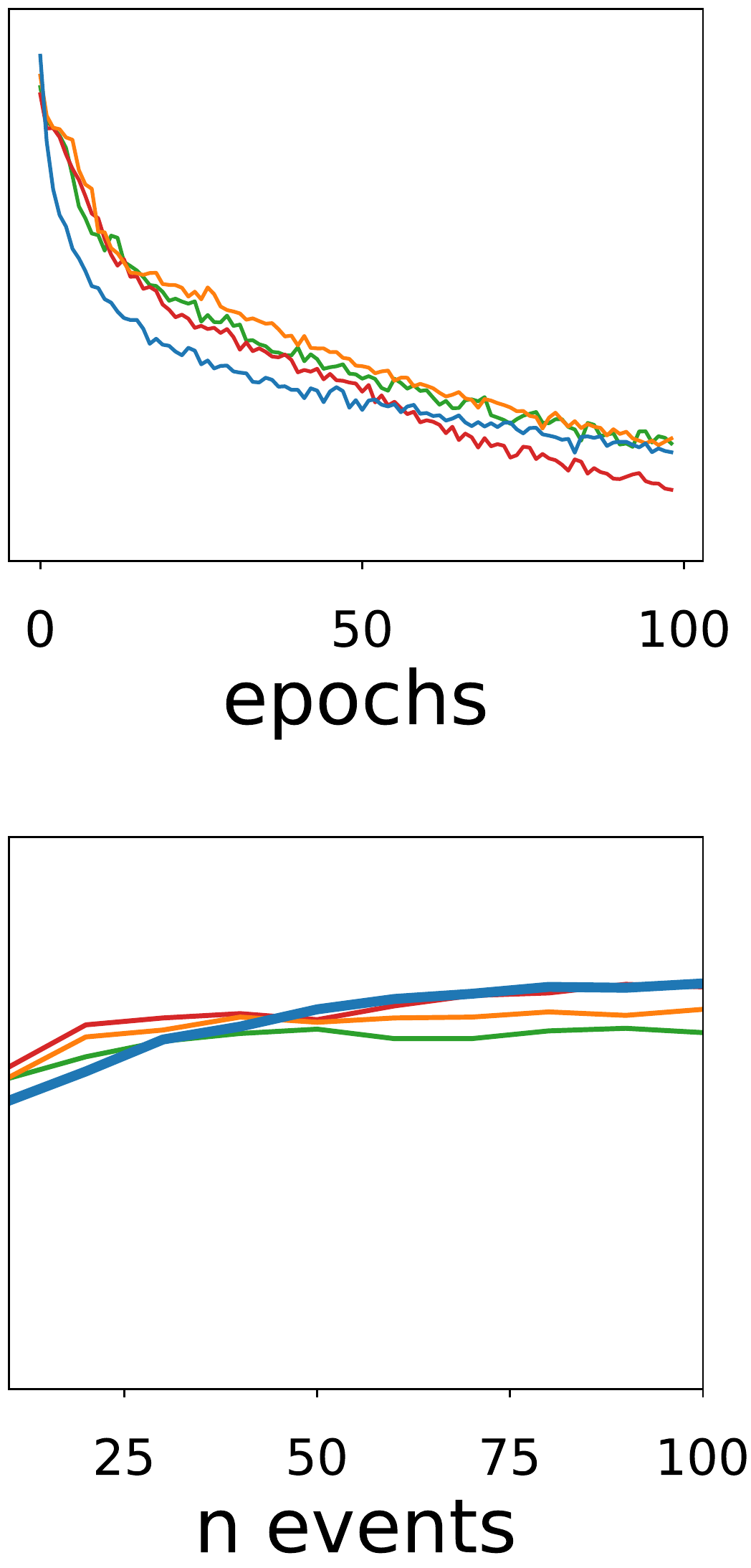}
        \caption{NCALTECH (100 epochs)}
        \vspace{5pt}
    \end{subfigure}
    \begin{subfigure}[b]{0.31\textwidth}
        \includegraphics[height=3.6cm, width=1.78cm]{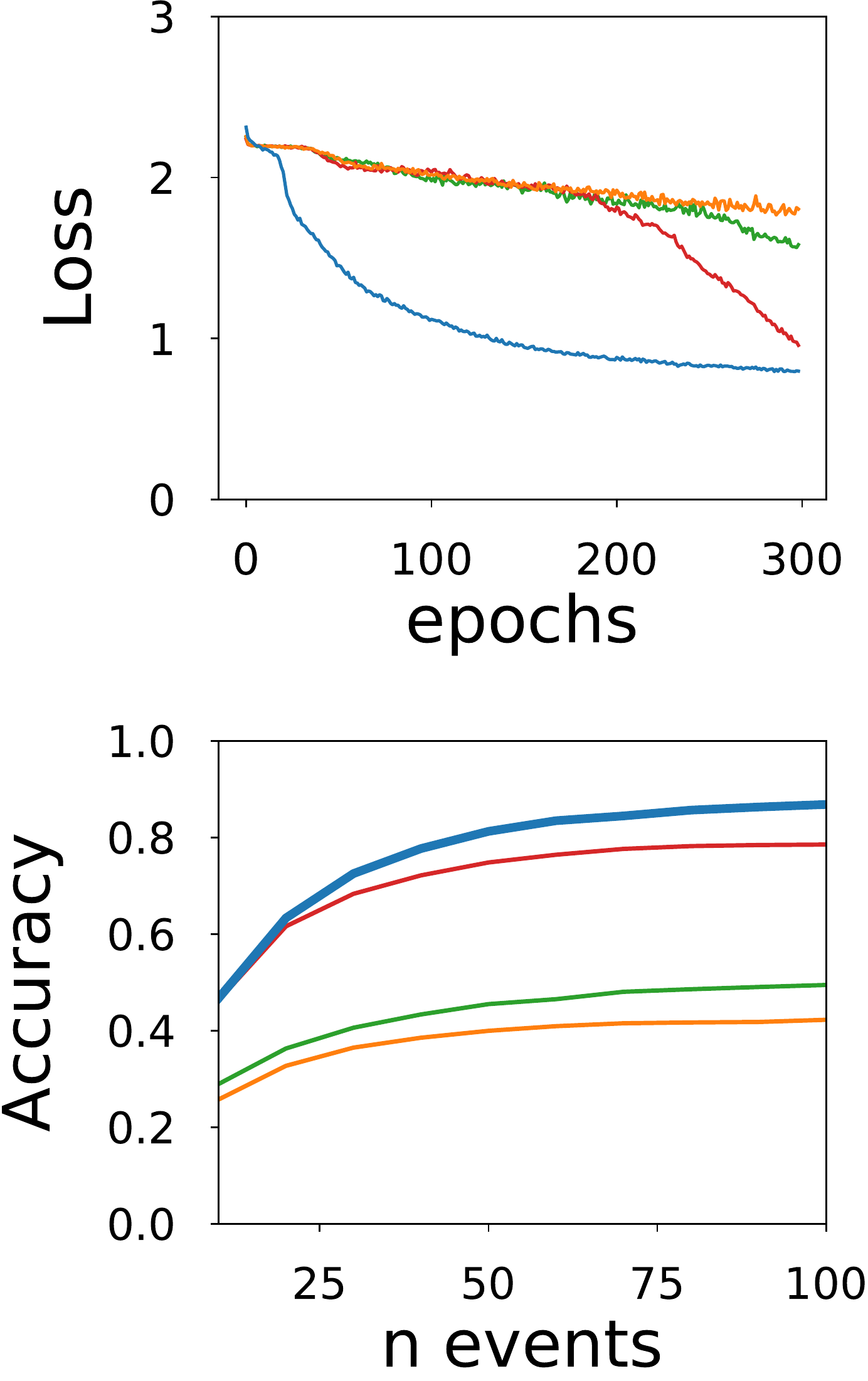}
        \includegraphics[height=3.6cm, width=1.36cm]{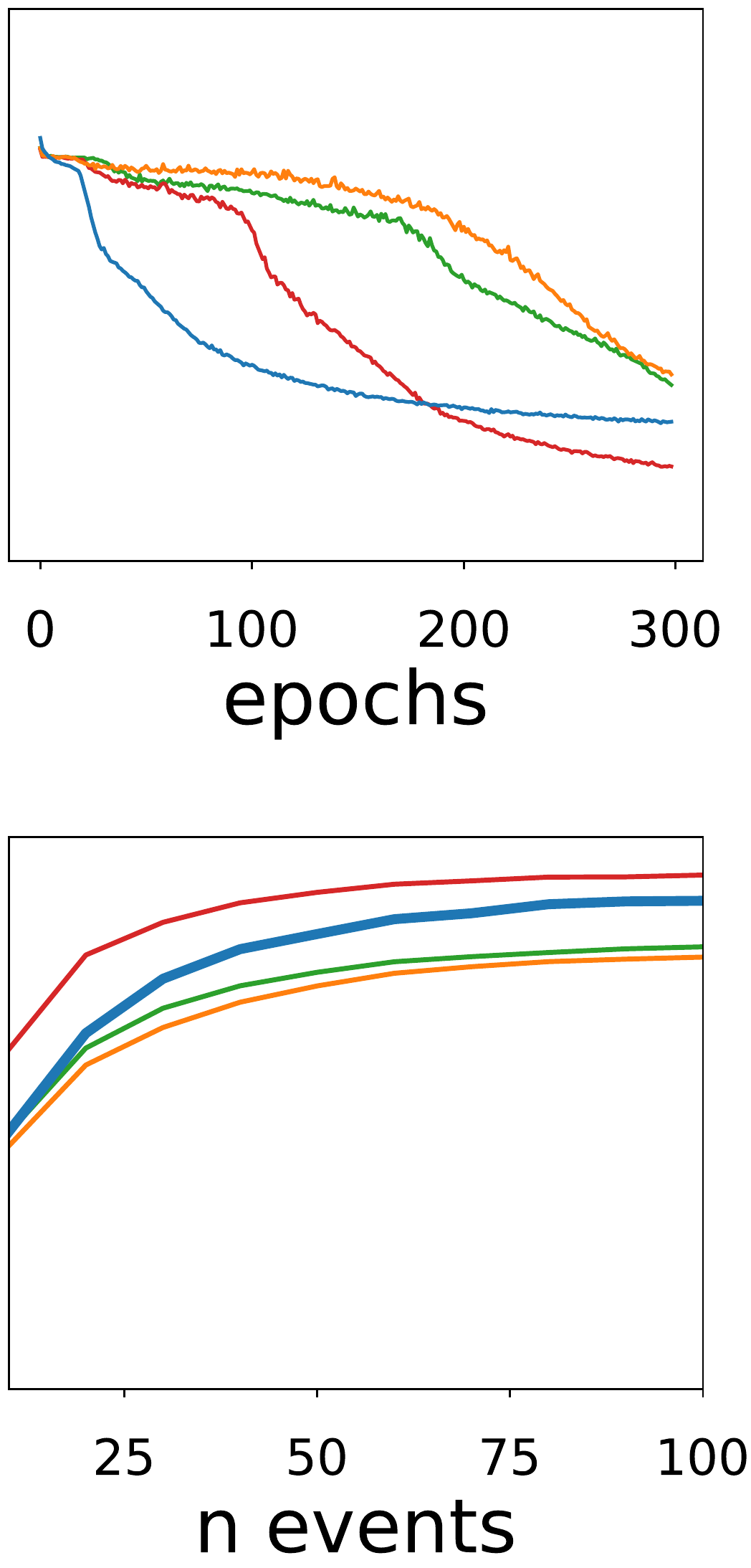}
        \includegraphics[height=3.6cm, width=1.36cm]{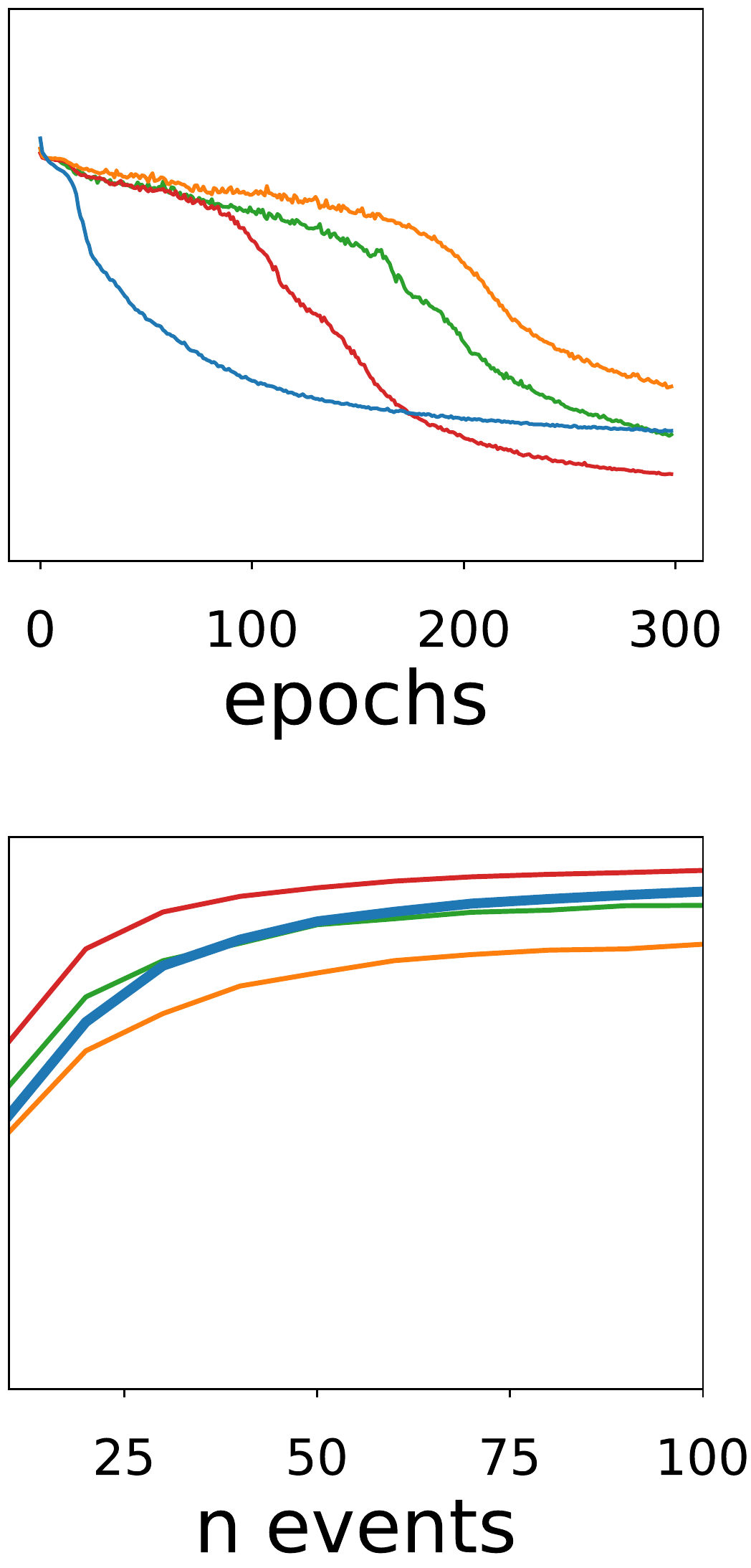}
        \caption{MNIST (300 epochs)}
    \end{subfigure}
    \begin{subfigure}[b]{0.31\textwidth}
        \includegraphics[height=3.6cm, width=1.78cm]{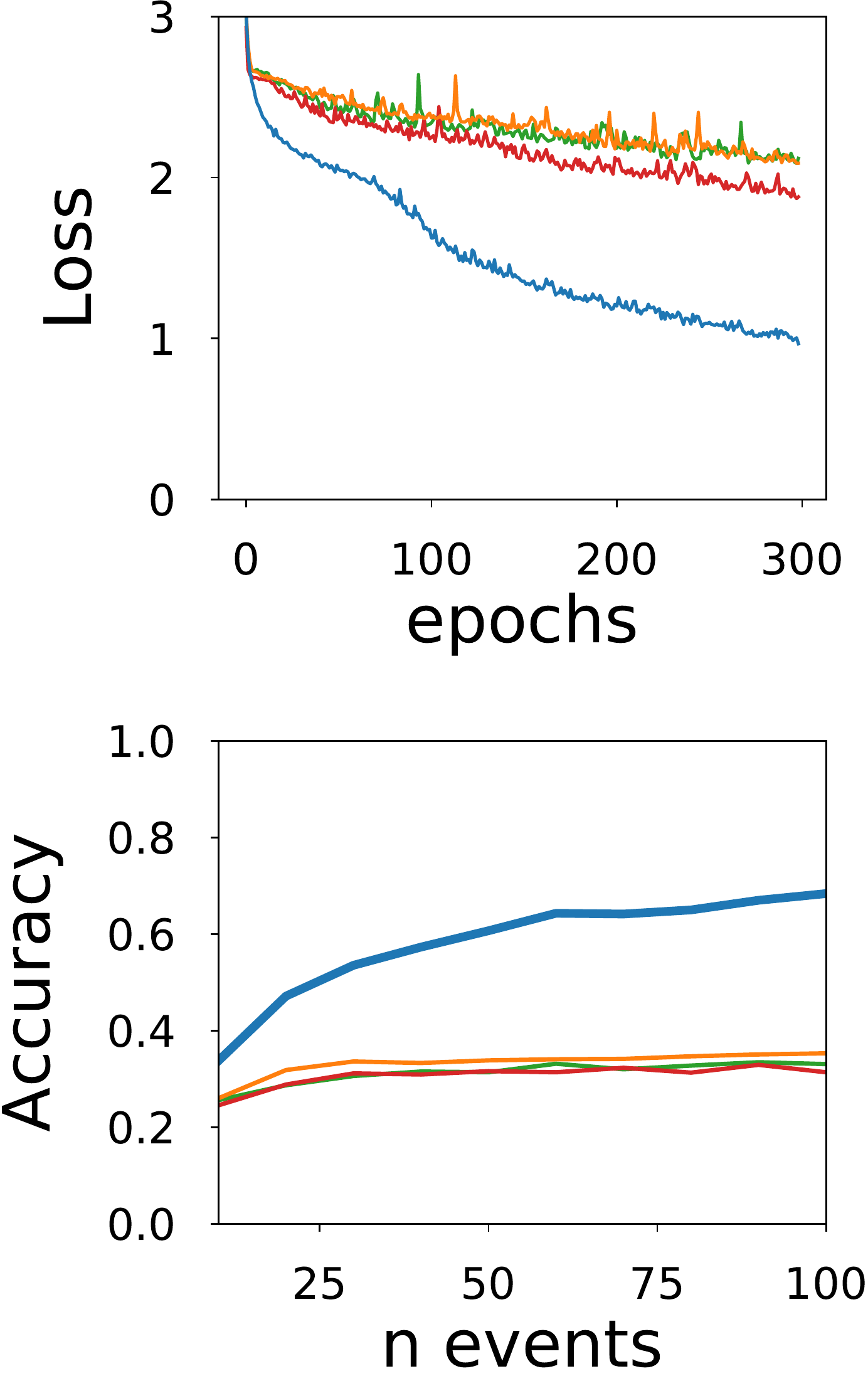}
        \includegraphics[height=3.6cm, width=1.36cm]{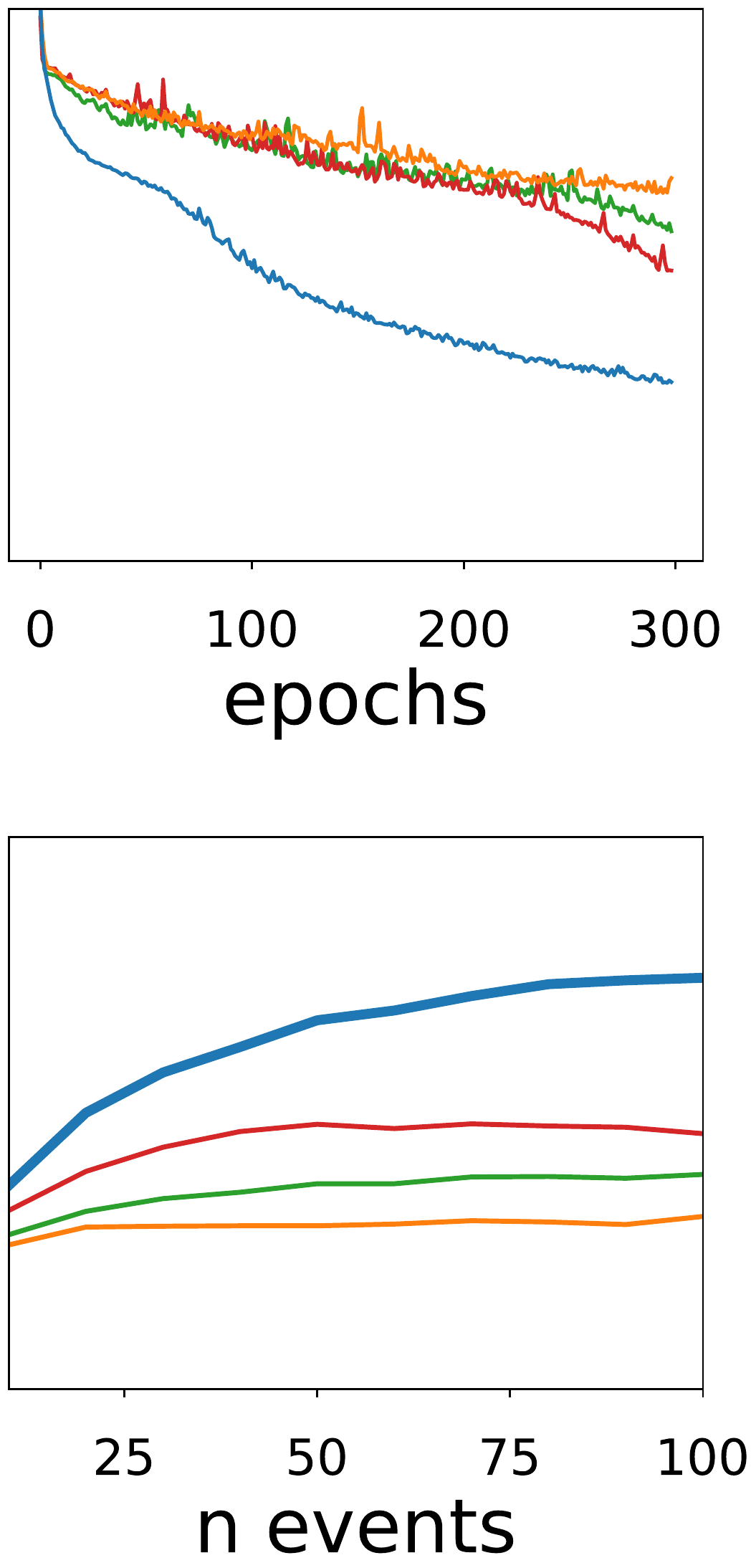}
        \includegraphics[height=3.6cm, width=1.36cm]{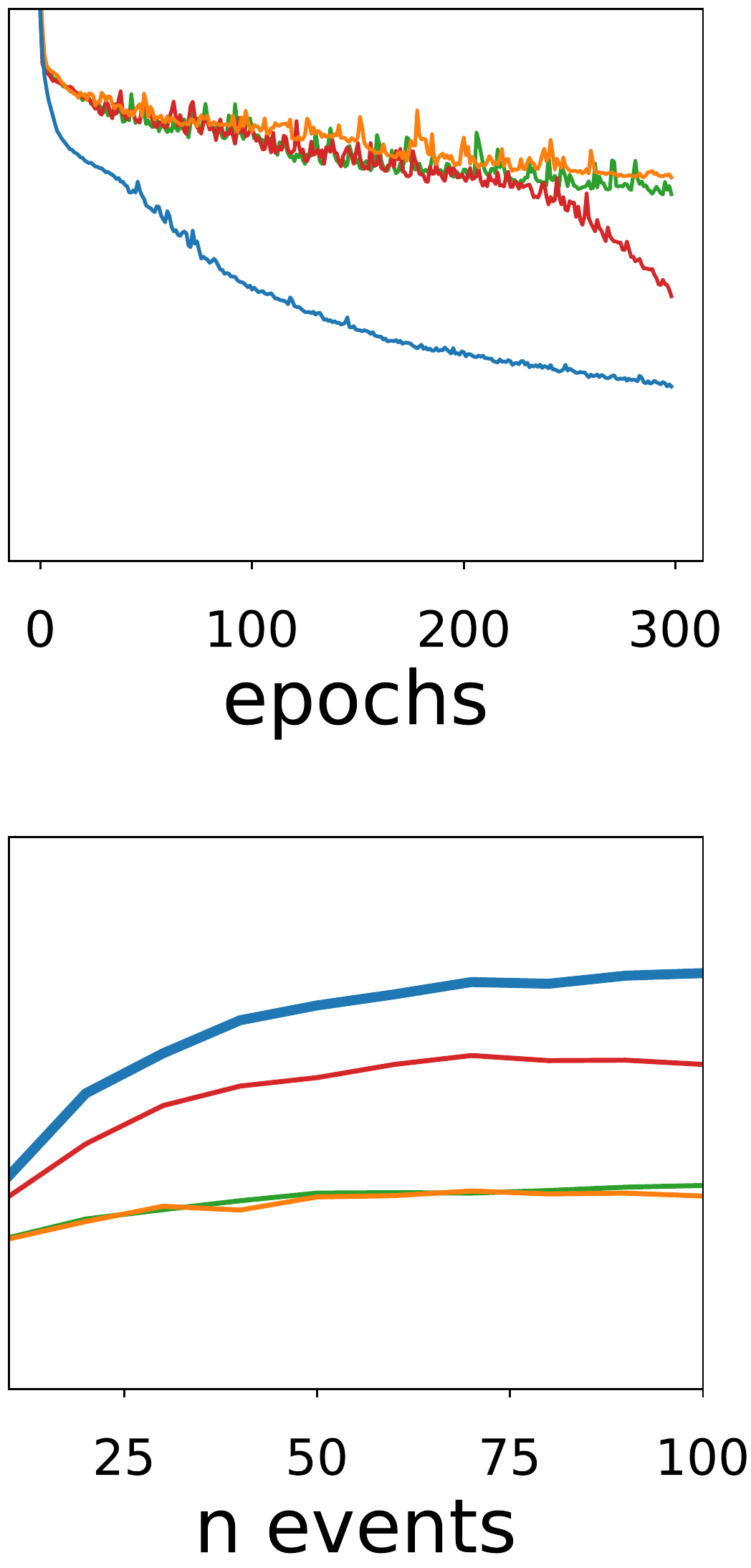}
        \caption{ASL (300 epochs)}
    \end{subfigure}
    \begin{subfigure}[b]{0.31\textwidth}
        \includegraphics[height=3.6cm, width=1.78cm]{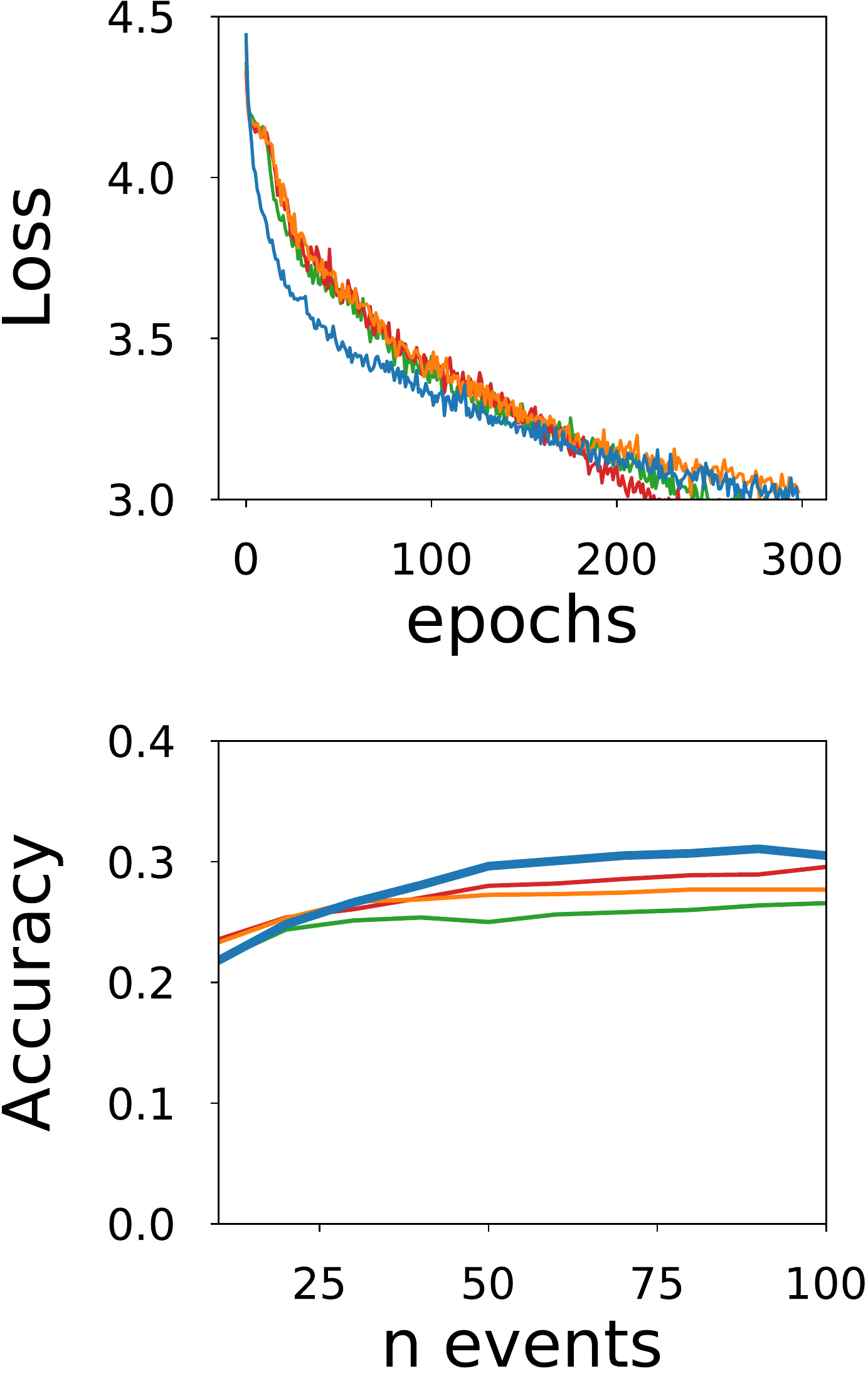}
        \includegraphics[height=3.6cm, width=1.36cm]{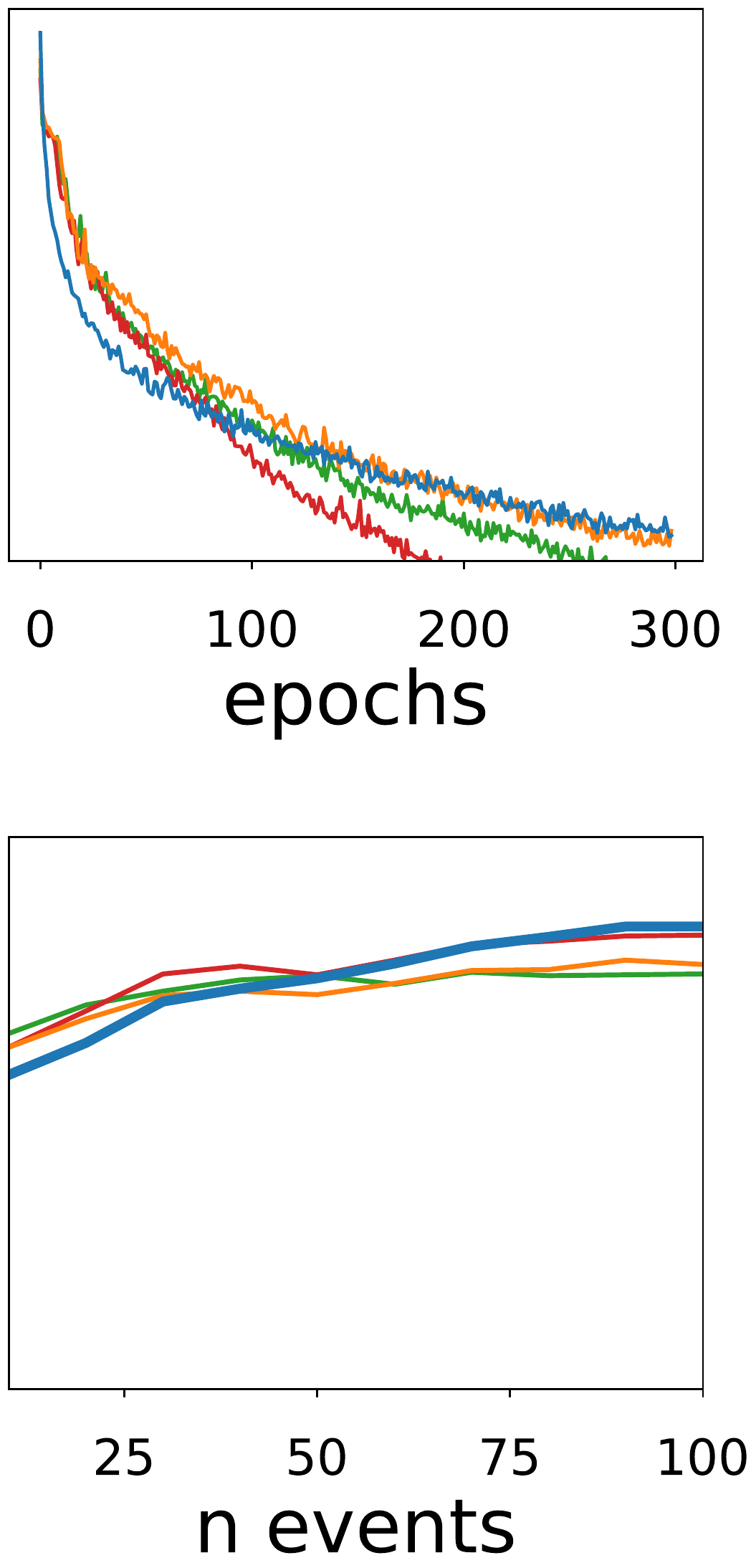}
        \includegraphics[height=3.6cm, width=1.36cm]{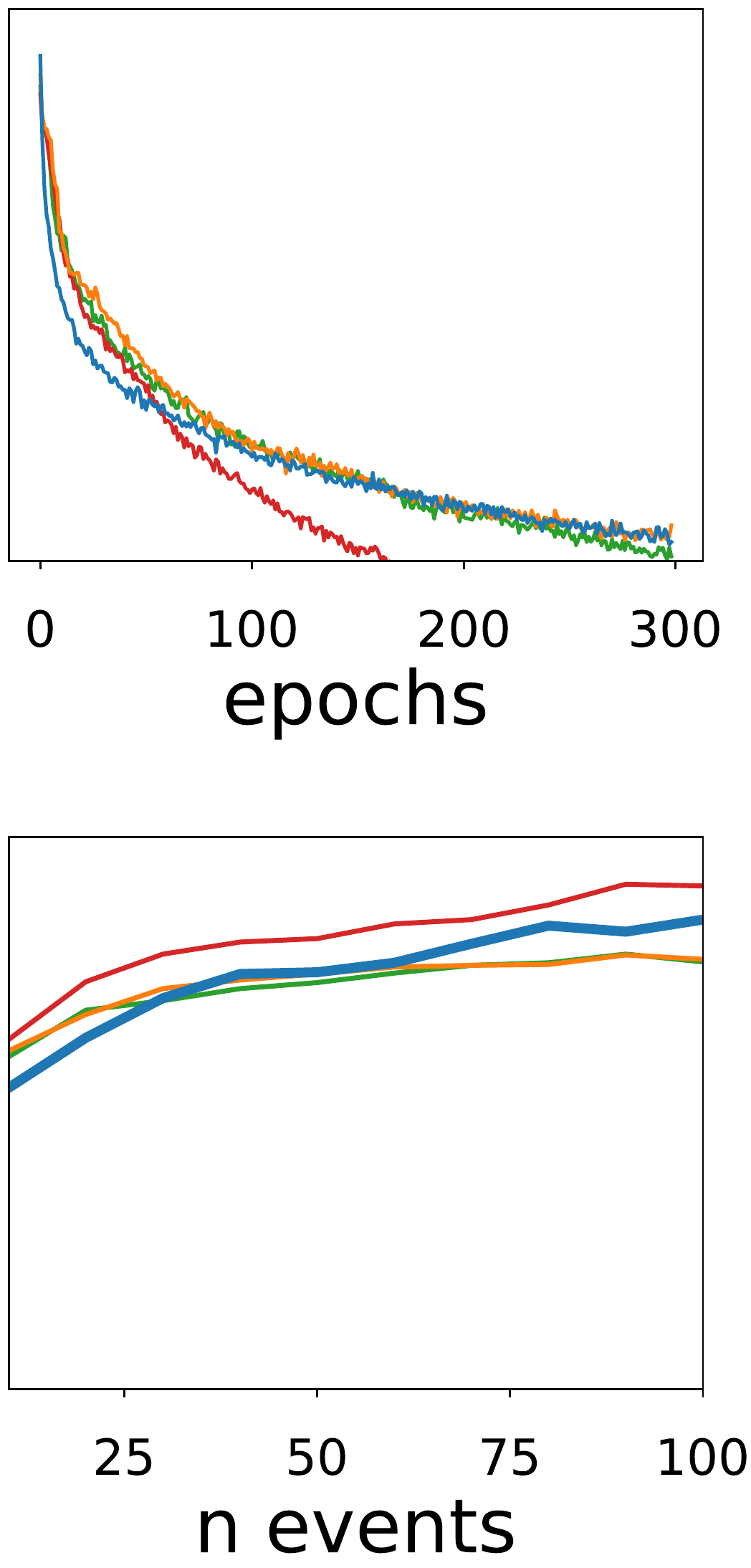}
        \caption{NCALTECH (300 epochs)}
    \end{subfigure}
    \caption{{\bf Summary of results.} Train/test losses and classification performance for INODE and multiple LSTM baselines, with increasing number of inference events per digit from 10 to 100. The three images for each dataset sub-figure correspond to training-set fraction of $20\%$ (left), $40\%$ (center), and $100\%$ (right).}
    \label{fig:results_summary}
\end{figure*}

\section{Conclusion}

This paper presents a novel approach for performing machine learning from event-camera streams. The proposed INODE model is devised to handle high-frequency event data, inherently making use of the precise timing information of each individual event, and does not require processing the raw data into different formats. 

We compared the approach to LSTM baselines on multiple DVS camera-based classification tasks. On the ASL task, the INODE significantly outperforms the baselines in fewer epochs. The network gains marginal predictive power as the complexity of the dataset increases or as the amount of data decreases. The baselines deliver a better performance only for simple datasets (MNIST) and if a large amount of data is availabile (NCALTECH).

INODE excels in the most realistic scenarios, when little training data and few events are available. This makes it suitable for real-time, low-computation settings where decisions must be taken with only few event such as collision avoidance and high-speed object recognition.

\section*{Acknoledgements}
The authors are grateful to Christian Osendorfer for his valuable input and feedback and to everyone at NNAISENSE for contributing to an inspiring R\&D environment.

\bibliography{biblio.bib}
\bibliographystyle{util/icml2020}

\clearpage
\newpage
\appendix 
\onecolumn

\section*{Appendix}

\section{Learning Dynamics}
\label{sec:appendix_figures}
\begin{figure*}[tb]
    \centering
    \includegraphics[width=.9\textwidth,height=0.25\textheight]{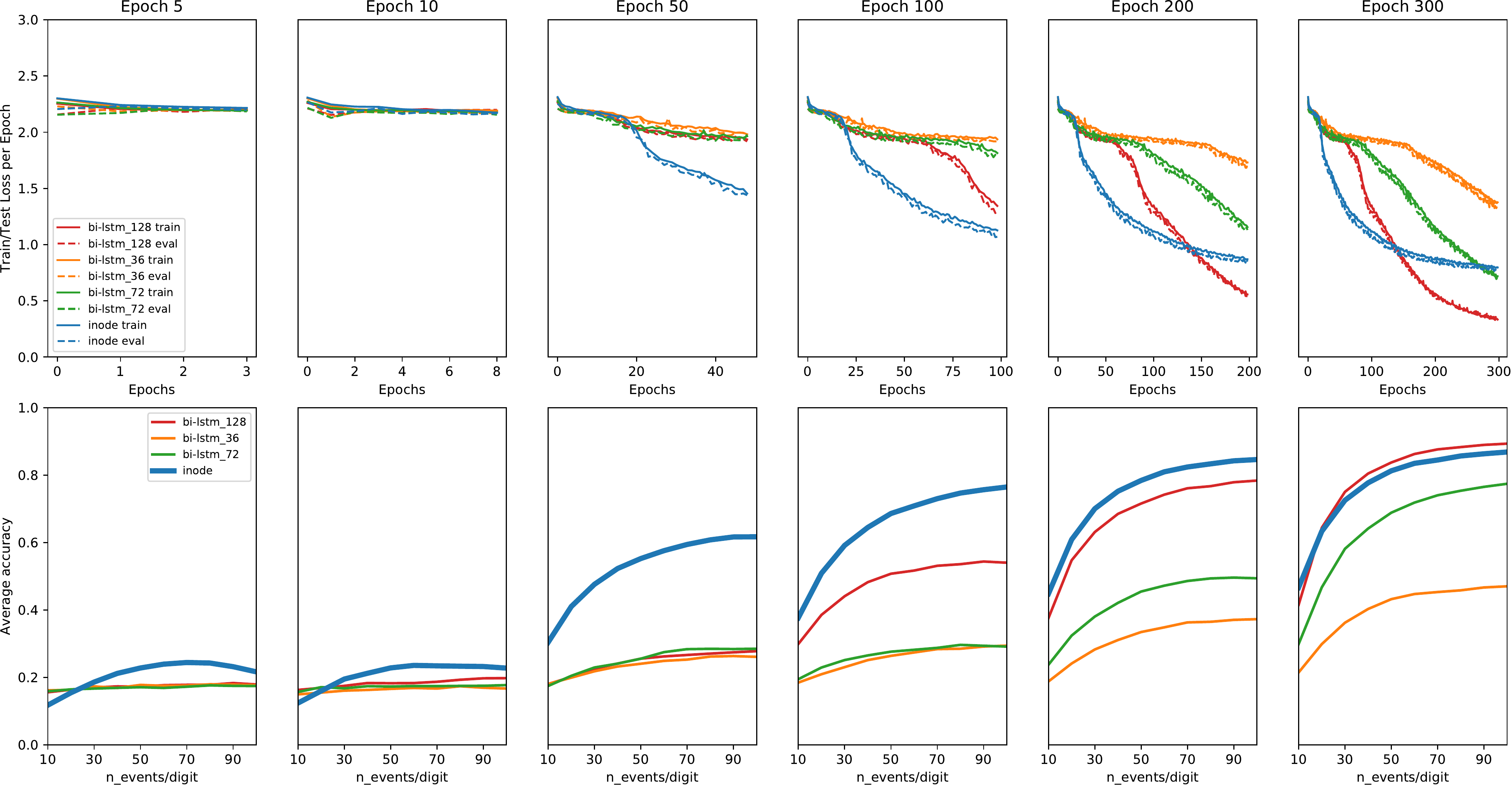}
    \vskip 0.2in
    \includegraphics[width=.9\textwidth,height=0.25\textheight]{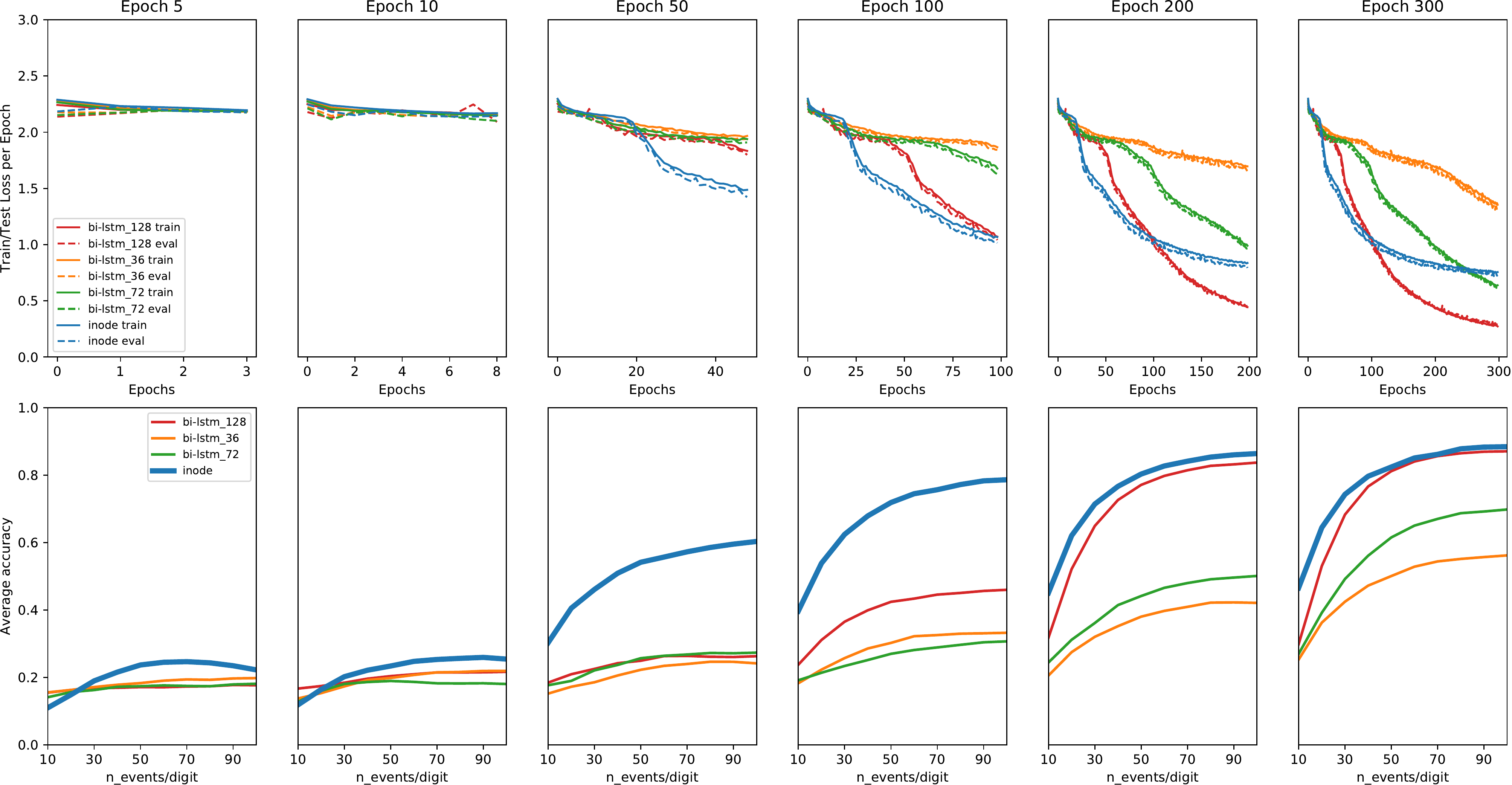}
    \vskip 0.2in
    \includegraphics[width=.9\textwidth,height=0.25\textheight]{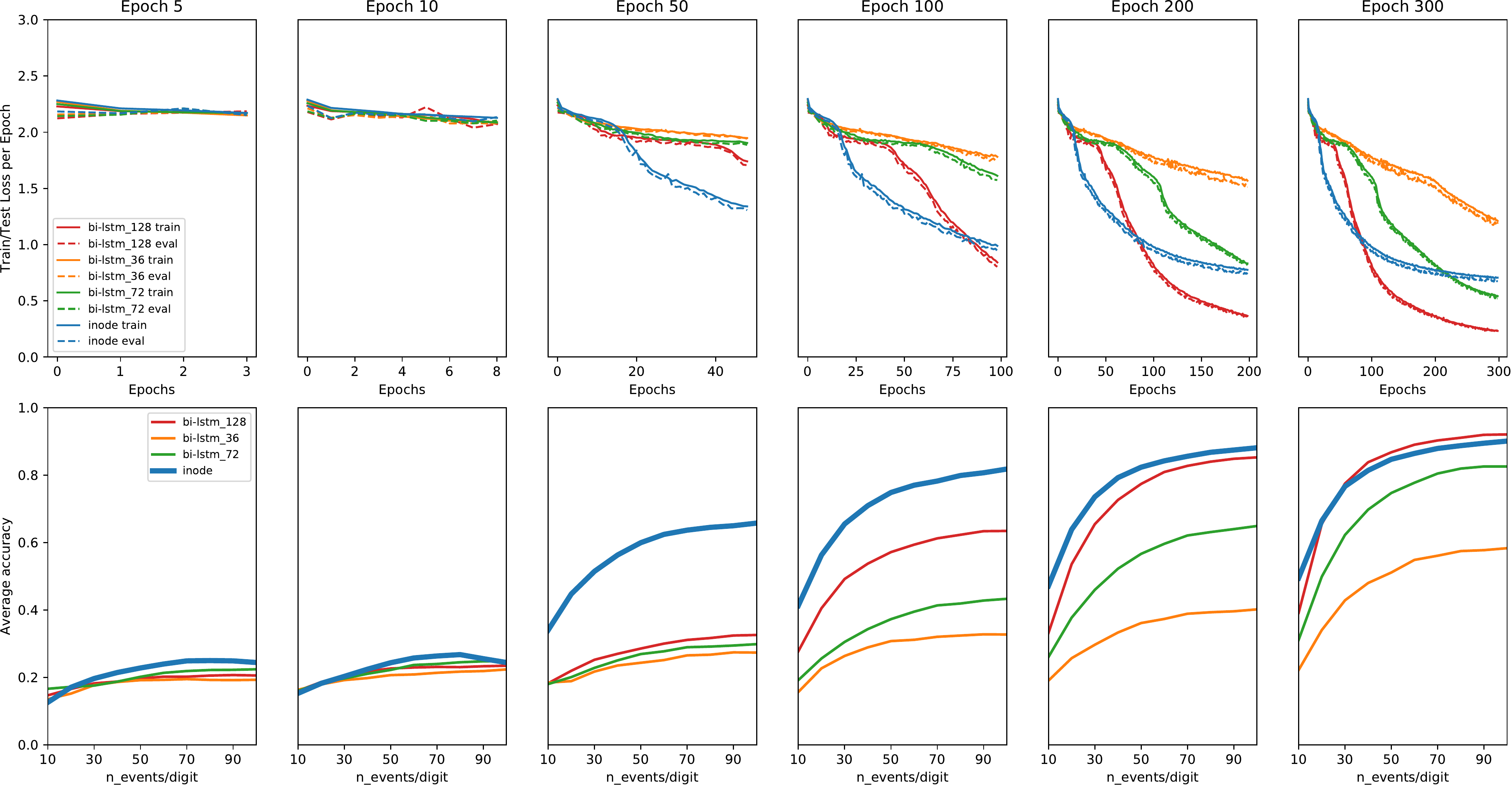}
    \caption{bi-LSTM baselines. NMNIST dataset.
    Training/test losses and classification performance for INODE and baselines increasing the number of events per digit from 10 to 100.
    Top $\rho=0.2$. Middle $\rho=0.4$. Bottom $\rho=1$.
    }\label{fig:mnistappendix}
\end{figure*}

\begin{figure*}[tb]
    \centering
    \includegraphics[width=.9\textwidth,height=0.25\textheight]{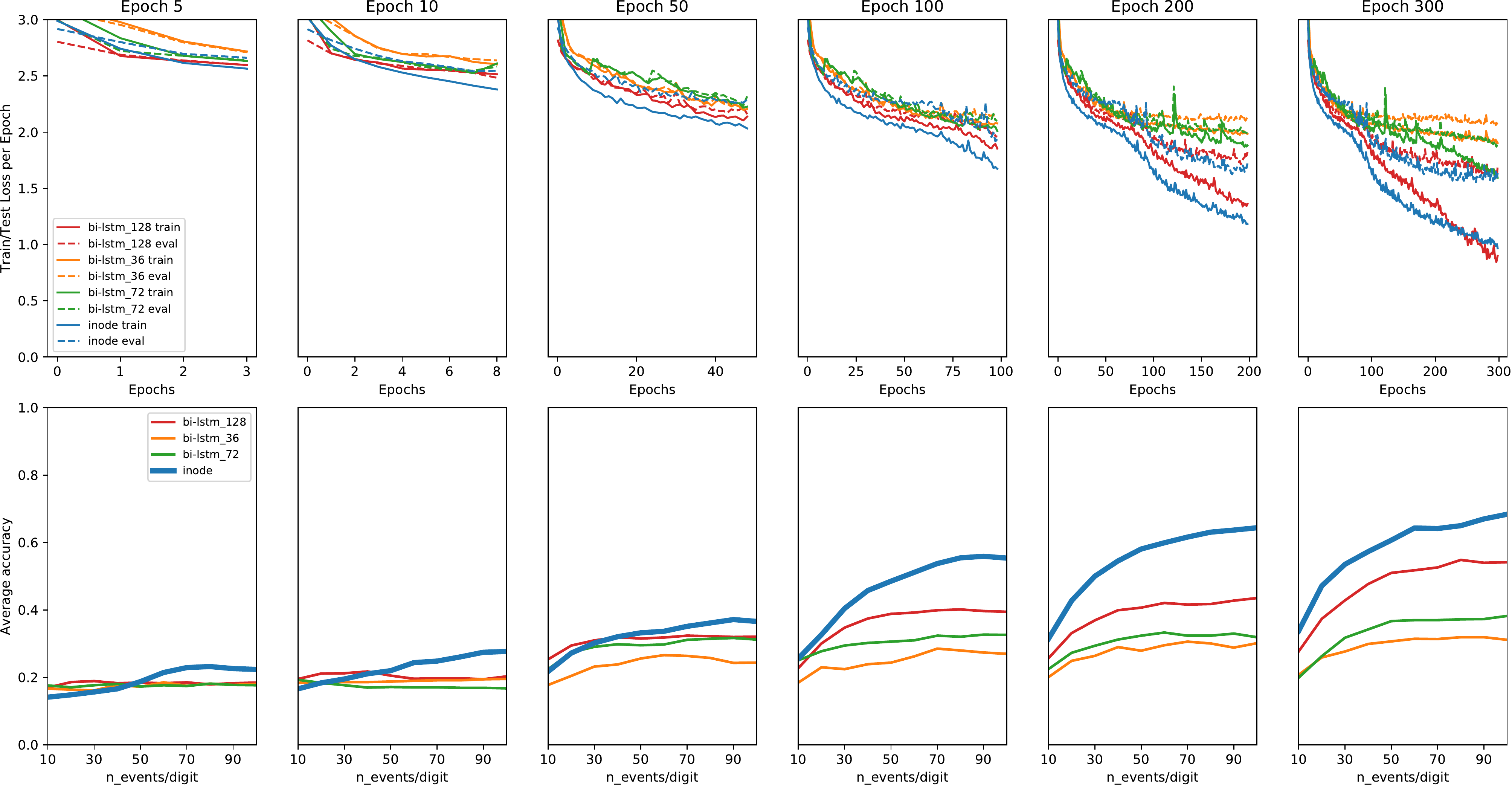}
    \vskip 0.2in
    \includegraphics[width=.9\textwidth,height=0.25\textheight]{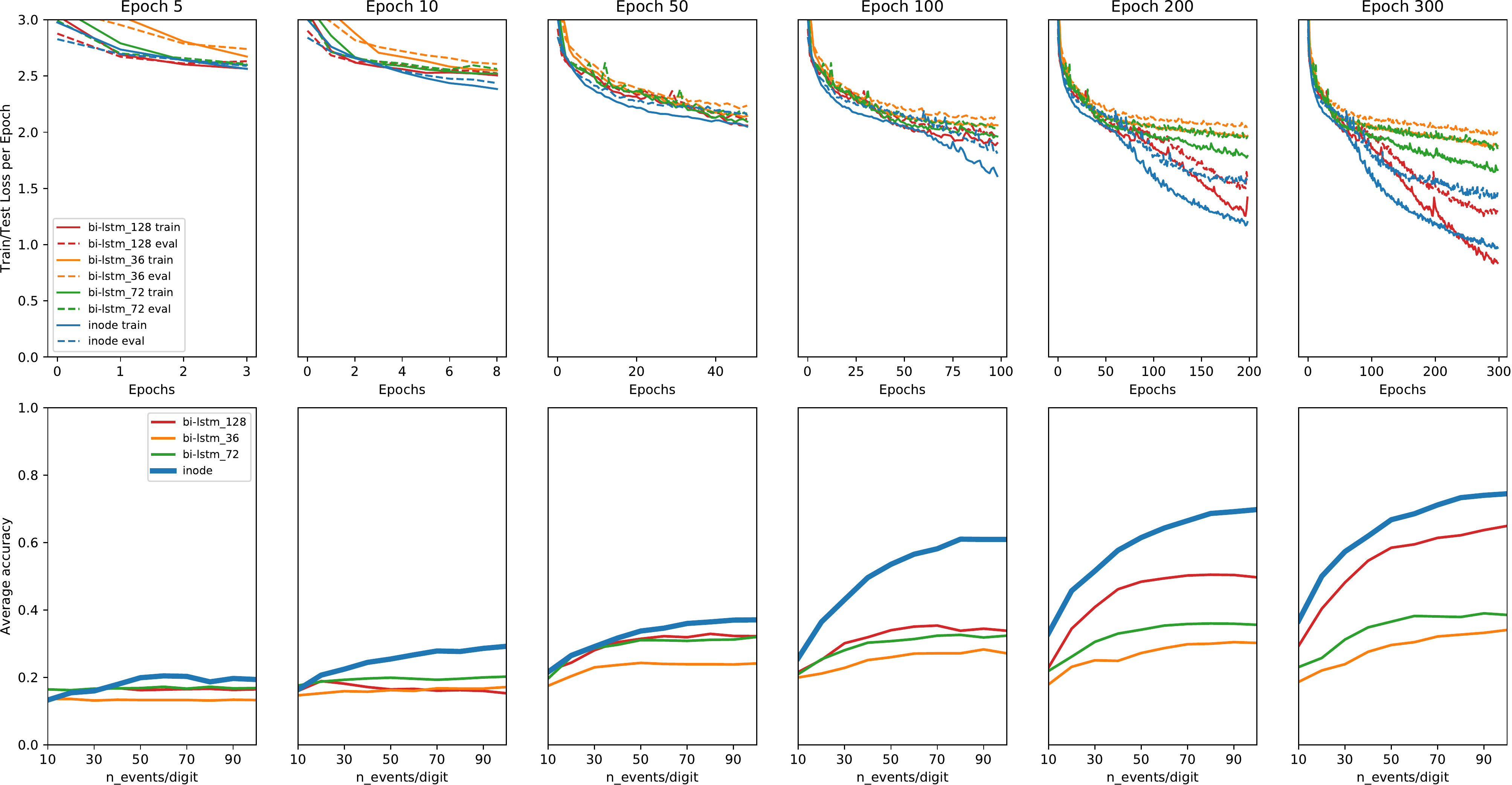}
    \vskip 0.2in
    \includegraphics[width=.9\textwidth,height=0.25\textheight]{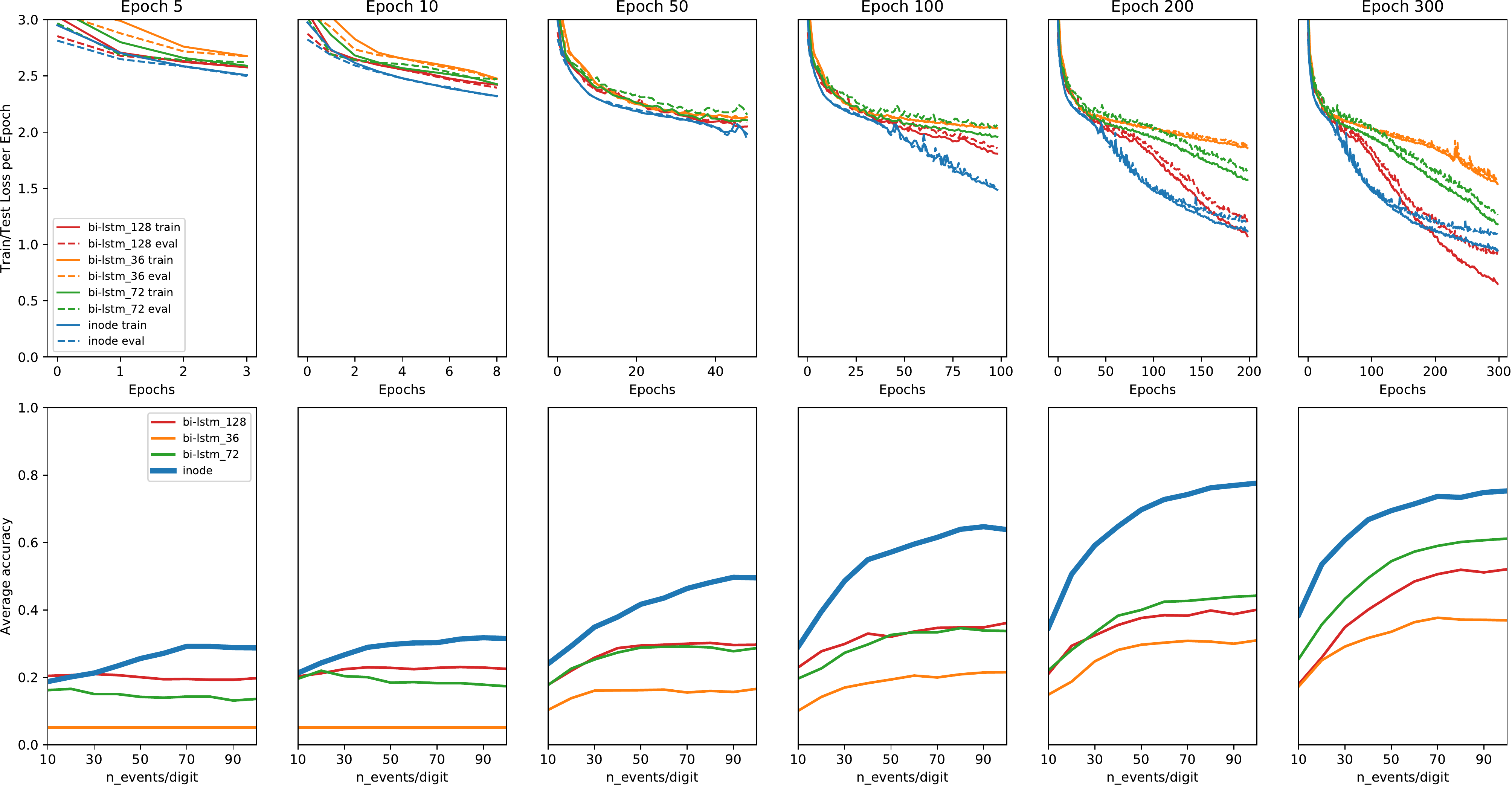}
    \caption{bi-LSTM baselines. ASL (12-16k) dataset.
    Training/test losses and classification performance for INODE and baselines increasing the number of events per digit from 10 to 100.
    Top $\rho=0.2$. Middle $\rho=0.4$. Bottom $\rho=1$.
    }\label{fig:aslappendix}
\end{figure*}

\begin{figure*}[tb]
    \centering
    \includegraphics[width=.9\textwidth,height=0.25\textheight]{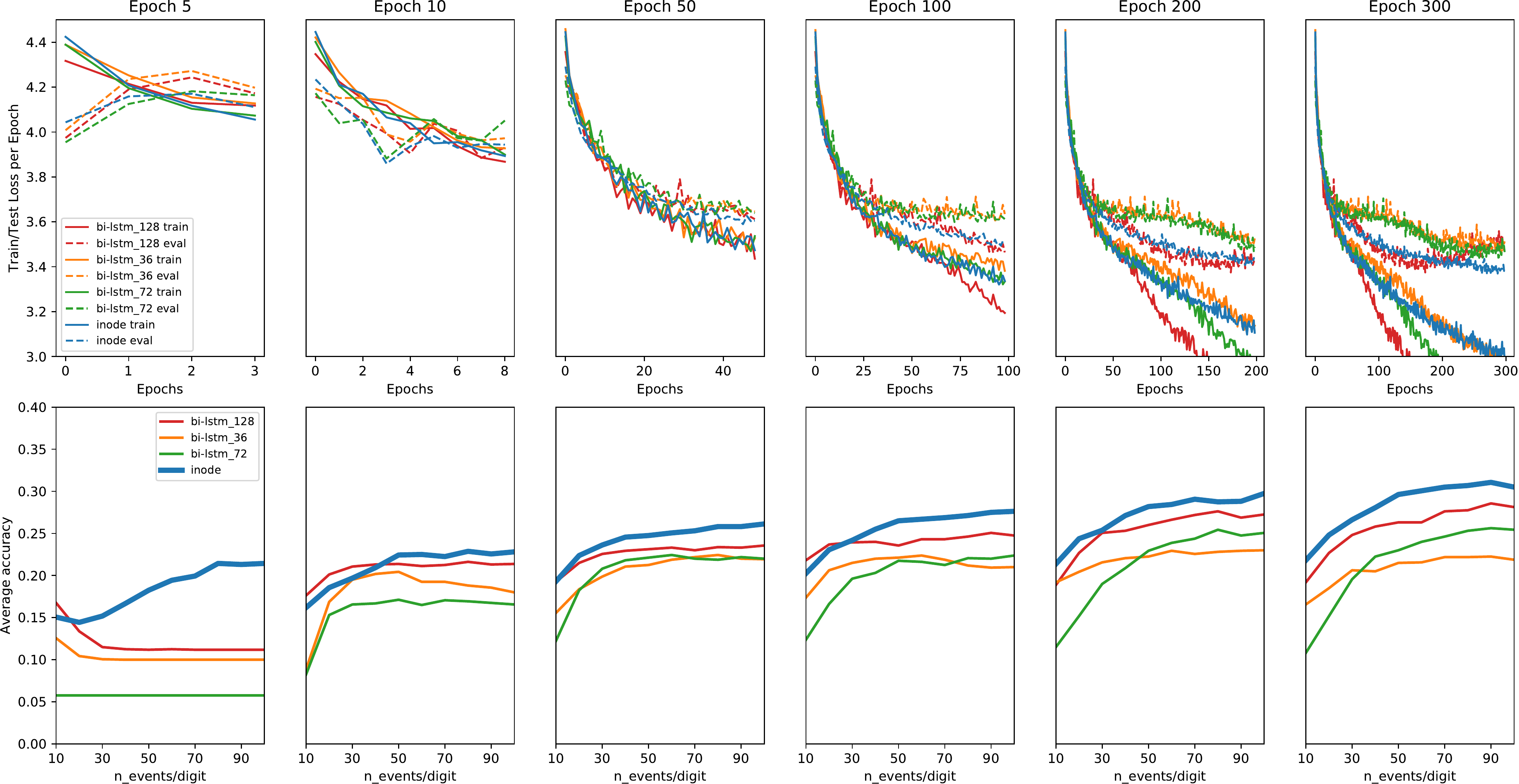}
    \vskip 0.2in
    \includegraphics[width=.9\textwidth,height=0.25\textheight]{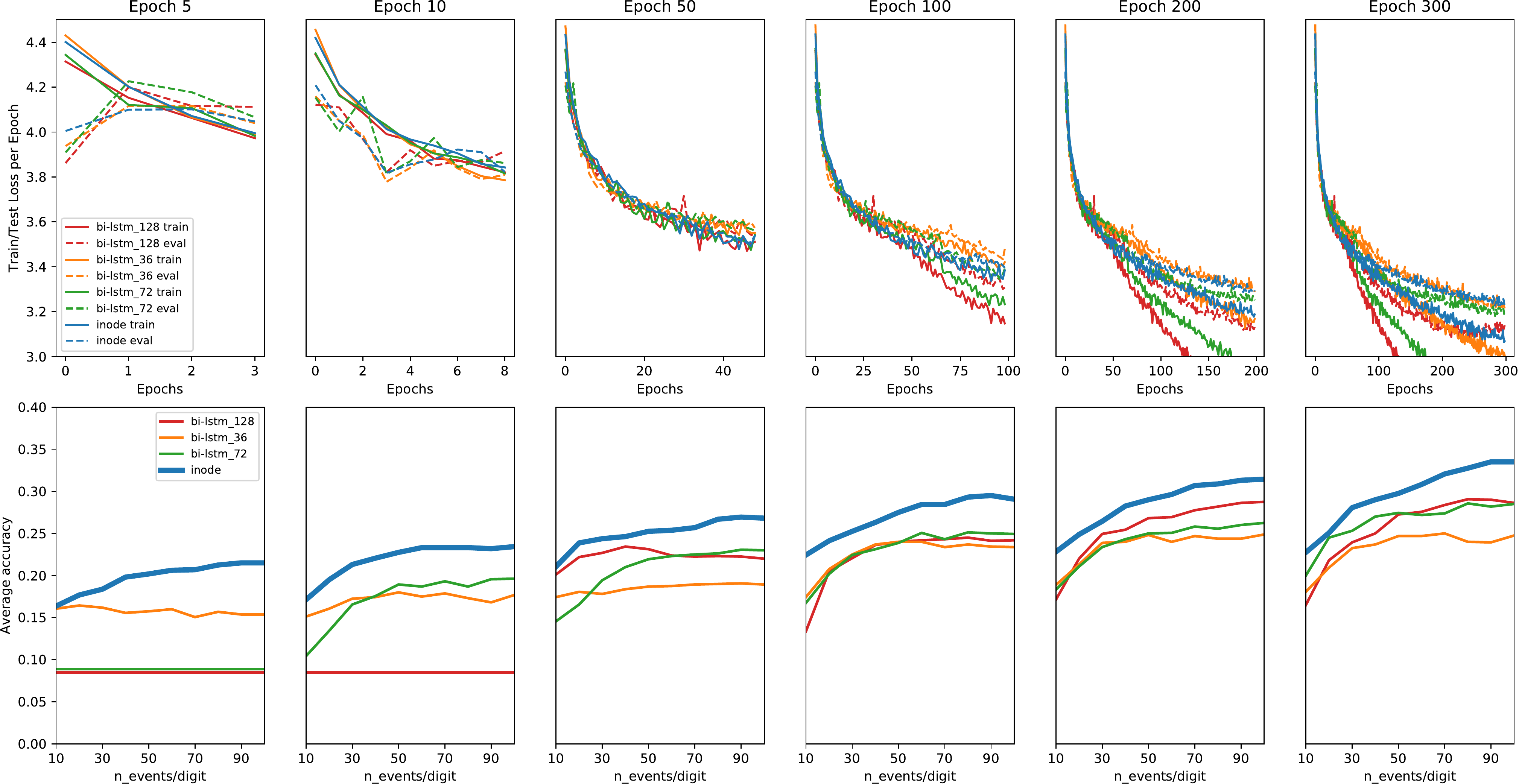}
    \vskip 0.2in
    \includegraphics[width=.9\textwidth,height=0.25\textheight]{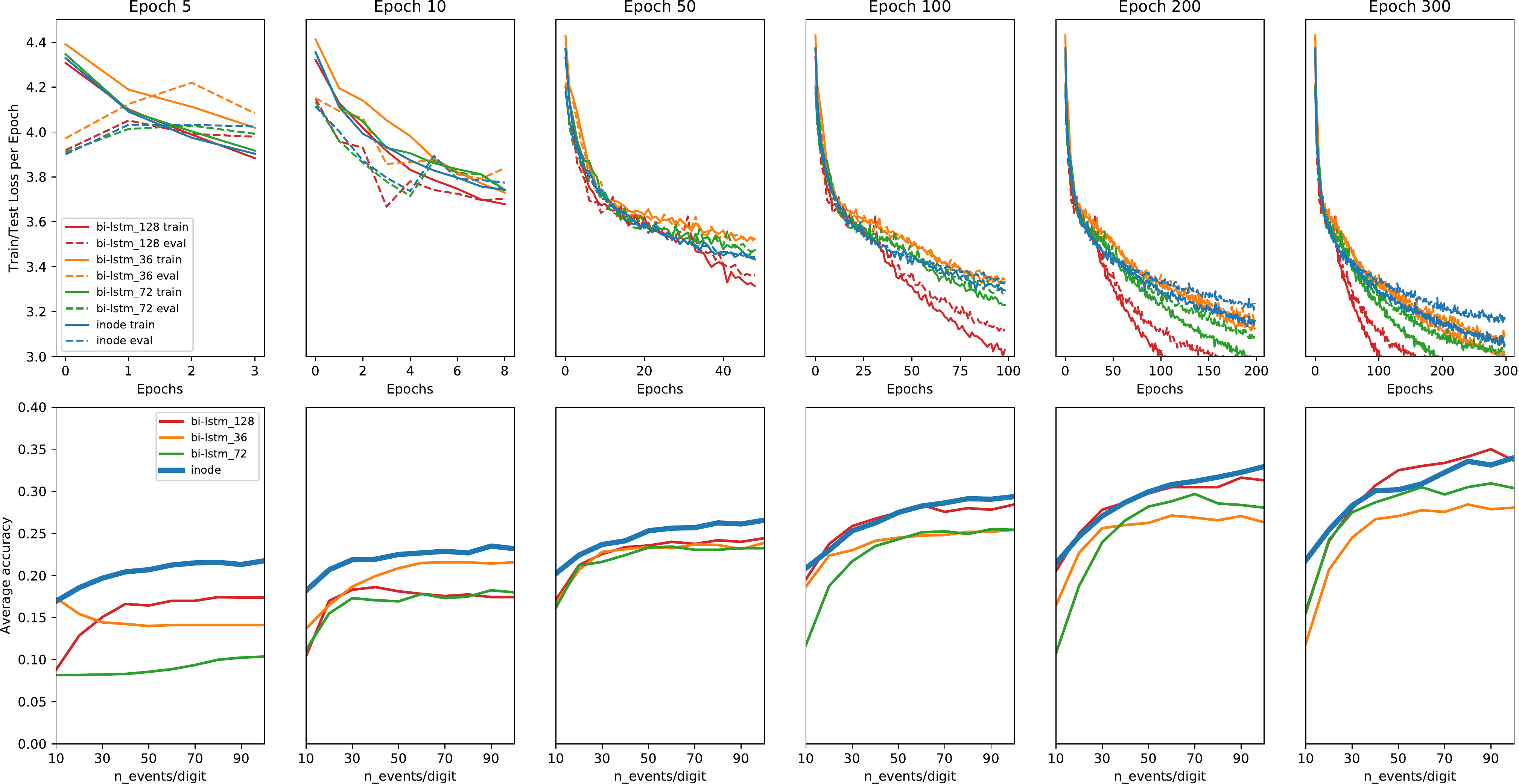}
    \caption{bi-LSTM baselines. NCALTECH dataset.
    Training/test losses and classification performance for INODE and baselines increasing the number of events per digit from 10 to 100.
    Top $\rho=0.2$. Middle $\rho=0.4$. Bottom $\rho=1$.
    }\label{fig:ncaltechppendix}
\end{figure*}

\begin{figure*}[tb]
    \centering
    \includegraphics[width=.9\textwidth,height=0.25\textheight]{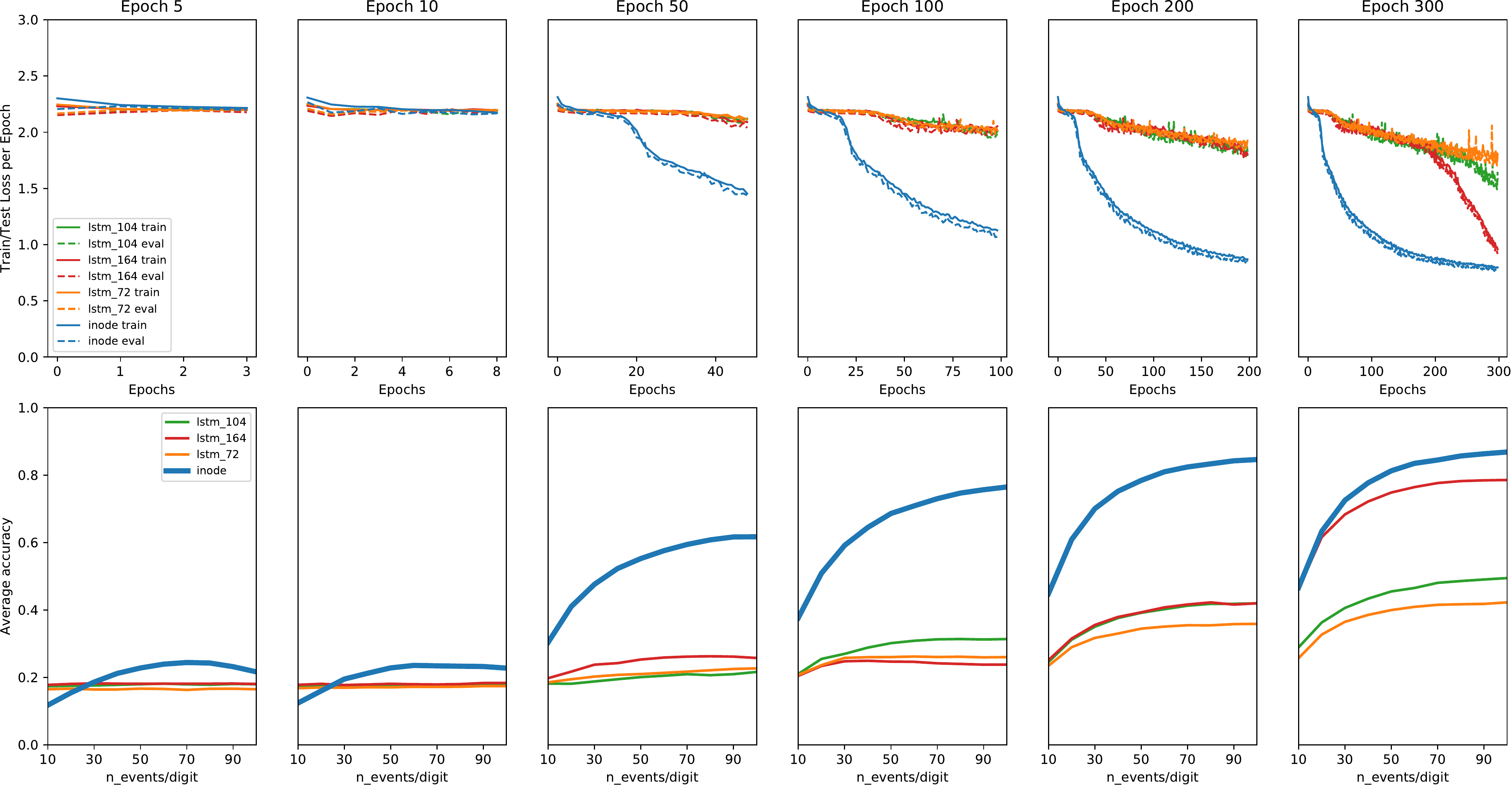}
    \vskip 0.2in
    \includegraphics[width=.9\textwidth,height=0.25\textheight]{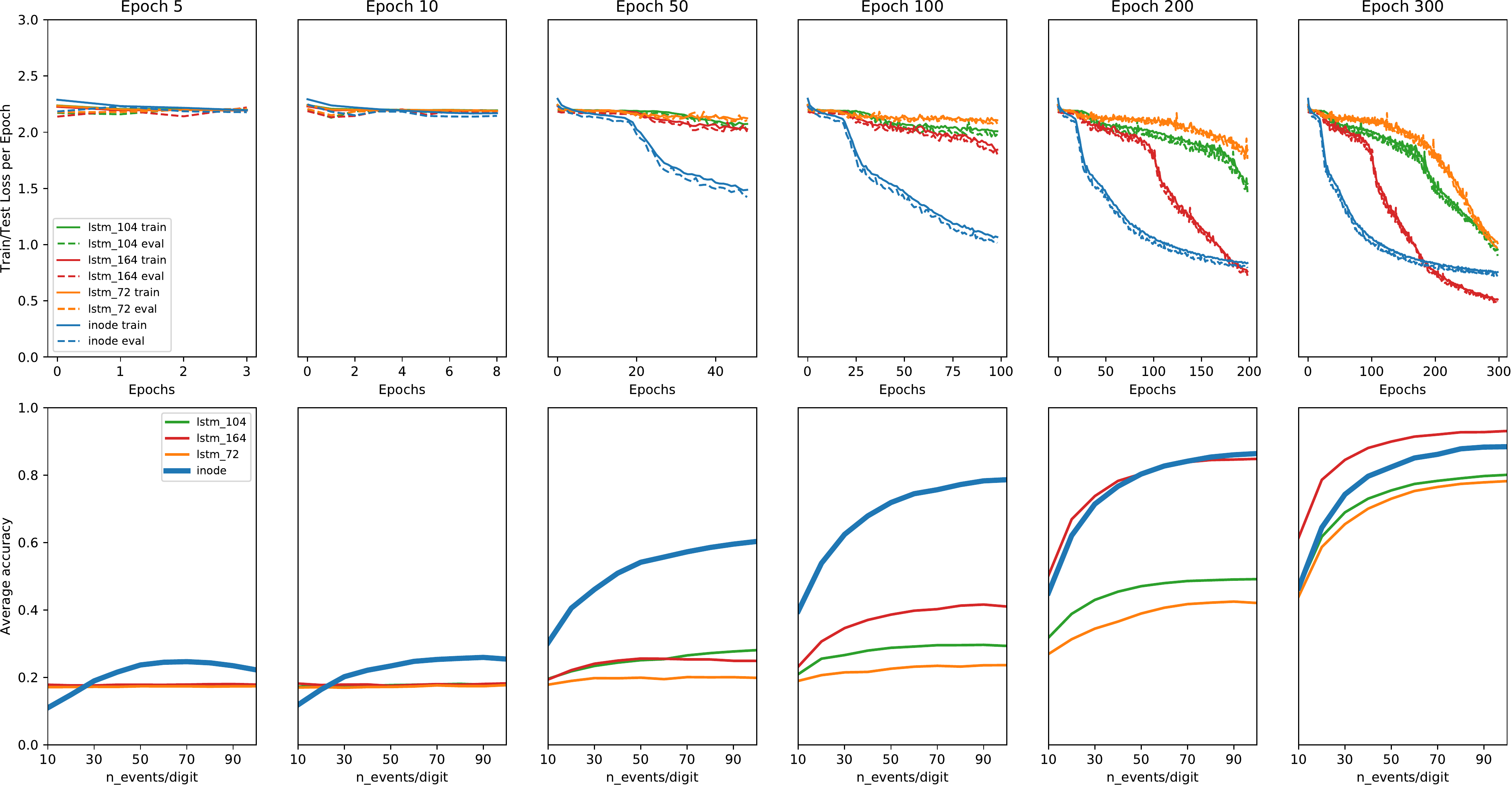}
    \vskip 0.2in
    \includegraphics[width=.9\textwidth,height=0.25\textheight]{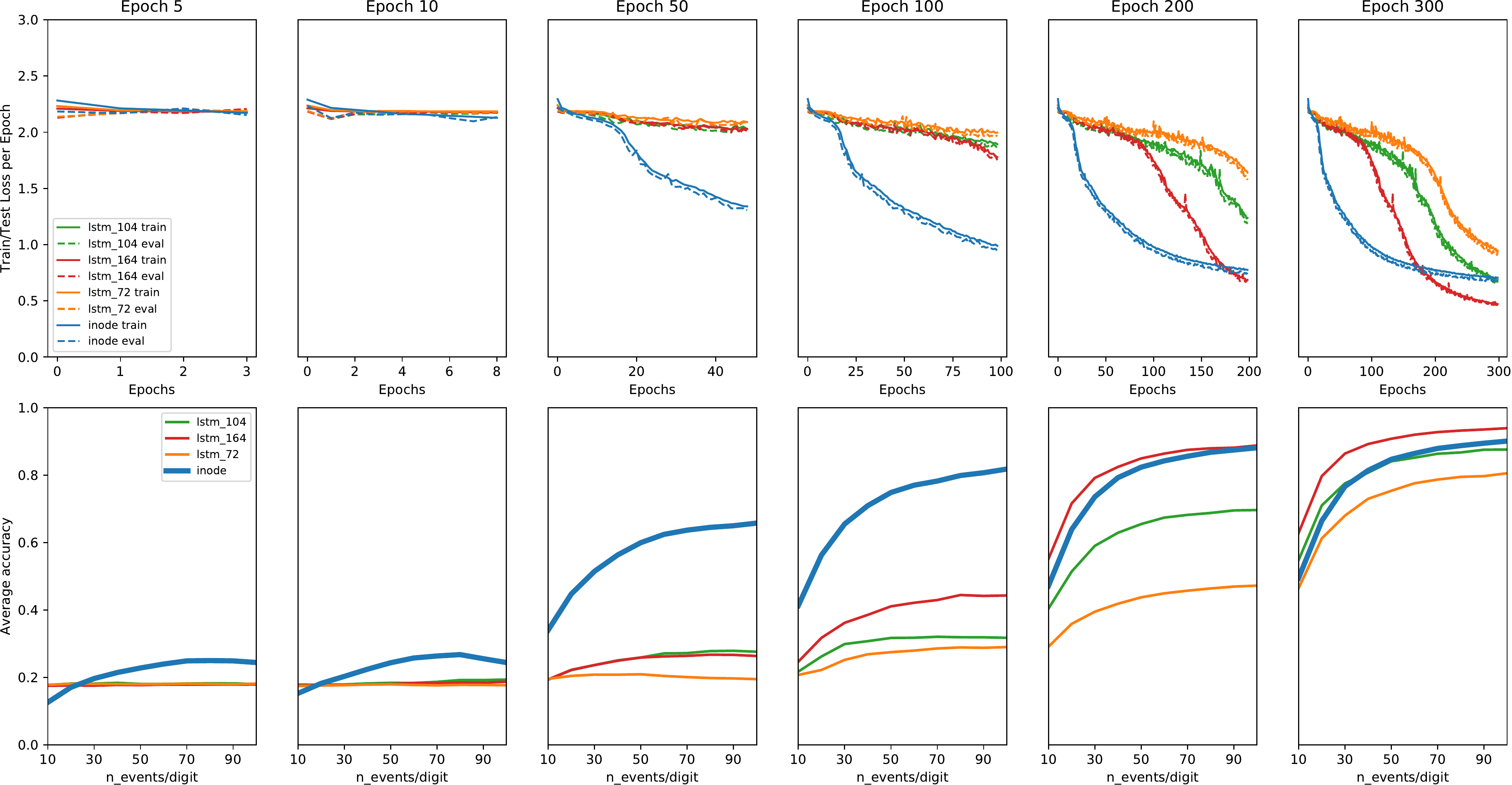}
    \caption{LSTM baselines. NMNIST dataset.
    Training/test losses and classification performance for INODE and baselines increasing the number of events per digit from 10 to 100.
    Top $\rho=0.2$. Middle $\rho=0.4$. Bottom $\rho=1$.
    }\label{fig:mnistappendix_uni}
\end{figure*}

\begin{figure*}[tb]
    \centering
    \includegraphics[width=.9\textwidth,height=0.25\textheight]{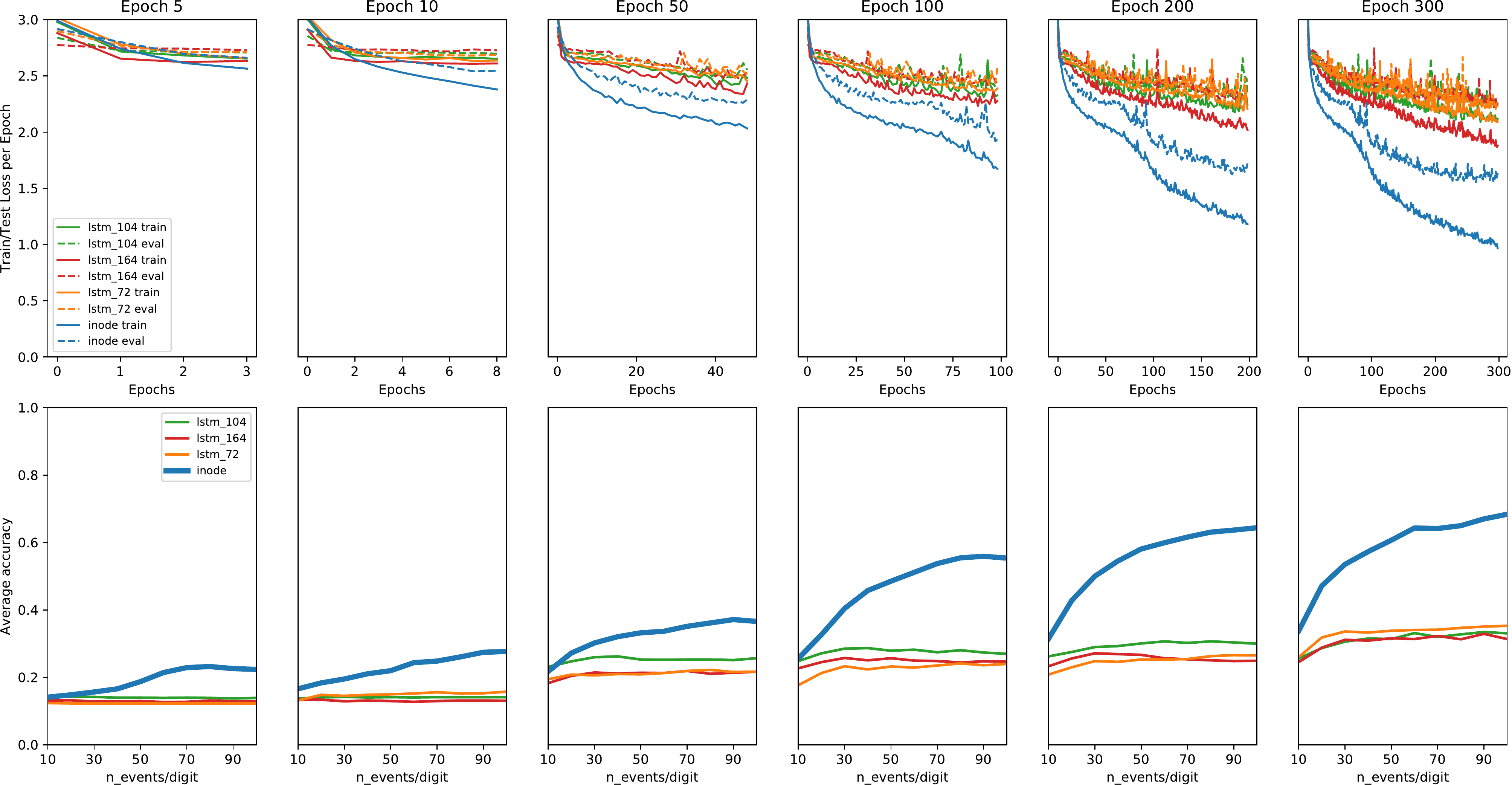}
    \vskip 0.2in
    \includegraphics[width=.9\textwidth,height=0.25\textheight]{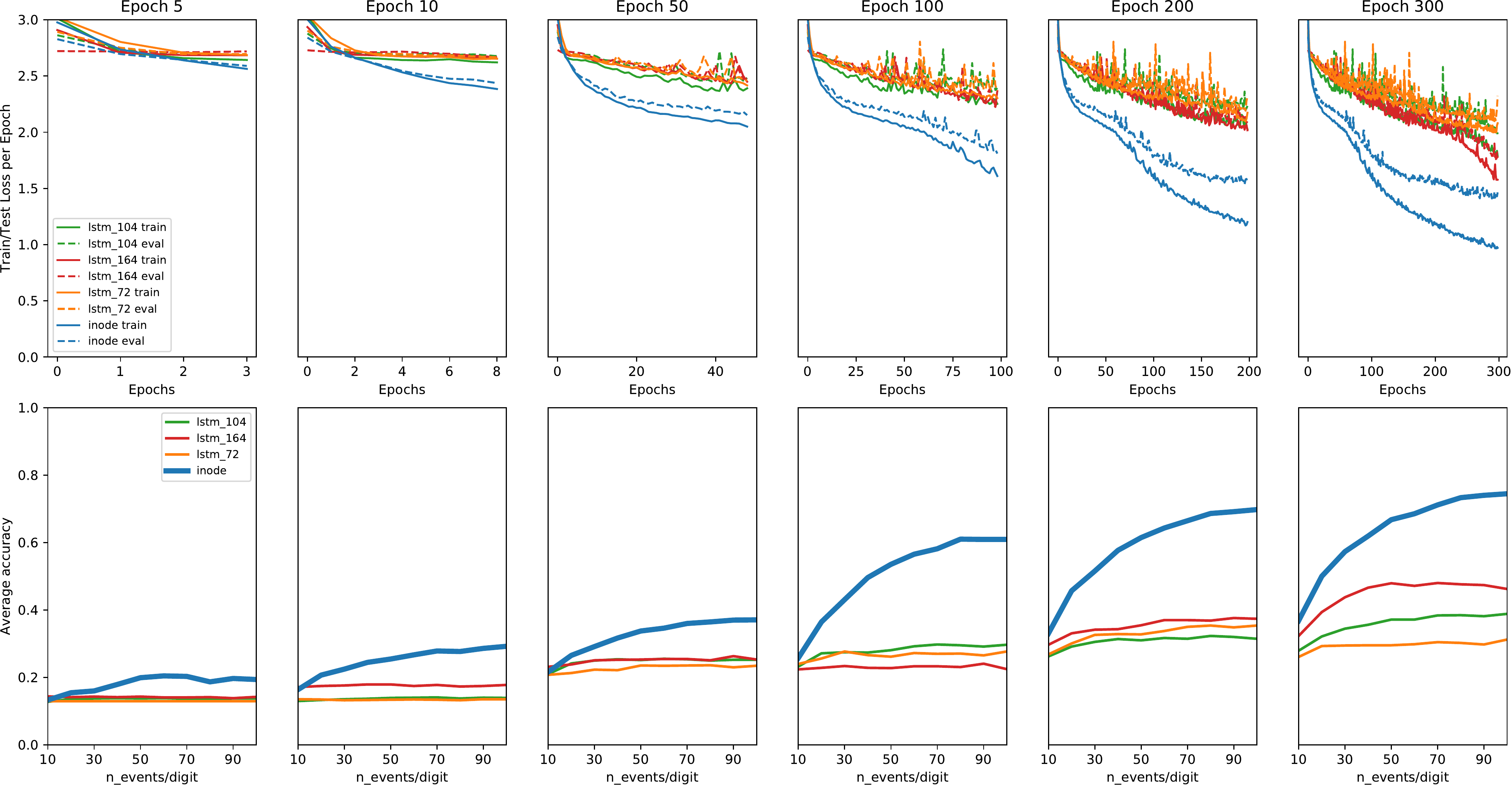}
    \vskip 0.2in
    \includegraphics[width=.9\textwidth,height=0.25\textheight]{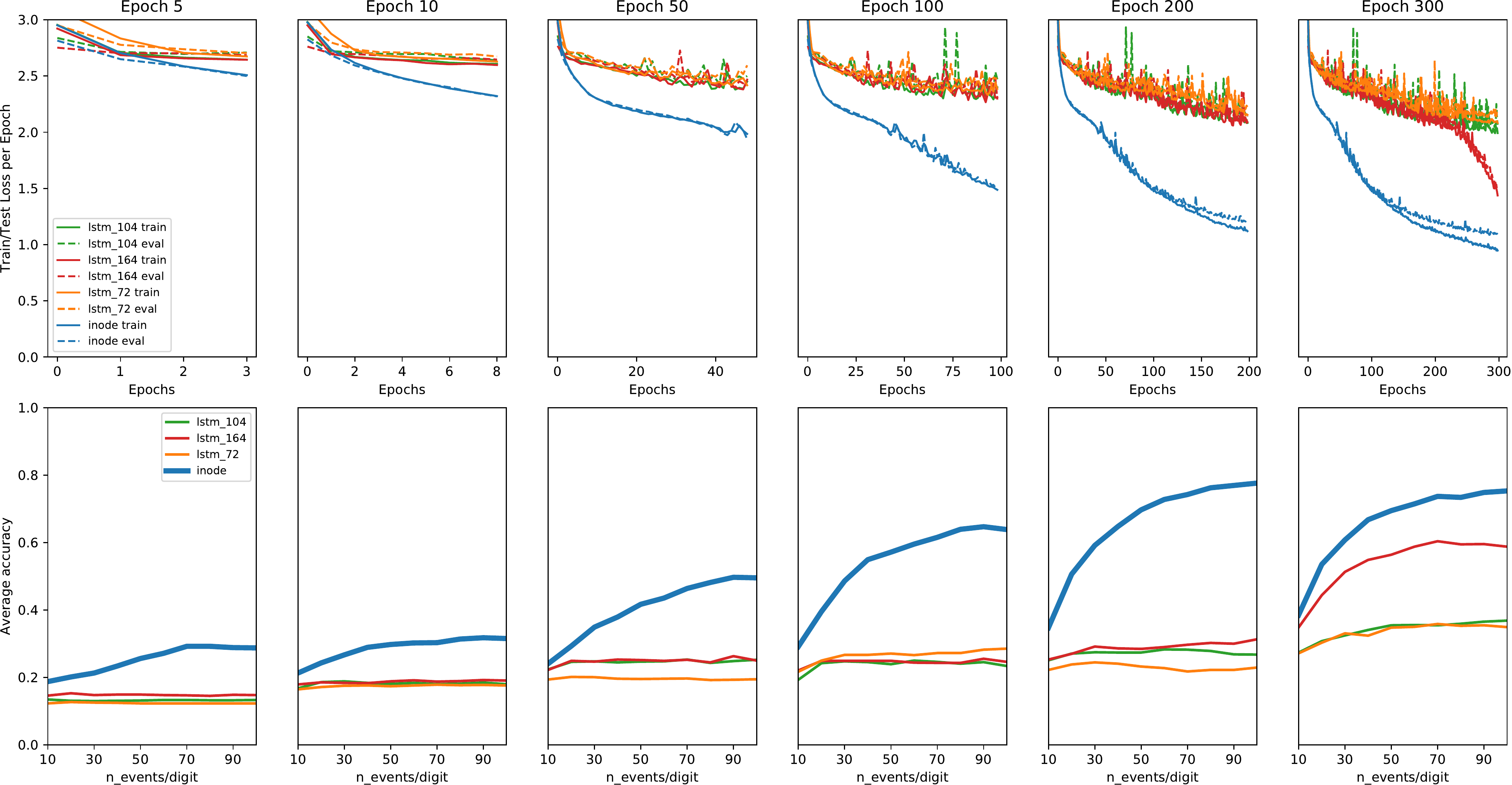}
    \caption{LSTM baselines. ASL (12-16k) dataset.
    Training/test losses and classification performance for INODE and baselines increasing the number of events per digit from 10 to 100.
    Top $\rho=0.2$. Middle $\rho=0.4$. Bottom $\rho=1$.
    }\label{fig:aslappendix_uni}
\end{figure*}

\begin{figure*}[tb]
    \centering
    \includegraphics[width=.9\textwidth,height=0.25\textheight]{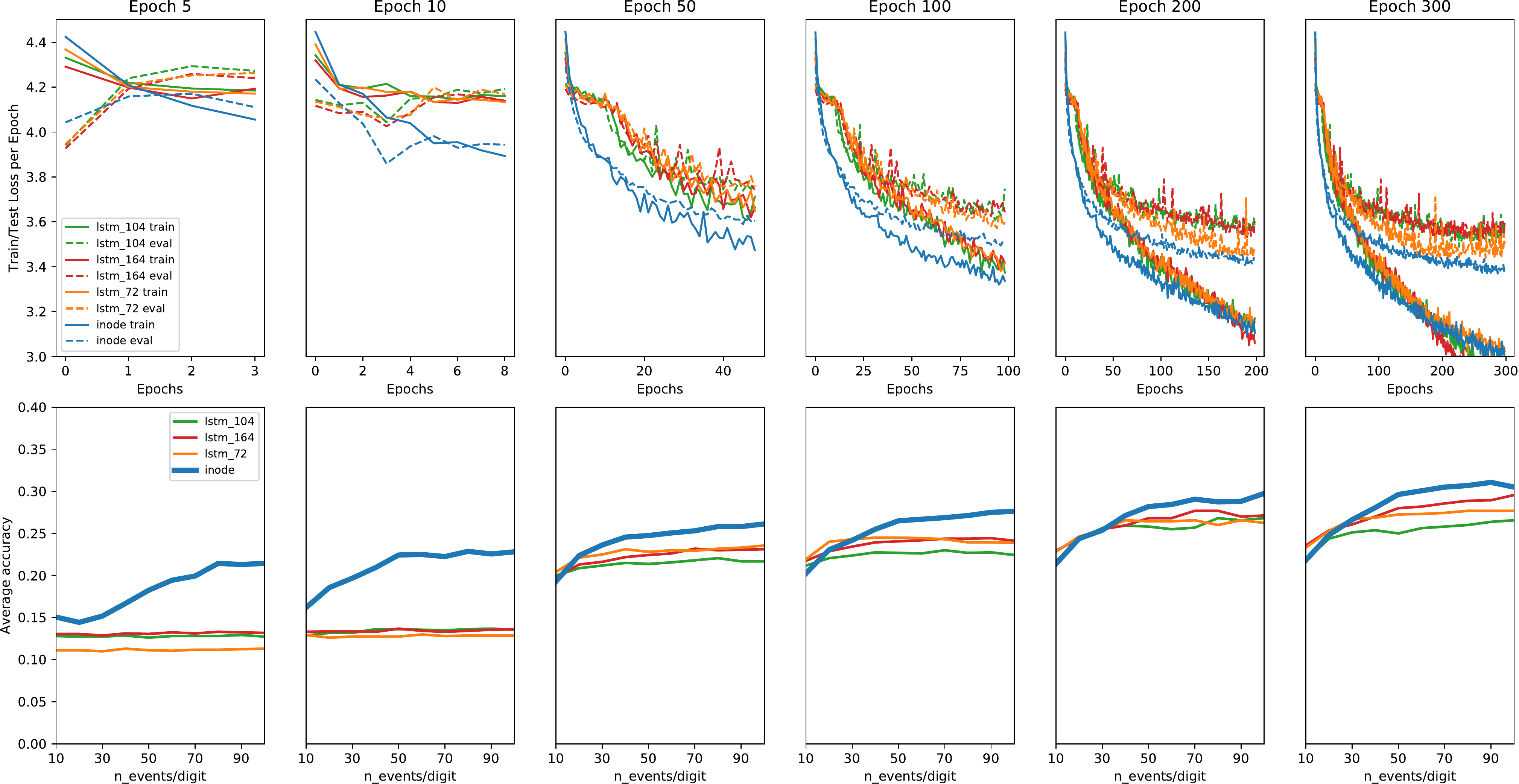}
    \vskip 0.2in
    \includegraphics[width=.9\textwidth,height=0.25\textheight]{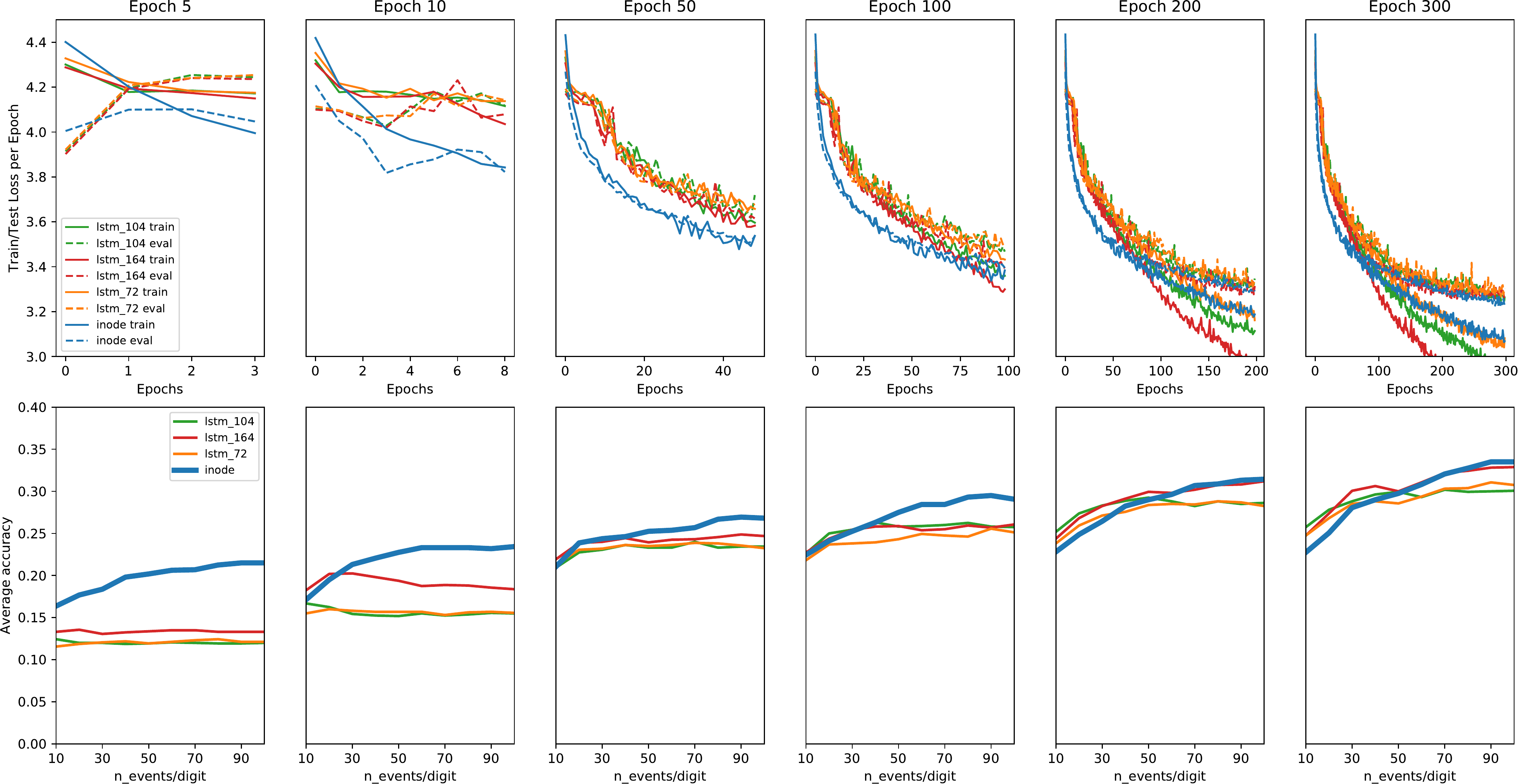}
    \vskip 0.2in
    \includegraphics[width=.9\textwidth,height=0.25\textheight]{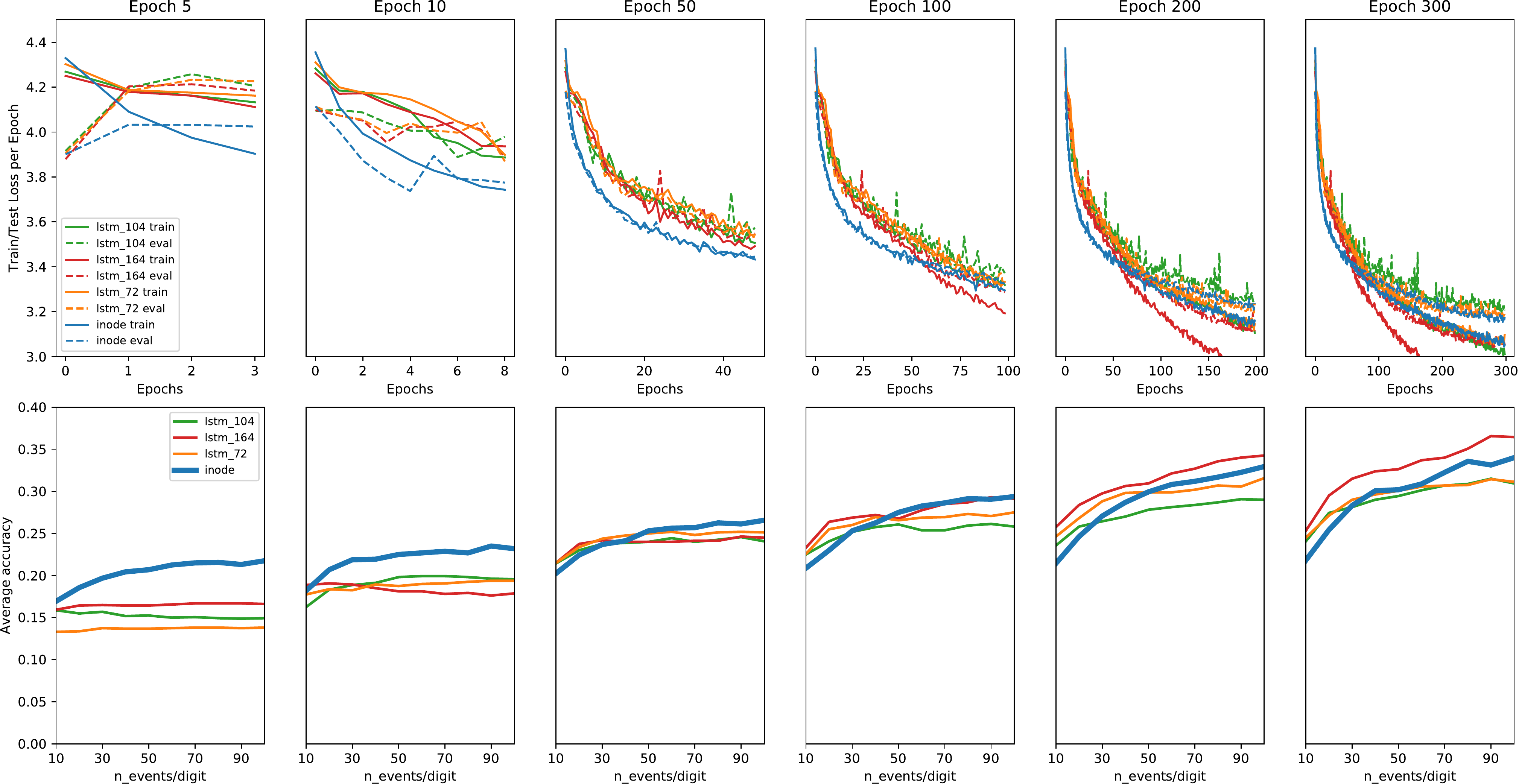}
    \caption{LSTM baselines. NCALTECH dataset.
    Training/test losses and classification performance for INODE and baselines increasing the number of events per digit from 10 to 100.
    Top $\rho=0.2$. Middle $\rho=0.4$. Bottom $\rho=1$.
    }\label{fig:ncaltechppendix_uni}
\end{figure*}

The learning curves and online inference trajectories for the proposed method and the bidirectional LSTM baselines are depicted in Figure \ref{fig:mnistappendix}, \ref{fig:ncaltechppendix} and \ref{fig:aslappendix} for, respectively, the NMNIST, NCALTECH and ASL dataset. On ASL, our method consistently outperforms the baselines b a large margin; and in Figure  \ref{fig:mnistappendix_uni}, \ref{fig:ncaltechppendix_uni} and \ref{fig:aslappendix_uni} for the LSTM baselines.

\end{document}